\documentclass[journal]{IEEEtran}

\newif\ifanonymoussubmission
\anonymoussubmissionfalse

\usepackage{url}
\usepackage[compress]{cite}
\usepackage{amsmath, amssymb, amsfonts}
\usepackage{mathtools, bm}
\usepackage{pifont}
\usepackage{siunitx}
\usepackage{bigints}
\usepackage{textcomp}
\usepackage{multirow} 
\usepackage{array}
\usepackage{adjustbox}
\usepackage{booktabs}
\usepackage{tabularx}
\usepackage{makecell}
\newcolumntype{L}[1]{>{\raggedright\arraybackslash}p{#1}}
\newcolumntype{C}[1]{>{\centering\arraybackslash}p{#1}}
\newcolumntype{Y}[1]{>{\hsize=#1\hsize\raggedright\arraybackslash}X}
\usepackage{graphicx}
\usepackage{stfloats}
\usepackage{xcolor}
\usepackage{listings}
\usepackage{hyperref}
\usepackage{cleveref}
\ifanonymoussubmission
  \hypersetup{
    pdfauthor={Anonymous Authors},
    pdftitle={SoRoMoX: Fast, Differentiable, and Parallelizable Soft Robot Models}
  }
\else
  \hypersetup{
    pdfauthor={Maximilian Stölzle, Solange Gribonval, Daniel Feliu-Talegon, Vito Daniele Perfetta, Michele Martini, Chuhan Zhang, Kiwan Wong, Mohammed Tarnini, Anup Teejo Mathew, Federico Renda, Daniela Rus, and Cosimo Della Santina},
    pdftitle={SoRoMoX: Fast, Differentiable, and Parallelizable Soft Robot Models}
  }
\fi
\usepackage[most]{tcolorbox}
\newtcolorbox{soromoxbox}[1][Sidebar]{breakable, enhanced, colback=black!4,
  colframe=black!55, coltitle=white, fonttitle=\bfseries\sffamily, title={#1},
  boxrule=0.6pt, arc=2pt, left=6pt, right=6pt, top=5pt, bottom=5pt,
  before skip=10pt, after skip=10pt}

\newcommand{\cmark}{\ding{51}}%
\newcommand{\xmark}{\ding{55}}%
\providecommand{\Ddot}[1]{\ddot{#1}}

\usepackage{acro}
\newcommand{\newac}[2]{\DeclareAcronym{#1}{short=#1,long=#2}}
\newac{AD}{Automatic Differentiation}
\newac{API}{Application Programming Interface}
\newac{AOT}{Ahead-of-Time}
\newac{CBF}{Control Barrier Function}
\newac{CLF}{Control Lyapunov Function}
\newac{CC}{Constant Curvature}
\newac{CI}{Continuous Integration}
\newac{CS}{Constant Strain}
\newac{COM}{Center of Mass}
\newac{DCM}{Discretized Cosserat Rod Model}
\newac{DOF}{Degrees of Freedom}
\newac{EB}{Euler-Bernoulli}
\newac{EOM}{Equations of Motion}
\newac{FEM}{Finite Element Method}
\newac{GD}{Gradient Descent} 
\newac{GVS}{Geometric Variable Strain}
\newac{HSA}{Handed Shearing Auxetics}
\newac{HOCBF}{High-Order Control Barrier Function}
\newac{HOCLF}{High-Order Control Lyapunov Function}
\newac{JIT}{Just-in-Time}
\newac{JSRM}{JAX Soft Robot Modeling}
\newac{LNN}{Lagrangian Neural Network}
\newac{LSTM}{Long Short-Term Memory}
\newac{MAE}{Mean Absolute Error}
\newac{ML}{Machine Learning}
\newac{MSE}{Mean Squared Error}
\newac{MPC}{Model Predictive Control}
\newac{MPPI}{Model Predictive Path Integral}
\newac{ODE}{Ordinary Differential Equation}
\newac{PAC}{Piecewise Affine Curvature}
\newac{PCS}{Piecewise Constant Strain}
\newac{PCC}{Piecewise Constant Curvature}
\newac{PDE}{Partial Differential Equation}
\newac{PPO}{Proximal Policy Optimization}
\newac{QP}{Quadratic Program}
\newac{RL}{Reinforcement Learning}
\newac{RMSE}{Root Mean-Squared Error}
\newac{SoRoMoX}{Soft Robot Models in jaX}
\newac{SOTA}{State-of-the-Art}
\newac{URDF}{Unified Robot Description Format}
\newac{VS}{Variable Strain}

\definecolor{soromoxcodebg}{RGB}{248,250,252}
\definecolor{soromoxcodeframe}{RGB}{190,198,208}
\definecolor{soromoxlinenumber}{RGB}{118,128,142}
\definecolor{soromoxkeyword}{RGB}{31,88,168}
\definecolor{soromoxclass}{RGB}{126,63,152}
\definecolor{soromoxfunction}{RGB}{23,112,99}
\definecolor{soromoxstring}{RGB}{34,124,70}
\definecolor{soromoxcomment}{RGB}{99,110,123}
\definecolor{soromoxoperator}{RGB}{165,80,30}

\lstdefinelanguage{SoRoMoXPython}[]{Python}{
    morekeywords=[2]{PCS,PCSParams,SystemState,GVS,GVSSegment,LinkSpec,JointSpec,StrainBasisSpec,IsotropicMaterialParams,ViserRenderer,OperationalSpaceDynamics,ActuationSpaceDynamics},
    morekeywords=[3]{jnp,jax,robot,params,trajectory,renderer},
}

\lstdefinelanguage{SoRoMoXBash}{
    sensitive=true,
    morekeywords={cd,git,uv,pip,install,sync,clone},
    morecomment=[l]{\#},
    morestring=[b]",
    morestring=[b]',
}

\lstdefinestyle{soromoxcommon}{
    basicstyle=\ttfamily\scriptsize,
    backgroundcolor=\color{soromoxcodebg},
    frame=single,
    rulecolor=\color{soromoxcodeframe},
    framerule=0.35pt,
    framesep=4pt,
    xleftmargin=4pt,
    xrightmargin=4pt,
    numbers=left,
    numberstyle=\tiny\color{soromoxlinenumber},
    numbersep=6pt,
    showstringspaces=false,
    keepspaces=true,
    columns=fullflexible,
    breaklines=true,
    breakatwhitespace=true,
    tabsize=4,
    captionpos=b,
    aboveskip=0pt,
    belowskip=0pt,
}

\lstdefinestyle{soromoxpython}{
    style=soromoxcommon,
    language=SoRoMoXPython,
    keywordstyle=\color{soromoxkeyword}\bfseries,
    keywordstyle=[2]\color{soromoxclass}\bfseries,
    keywordstyle=[3]\color{soromoxfunction},
    commentstyle=\color{soromoxcomment}\itshape,
    stringstyle=\color{soromoxstring},
}

\lstdefinestyle{soromoxbash}{
    style=soromoxcommon,
    language=SoRoMoXBash,
    basicstyle=\ttfamily\footnotesize,
    numbers=none,
    keywordstyle=\color{soromoxkeyword}\bfseries,
    commentstyle=\color{soromoxcomment}\itshape,
    stringstyle=\color{soromoxstring},
}

\begin{document}


\title{SoRoMoX: Fast, Differentiable, and Parallelizable Soft Robot Models}

\ifanonymoussubmission
\else
\author{{M}aximilian Stölzle, Solange Gribonval, Daniel Feliu-Talegon,\\ Vito Daniele Perfetta\textsuperscript{*}, Michele Martini\textsuperscript{*}, Chuhan Zhang\textsuperscript{*}, Kiwan Wong\textsuperscript{*},\\ Mohammed Tarnini, Anup Teejo Mathew, Federico Renda,\\ Daniela Rus, and Cosimo Della Santina%
\thanks{M. Stölzle, S. Gribonval, D. Feliu-Talegon, V. D. Perfetta, M. Martini, C. Zhang, and C. Della Santina are with Delft University of Technology, Delft, The Netherlands.}%
\thanks{M. Stölzle, K. Wong, and D. Rus are with the Massachusetts Institute of Technology, Cambridge, MA, USA.}%
\thanks{M. Martini is also with the Italian Institute of Technology, Genoa, Italy.}%
\thanks{M. Tarnini, A. T. Mathew, and F. Renda are with Khalifa University, Abu Dhabi, United Arab Emirates.}%
\thanks{\textsuperscript{*}V. D. Perfetta, M. Martini, C. Zhang, and K. Wong contributed equally.}%
}

\markboth{IEEE Robotics and Automation Magazine}{Stölzle \MakeLowercase{\textit{et al.}}: SoRoMoX}
\fi

\maketitle

\begin{abstract}
Reduced-order models based on Cosserat-rod theory are now well established, and modeling theory is no longer the primary bottleneck in soft-robot control. Their implementations, however, do not support the differentiable, GPU-parallel, and control-oriented workflows that underpin advanced rigid-robotics applications. Here, we fill this gap with \emph{SoRoMoX} (Soft Robot Models in JAX), a fully numerical, JIT-compilable Python/JAX framework. SoRoMoX implements articulated, Piecewise Constant Strain, and Variable Strain models through a unified, control-ready interface that provides inertia matrices, gravitational and elastic forces, Jacobians, and their derivatives. To our knowledge, it is the first rod/strain-based soft-robot modeling framework that runs directly on GPUs, with Warp kernels accelerating parallel continuum-model execution, and is end-to-end differentiable with respect to states, inputs, and parameters. Sequential CPU rollouts are up to $27.0\times$ faster than SoRoSim, while GPU-parallel GVS rollouts increase throughput by up to $679.7\times$. This performance enables workflows that were previously impractical or impossible: static-equilibrium system identification with 66\% lower marker RMSE; residual-force learning with a further 64\% reduction; computed-torque tracking with RMSE reduced by a factor of approximately $500$ relative to model-free PD; control-gain optimization with up to 98\% lower loss than untuned gains; safety-constrained control using high-order control barrier functions to keep the peak contact force within a prescribed \SI{5}{N} bound, compared with \SI{33.5}{N} without the safety constraint; and reinforcement-learning policy training up to $7\times$ faster than a CPU PyElastica discrete-rod baseline through massively parallel rollouts.
\end{abstract}


\section{Introduction}


\IEEEPARstart{S}{oft} robots derive physical intelligence from their compliance~\cite{milana2025physical}, but this same property complicates their control. Over the past decade, the soft robotics community has developed control-oriented formulations that reduce the theoretically infinite-dimensional deformation of soft bodies to tractable coordinates while retaining the dominant nonlinear, geometric, and material couplings~\cite{webster2010design, della2019control, grazioso2019geometrically, della2023model, caasenbrood2023control, alessi2024rod, mathew2025reduced}. We focus here on a class of models that is becoming a practical reference: Cosserat-rod models based on modal parametrizations of the strain field, including \ac{PCS}~\cite{renda2018discrete} and variable-strain formulations, particularly their \ac{GVS} implementation~\cite{della2019control, renda2020geometric,boyer2020dynamics,mathew2025reduced}.

The central limitation has therefore shifted from model derivation to model deployment. In rigid robotics, differentiable dynamics engines and accelerator-based parallelism have made the \ac{EOM} directly usable within large-scale optimization and learning pipelines, spanning system identification, trajectory optimization, reinforcement learning, and design~\cite{brax2021github,makoviychuk2isaac}. Comparable capabilities remain largely unavailable for soft robots. Existing implementations are often closed source or embedded in legacy computational ecosystems, do not differentiate end to end with respect to states, inputs, and parameters, cannot exploit modern accelerators, or expose only a subset of the quantities needed by model-based control algorithms~\cite{della2023model,mathew2024analytical,mathew2022sorosim}.

We address this limitation with \ac{SoRoMoX}, a JAX/Python framework that brings expressive soft-robot models into a modern computational stack~\cite{jax2018github}. SoRoMoX provides a common interface to control-relevant quantities, supports \ac{JIT} compilation and \ac{AD}, and evaluates single models or large batches on CPUs, GPUs, and TPUs. We use this infrastructure for gradient-based static-equilibrium system identification, residual-force learning, model-based control~\cite{della2023model, stolzle2025phdthesis}, control-gain optimization, massively parallel \ac{RL}, and safety-constrained control with \acp{HOCLF} and \acp{HOCBF}~\cite{wong2025contact}. These case studies show how computational tools already common in rigid robotics can be transferred to reduced-order soft-robot models.

In summary, this article makes the following contributions.
\begin{itemize}
\item SoRoMoX implements articulated, \ac{PCS}, and \ac{GVS} models through a numerical core that supports \ac{JIT} compilation, differentiation with respect to states, inputs, and parameters, and direct execution on GPUs, with Warp kernels accelerating parallel continuum-model execution.
\item A common interface provides kinematics, dynamics, energies, fused forward dynamics, operational- and actuation-space maps, and packaged model-based controllers across model families and actuation types.
\item Benchmarks show up to $27.0\times$ faster sequential CPU execution than SoRoSim~\cite{mathew2022sorosim}, while GPU-parallel GVS rollouts increase throughput by up to $679.7\times$. Six case studies cover static-equilibrium identification, residual learning, model-based control, control-gain optimization, safety-constrained control with \acp{HOCBF}~\cite{wong2025contact}, and parallel \ac{RL}.
\item Sidebars 1 to 4 collect the implementation practices that support these results, including a control-ready interface, structure-aware batched evaluation, singularity- and gradient-safe maps, and continuous-integration testing.
\end{itemize}

\paragraph*{Relation to existing tools}
SoRoMoX connects three families of existing software. JAX-based dynamics engines such as MuJoCo XLA (MJX)~\cite{todorov2012mujoco, mjx2023mujoco}, JaxSim~\cite{ferretti2026contact}, and Adam~\cite{adam2026}, together with GPU learning platforms such as Isaac Sim/Gym~\cite{makoviychuk2isaac} and Newton~\cite{newton}, provide contemporary execution models but primarily target rigid multibody systems. Particle- and MPM-based differentiable simulators such as ChainQueen~\cite{spielberg2023advanced} and SoftZoo~\cite{wang2023softzoo} resolve rich deformations but do not expose reduced-order dynamics in a form tailored to control. Rod-based packages are the closest alternatives. SoRoSim~\cite{mathew2022sorosim} implements \ac{GVS} models in MATLAB but is CPU-bound and not \ac{AD}-native\footnote{SoRoSim does not support analytical differentiation with respect to arbitrary quantities, such as robot parameters or control inputs, although it has recently introduced analytical derivatives with respect to system states~\cite{mathew2024analytical}.}, while PyElastica~\cite{naughton2021elastica} provides a fast CPU-based\footnote{A recent experimental repository includes a JAX implementation that enables GPU \& TPU parallelization.} Cosserat-rod solver without a control-oriented model interface. SoRoMoX combines the reduced-order structure of \ac{PCS} and \ac{GVS} models with \ac{AD}, \ac{JIT} compilation, and batched execution on modern accelerators.

\begin{table*}[t]
\centering
\scriptsize
\setlength{\tabcolsep}{1.5pt}
\renewcommand{\arraystretch}{1.12}
\caption{\textbf{Capability matrix for selected rod/strain-based soft robot model implementations.}
We use the following abbreviation: Automatic Differentiation (AD).
If a package supports a general strain model (e.g., \ac{VS}), we report
more specialized strain models (e.g., PCC, PCS) only if it contains a
dedicated implementation for those cases.}
\label{tab:rod_strain_capability_matrix}
\begin{tabularx}{\textwidth}{@{}
>{\raggedright\arraybackslash}p{0.11\textwidth}
>{\raggedright\arraybackslash}X
>{\raggedright\arraybackslash}X
>{\raggedright\arraybackslash}p{0.078\textwidth}
>{\centering\arraybackslash}p{0.050\textwidth}
>{\raggedright\arraybackslash}p{0.090\textwidth}
>{\centering\arraybackslash}p{0.052\textwidth}
>{\raggedright\arraybackslash}X
>{\centering\arraybackslash}p{0.058\textwidth}
@{}}
\toprule
\textbf{Implementation}
&
\textbf{Rod Models}
&
\textbf{Actuator Models}
&
\textbf{Interfaces}
&
\textbf{\makecell{Control\\API}}
&
\textbf{Compilation}
&
\textbf{Accel.}
&
\textbf{Differentiation}
&
\textbf{Docs}
\\
\midrule
SoMoGym~\cite{graule2022somogym}
&
Segmented rigid-link approximation
&
\xmark
&
Python; PyBullet; Gym
&
\xmark
&
PyBullet C++ backend
&
CPU
&
\xmark
&
Sparse
\\
SoftManiSim~\cite{kasaei2025softmanisim}
&
PyBullet PCC proxy
&
\xmark
&
Python; PyBullet; Gym
&
\xmark
&
PyBullet C++
&
CPU
&
\xmark
&
Sparse
\\
TMTDyn~\cite{sadati2021tmtdyn}
&
Lumped segmented Euler--Bernoulli beam
&
Fluidic pressure (equivalent chamber wrenches)
&
MATLAB
&
\xmark
&
MATLAB codegen to C-MEX
&
CPU
&
\xmark
&
Sparse
\\
PyElastica~\cite{naughton2021elastica}
&
Discrete Cosserat rods
&
Prescribed distributed muscle torques
&
Python; NumPy
&
\xmark
&
Numba JIT
&
CPU
&
\xmark
&
Excellent
\\
DisMech~\cite{choi2024dismech}
&
Discrete elastic rods
&
\xmark
&
C++; Python bindings
&
\xmark
&
C++ AOT
&
CPU
&
\xmark
&
Good
\\
Sorotoki~\cite{caasenbrood2024sorotoki}
&
PCC; PCS; VS
&
Threadlike actuation
&
MATLAB
&
\cmark
&
\xmark
&
CPU
&
\xmark
&
Good
\\
SoRoSim~\cite{mathew2022sorosim}
&
GVS
&
Threadlike actuation
&
MATLAB
&
\cmark
&
MATLAB codegen to C-MEX
&
CPU
&
Analytical state derivatives~\cite{mathew2024analytical}
&
Sparse
\\
SoRoMoX (ours)
&
PCC; PCS; GVS
&
Threadlike actuation; articulated tendon; McKibben; functional metamaterial
&
Python; JAX/Warp
&
\cmark
&
JAX/XLA JIT; Warp kernels
&
CPU; GPU; TPU
&
AD: states, inputs, parameters
&
Excellent
\\
\bottomrule
\end{tabularx}
\end{table*}

\paragraph*{Access to the tool}

\ifanonymoussubmission
\else
The \ac{SoRoMoX} package is available as open source under the MIT license on GitHub\footnote{\url{https://github.com/tud-phi/soromox}}, with documentation hosted on GitHub Pages\footnote{\url{https://tud-phi.github.io/soromox/}}.
\fi

\begin{figure}
\centering
\includegraphics[width=1.0\linewidth]{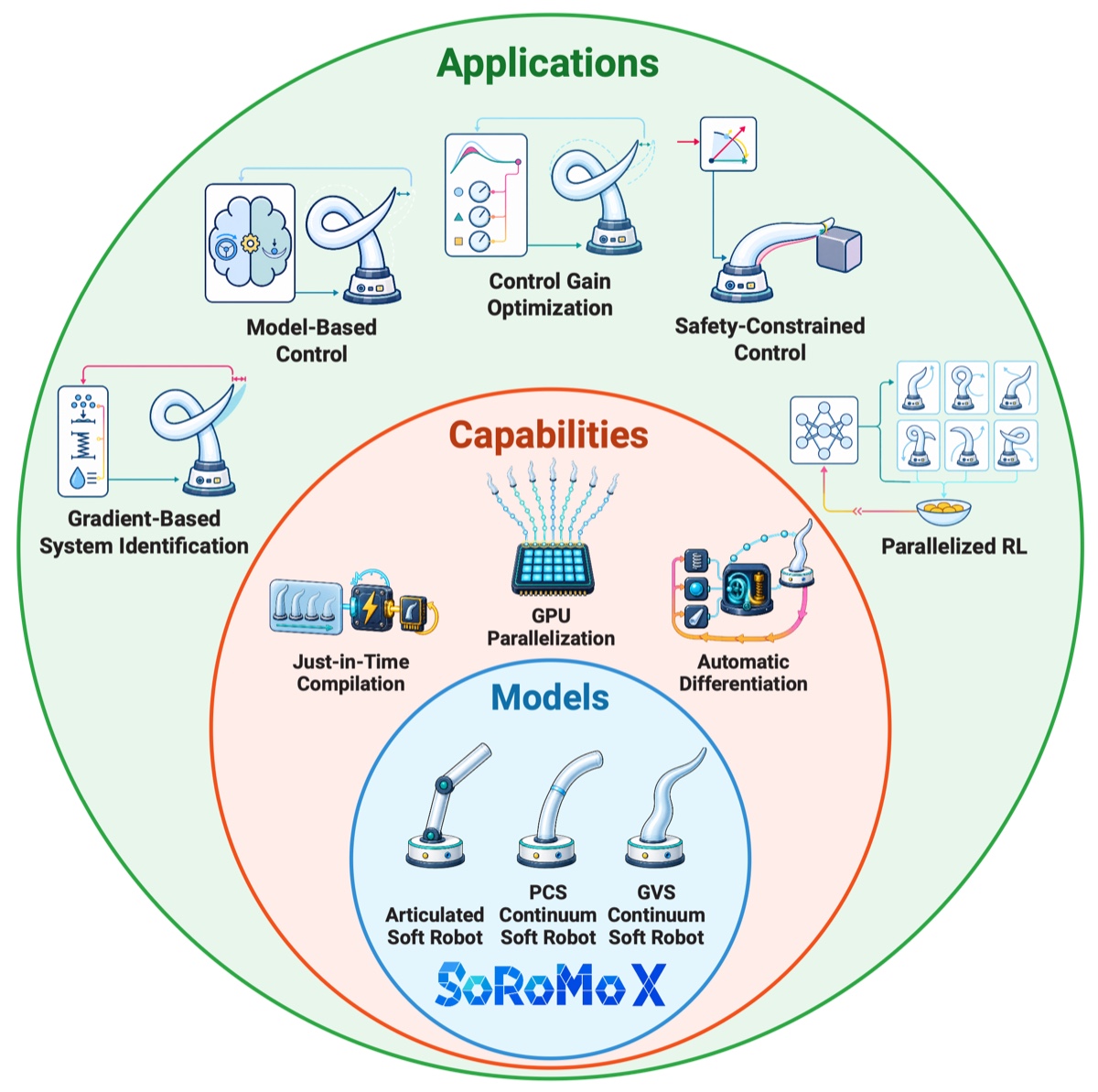}
\caption{Overview of SoRoMoX. Implemented in Python/JAX, the framework provides control-oriented models for soft robots, including articulated systems and continuum rod formulations such as \ac{PCS}~\cite{renda2018discrete} and \ac{GVS}~\cite{boyer2020dynamics, renda2020geometric}. \ac{JIT} compilation, automatic differentiation, and batched accelerator execution support gradient-based system identification, model-based control, control-gain optimization, parallel \ac{RL}, and \ac{CBF}-based safety-constrained control.}
\label{fig:overview}
\vspace{-0.1cm}
\end{figure}

\section{What SoRoMoX Provides}

\subsection{Model Families}
\ac{SoRoMoX} focuses on soft robot models that are rich enough to capture the dominant deformations of real systems, but compact enough to be useful for simulation, control, optimization, and learning. The package currently covers two broad families: articulated soft robots, which keep the serial-chain structure of classical robots while adding compliance, and continuum soft robots, whose shape is described by a small number of strain coordinates along a deformable backbone. Every soft robot model family supports an optional floating-base dynamics formulation.

Across these families, the models share a common control-oriented form. Rather than introducing the full derivation here, we use the following equation to fix the notation used throughout the article~\cite{della2023model, stolzle2025phdthesis}:
\begin{equation}
    M(q) \, \Ddot{q} + C(q,\dot{q}) \, \dot{q} + G(q) + K(q) + D \, \dot{q} = A(q) \, u.
    \label{eq:system_dynamics}
\end{equation}
Here, $q \in \mathbb{R}^n$ collects the model coordinates and $\dot{q}$ their velocities. The terms on the left describe inertia $M(q)$, Coriolis and Centrifugal effects $\tau_\mathrm{C} = C(q,\dot{q})\dot{q}$,\footnote{The factorization of the Coriolis term is not unique; the physically relevant quantity is the product $C(q,\dot{q})\dot{q}$.} gravity $G(q)$, elasticity $K(q)$, and damping $D\dot{q}$. The right-hand side maps actuator commands $u \in \mathbb{R}^m$ into generalized forces through the actuation matrix $A(q)$. We also write $\tau_\mathrm{C}=C(q,\dot{q})\dot{q}$ and $\tau_\mathcal{U}=G(q)+K(q)$ for the Coriolis and potential generalized-force terms.

The same notation carries over to kinematics for a robot with length $L$. A point at curvilinear abscissa $s \in (0, L]$ along the robot has pose $g(q,s) \in SE(3)$, and its spatial velocity is written as $\eta=J(q,s)\dot{q} \in \mathbb{R}^6$\footnote{This assumes a configuration-independent and time-invariant strain basis.}, where $J(q,s)$ is the Jacobian. When accelerations are needed, \ac{SoRoMoX} also exposes the Jacobian time derivative $\dot{J}(q,\dot{q},s)$. For compactness, we write these quantities in the spatial $SE(3)$ setting; the planar models use the analogous $SE(2)$ quantities.

\Cref{tab:implemented_models} summarizes the model families, actuation modalities, and representative robot-specific instantiations currently provided in \ac{SoRoMoX}.

\begin{table*}[t]
\centering
\caption{Soft robot model families implemented in \ac{SoRoMoX}.}
\label{tab:implemented_models}
\small
\setlength{\tabcolsep}{3pt}
\renewcommand{\arraystretch}{1.08}
\begin{tabular}{@{}>{\raggedright\arraybackslash}p{0.24\textwidth}>{\centering\arraybackslash}p{0.13\textwidth}>{\raggedright\arraybackslash}p{0.34\textwidth}>{\raggedright\arraybackslash}p{0.23\textwidth}@{}}
\toprule
\textbf{Model} & \textbf{Planar Implementation} & \textbf{Actuation Modalities} & \textbf{Example Instantiations} \\
\midrule
\multicolumn{4}{@{}l}{\textsc{Articulated Soft Robot Models}} \\
\cmidrule(lr){1-4}
Articulated Soft Robot~\cite{huang2020dynamic} &
\cmark\ (Soft Pendulum) &
Generalized-coordinate actuation (joint torques); articulated-tendon actuation; McKibben actuation &
UMArm~\cite{zuo2025umarm} \\
\addlinespace[0.8em]
\multicolumn{4}{@{}l}{\textsc{Continuum Soft Robot Models}} \\
\cmidrule(lr){1-4}
Piecewise Constant Strain (PCS)~\cite{renda2016discrete,renda2018discrete} &
\cmark\ (Planar PCS) &
Generalized-strain actuation; threadlike actuation; functional-metamaterial actuation (HSA) &
I-SUPPORT~\cite{arleo2021isupport}; planar HSA robot~\cite{stolzle2024experimental} \\
\addlinespace[0.45em]
Geometric Variable Strain (GVS)~\cite{renda2020geometric,boyer2020dynamics,mathew2025reduced} &
\xmark &
Generalized-coordinate actuation; threadlike actuation &
Tapered cable-driven soft tentacle~\cite{ramirez2024mastering} \\
\bottomrule
\end{tabular}
\end{table*}

\subsubsection{Articulated Soft Robot Models}
Articulated soft robots combine rigid or nearly rigid links with compliant joints, deformable transmissions, or soft actuators~\cite{della2020soft,huang2020dynamic}. In \ac{SoRoMoX}, this family is represented by \texttt{ArticulatedSoftRobot}, an open spatial serial chain whose coordinates are scalar joint variables. Each joint is described by a screw axis, fixed transforms define the parent-to-joint frames, and each link contributes inertia, gravity, and optional stiffness, damping, and rest-position terms.

This representation allows the same class structure to cover rigid serial mechanisms, soft-joint chains, tendon-actuated pendulums, and fluidic-pressure-actuated examples such as the McKibben-driven UMArm~\cite{zuo2025umarm}. For controller design, \ac{SoRoMoX} exposes the same quantities as in \eqref{eq:system_dynamics}: inertia, gravity, elastic and damping forces, actuation maps, and forward dynamics. Internally, these quantities are assembled from standard rigid-body kinematics and energy expressions~\cite{murray2017mathematical,hollerbach1980recursive,walker1982efficient}. When only accelerations are required, the implementation can use the articulated-body algorithm to avoid repeatedly solving dense equations of motion~\cite{featherstone1983calculation}.

\subsubsection{Continuum Soft Robot Models}
Continuum soft robots are described through reduced strain coordinates derived from Cosserat rod theory~\cite{renda2016discrete,renda2018discrete,renda2020geometric,boyer2020dynamics,mathew2025reduced}. Instead of representing the full elastic body with an infinite-dimensional field, the configuration $q$ captures a finite set of dominant bending, shear, torsional, and axial deformation modes. This gives a finite-dimensional model while preserving the mechanical structure needed for simulation, differentiation, and model-based control~\cite{armanini2023soft,della2023model}.

\ac{PCS} models use one particularly simple idea: within each segment, the strain is constant. This leads to compact and efficient models, with a spatial \texttt{PCS} class for $SE(3)$ motion and a \texttt{PlanarPCS} specialization for planar bending, axial, and shear deformation. Composable threadlike actuators reuse the same constant-strain backbone and provide actuator-specific maps $A(q)$ through presets for tendons, push rods, simplified muscles, and equivalent pressure chambers, including the fluidic-pressure-actuated spatial I-SUPPORT model~\cite{arleo2021isupport}.

\ac{VS} denotes the general class of strain-based continuum descriptions in which the strain field is allowed to vary along the backbone, rather than being constant within each segment~\cite{della2019control, della2023model, mathew2025reduced}.
Without a spatial reduction, this strain field is infinite-dimensional and the corresponding Cosserat-rod dynamics are governed by \acp{PDE}.
The \ac{GVS} formulation~\cite{renda2020geometric,boyer2020dynamics,mathew2025reduced} obtains a finite-dimensional reduced-order model by approximating the backbone strain with a prescribed finite set of basis functions, encoded by the strain basis.\footnote{Under a \ac{VS} approximation, the strain basis need not be fully spatially continuous; it can also be discontinuous or piecewise. Indeed, \ac{GVS} is implemented in both SoRoSim and \ac{SoRoMoX} as Piecewise Variable Strain.}
The generalized coordinates are then the time-varying coefficients of this strain parametrization, and the resulting dynamics take the \ac{ODE} form used in \eqref{eq:system_dynamics}.
In \ac{SoRoMoX}, each \ac{GVS} segment can define its own link properties, joint type, strain basis, active strain components, and quadrature resolution, while fixed-shape runtime arrays keep the implementation compatible with JAX compilation, differentiation, and batching.
\ac{GVS} systems support threadlike actuation through fixed material-frame path routing on top of the same segment representation.

Although \ac{PCS} models can be expressed as special cases of \ac{VS}, \ac{SoRoMoX} keeps dedicated planar and spatial \ac{PCS} implementations. The reason is practical: specialized models are easier to read, adapt, and test, and they can exploit closed-form constant-strain kinematics instead of numerical integration. The computational benefit of this specialization is reported in \Cref{sec:benchmarking}.

Additional continuum instantiations include the functional-metamaterial-driven planar \ac{HSA} robot~\cite{stolzle2023modelling,stolzle2024experimental} and the tapered cable-driven soft tentacle model used in contact-rich manipulation~\cite{ramirez2024mastering}. Detailed derivations are available in the original modeling papers and recent surveys~\cite{renda2018discrete,boyer2020dynamics,mathew2025reduced,armanini2023soft}.

\begin{figure*}
    \centering    
    \includegraphics[width=1\linewidth]{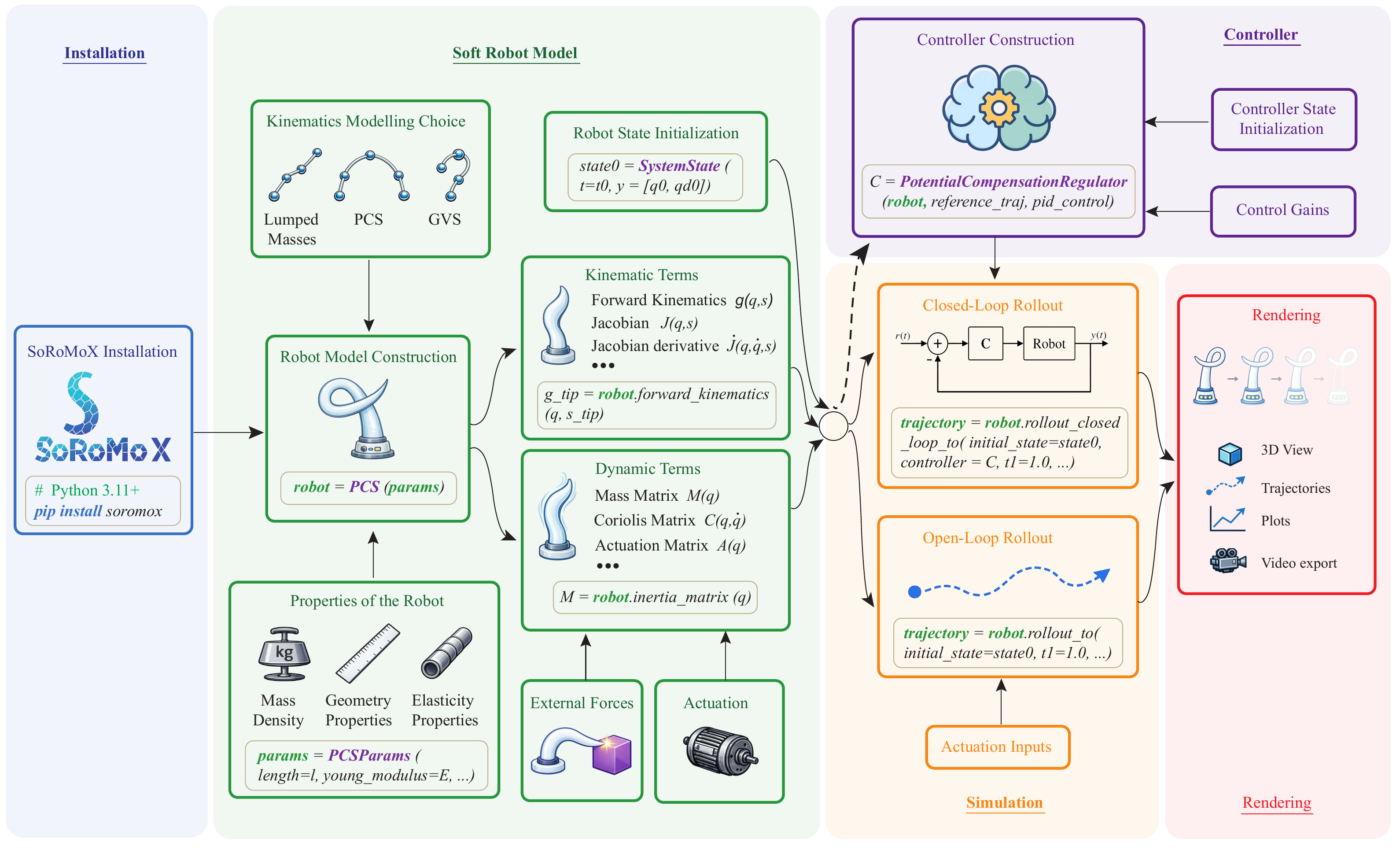}
    \caption{\textbf{SoRoMoX user workflow.} High-level fixed-base-default path from installation and robot definition to querying model terms, constructing model-based controllers, running open- and closed-loop simulations, and rendering resulting trajectories and shape evolutions.}\label{fig:soromox_user_workflow}
\end{figure*}

\subsection{A Control-Ready Model Interface}

The central software idea in \ac{SoRoMoX} is that a model should be more than a black-box simulator. A rollout trajectory is useful, but many downstream algorithms also need direct access to the structure of the model: controllers need gravity and elastic compensation, optimization routines need derivatives, safety filters need Jacobians and constraint sensitivities, and benchmarking needs comparable model quantities across formulations.

For this reason, the articulated, \ac{PCS}, and \ac{GVS} classes expose a common interface for forward kinematics $g(q,s)$, Jacobians $J(q,s)$, Jacobian derivatives $\dot{J}(q,\dot{q},s)$, energies $\mathcal{T}(q,\dot{q})$ and $\mathcal{U}(q)$, the dynamic terms in \eqref{eq:system_dynamics}, and fused forward dynamics. \ac{SoRoMoX} also provides transformations to operational-space dynamics for task-space objectives~\cite{khatib2003unified} and actuation-space dynamics for underactuated systems with integrable actuator-coordinate maps, including systems with threadlike actuation~\cite{pustina2024input}.

On the implementation side, robot parameters, system states, controller states, and optional environment states are represented as typed JAX/Equinox objects. This keeps the numerical data compatible with \ac{JIT} compilation, automatic differentiation, vectorized batch evaluation, and execution on CPU, GPU, or TPU backends~\cite{jax2018github}. Open-loop, continuous closed-loop, and sampled-data closed-loop rollouts are integrated with Diffrax~\cite{kidger2021on}. \Cref{sec:getting_started} turns this interface into concrete user-facing code examples, while the sidebars before the conclusion summarize the underlying interface and implementation choices.

\subsection{Execution Backends}\label{sub:execution_backends}

\ac{SoRoMoX} provides two execution backends. JAX offers strong CPU performance and end-to-end automatic differentiation across model families, whereas Warp accelerates parallel GPU rollouts of computationally complex \ac{PCS}- and \ac{GVS}-based continuum models~\cite{warp2022}. Current backend coverage is maintained in the online documentation.

\subsection{Model-Based Controllers}

\ac{SoRoMoX} implements model-based controllers around the same model interface used for simulation. This avoids maintaining separate robot descriptions for dynamics, control, and rollout evaluation. At a high level, each controller follows the map
\begin{equation}
    \pi_{\zeta^\mathrm{d}}: (t, y, z) \mapsto (u, \dot{z}),
    \qquad
    u = u_\mathrm{mb} + u_\mathrm{fb},
\end{equation}
where $\zeta=\psi(q)$ is the controlled output, $\zeta^\mathrm{d}$ is the desired reference, $y=[q^\top,v^\top]^\top$ is the robot state, $z$ is an optional controller state, $u_\mathrm{mb}$ is a model-based cancellation or feedforward term, and $u_\mathrm{fb}$ is the feedback term. This notation covers both stateless controllers and controllers with internal states, such as integral-action controllers.

References can be setpoints or time-varying trajectories. Depending on the controller, they may be expressed in configuration space, task space, or actuation space, with optional desired velocities and accelerations. \ac{SoRoMoX} supports both continuous-time closed-loop rollouts, where the controller is queried during numerical integration, and sampled-data rollouts, where the actuation is updated at a fixed control period and held constant between updates.

The implemented controller families are organized by the space in which the feedback law is defined. Configuration-space controllers include PID feedback, gravity and potential compensation, feedforward compensation, mixed-state feedback, and computed-torque tracking~\cite{della2020improved,della2020model,della2023model,stolzle2025phdthesis}. Actuation-space controllers support underactuated systems with integrable actuator-coordinate maps, including systems with threadlike actuation, through transformed collocated dynamics~\cite{pustina2024input,pustina2025analysis}. Operational-space controllers support task-space objectives such as impedance tracking and synergistic control~\cite{khatib2003unified,della2020model,stolzle2024guiding,della2019exact}. \Cref{tab:implemented_model_based_controllers} provides the detailed controller overview in the appendix.

\subsection{Extensibility}

The implementation is designed so that users can adapt existing model families to new robots without copying an entire system implementation. Robot-specific instantiations, such as the McKibben-actuated UMArm~\cite{zuo2025umarm} and the I-SUPPORT pneumatic arm~\cite{arleo2021isupport}, illustrate this pattern. A custom system can inherit from the common \texttt{SoftRobot} interface or from an existing concrete model, such as a \ac{PCS} or articulated-robot class, and override only the quantities that change.

In practice, these model-specific quantities often include geometry, cross-sectional area, second moments of area, material parameters, actuator routing, or the actuation matrix that maps actuator inputs to generalized forces. The separation between kinematics, dynamics, actuation, simulation, control, rendering, and differentiation also leaves room for larger extensions, including optimized specializations for particular kinematic parametrizations or open-chain kinematic trees. This extensibility is supported by explicit interfaces, extensive docstrings, and the common API contract summarized in Sidebar 1.

\subsection{Rendering and Visualization}

\ac{SoRoMoX} includes rendering tools because visual inspection is often the fastest way to understand whether a soft robot model, controller, or learned policy is behaving as intended. The renderers share a common abstraction: they sample forward kinematics along the robot backbone and expose common methods for rendering a single configuration, a full trajectory, or an interactive scene. This keeps visualization independent of whether the underlying robot is represented by a planar \ac{PCS}, spatial \ac{PCS}, or more general \ac{GVS} model~\cite{renda2018discrete,boyer2020dynamics,renda2020geometric}.

The four backends offer different display environments for robot configurations and trajectories. OpenCV draws planar robots on an image canvas for lightweight animations and video generation. Matplotlib provides planar and spatial plots with axes for figures and notebooks. Open3D renders spatial mesh scenes in a desktop viewer or offscreen for images and videos~\cite{zhou2018open3d}. Viser displays spatial mesh scenes in a web browser with interactive controls and live state streaming~\cite{yi2025viser}. Shared configuration covers camera, colors, geometry, scene appearance, and output settings, while editable presets give Open3D and Viser polished visual styles out of the box.
\Cref{sec:app:rendering} describes geometry reconstruction, backend selection, and visualization of robot motion.

\section{Getting Started and User Interface}
\label{sec:getting_started}

A central design goal of \ac{SoRoMoX} is to make control-oriented soft robot models directly usable from Python. The package exposes robot models as typed JAX/Equinox objects, with a common interface for kinematics, dynamics, simulation, control, rendering, and differentiation. The high-level workflow is summarized in \Cref{fig:soromox_user_workflow}, while this section turns the main steps into short user-facing code snippets; longer tutorials, complete scripts, and the full \ac{API} reference are provided in the online documentation\ifanonymoussubmission\else\footnote{\url{https://tud-phi.github.io/soromox/}}\fi.

\subsection{Installation and Running Your First Simulation}\label{sub:installation}

\Cref{fig:soromox_install} shows two common installation paths. The first installs the core package from PyPI, with a separate extra for Warp support. The second installs a source checkout with the dependencies needed to reproduce the examples. Current platform and version requirements are maintained in the online documentation.

\begin{figure}[t]
\ifanonymoussubmission
\else
\begin{lstlisting}[style=soromoxbash]
# Lightweight installation from PyPI.
pip install soromox

# Install Warp support.
pip install "soromox[warp]"

# Reproducible source checkout with example dependencies.
git clone https://github.com/tud-phi/soromox.git
cd soromox
uv sync --extra examples
\end{lstlisting}
\fi
\caption{\textbf{Installing SoRoMoX.} The PyPI commands install the core package and Warp support, while the \texttt{uv sync} workflow installs the source checkout with example dependencies.}
\label{fig:soromox_install}
\end{figure}

\Cref{fig:soromox_pcs_minimal} gives a compact first simulation using the spatial \ac{PCS} model. The example illustrates the three objects users encounter most often: per-link construction specifications, the robot object, and a \texttt{SystemState} object whose plant state is packed with the robot's state-layout helper. The construction factory converts the link specifications into the nested typed parameter tree used at runtime. The same robot object then exposes kinematic quantities, dynamic matrices, rollout methods, and renderer calls.

\begin{figure}[t]
\begin{lstlisting}[style=soromoxpython]
import jax.numpy as jnp
from soromox.rendering import ViserRenderer
from soromox.systems import LinkSpec, PCS, SystemState

# Two-segment spatial PCS rod with shared link specs.
num_segments = 2
ref_strain = jnp.array([0.0, 0.0, 0.0, 1.0, 0.0, 0.0])
links = [
    LinkSpec.circular(
        length=0.15,
        radius=0.01,
        density=1000.0,
        young_modulus=1e6,
        shear_modulus=1e5,
        material_damping_coefficient=362.0,  # Pa s
        reference_strain=ref_strain,
    )
    for _ in range(num_segments)
]

# Defaults: upright +z mounting and world -z gravity.
robot = PCS.from_links(links)
q = jnp.zeros(robot.num_coordinates)
v = jnp.zeros(robot.num_velocities)
s_tip = robot.segment_length.sum()
state0 = SystemState(t=0.0, y=robot.pack_state(q, v))

# Control-oriented model quantities at the tip.
g_tip = robot.forward_kinematics(q, s_tip)
J_tip = robot.jacobian(q, s_tip)
M = robot.inertia_matrix(q)
tau_U = robot.potential_force(q)
A = robot.actuation_matrix(q)
u0 = jnp.zeros(robot.num_actuators)

yd = robot.forward_dynamics(
    t=0.0,
    y=state0.y,
    actuation_args=(u0,),
)

# One second of open-loop dynamics with zero actuation.
trajectory = robot.rollout_to(
    initial_state=state0,
    u=u0,
    t1=1.0,
    solver_dt=1e-4,
    save_dt=1e-2,
)

# Render the saved configuration trajectory.
q_ts, v_ts, _ = robot.split_state(trajectory.y)
renderer = ViserRenderer(robot)
renderer.render_sequence(
    trajectory.t,
    q_ts,
    playback_speed=1.0,
    loop=True,
    autoplay=True,
)
\end{lstlisting}
\caption{\textbf{Minimal spatial PCS workflow.} Shared \texttt{LinkSpec} objects describe the physical links, and \texttt{PCS.from\_links} converts them into canonical runtime parameters. With no environment arguments, the backbone points along world $+z$ under gravity directed along world $-z$. The system class exposes kinematics and dynamics, \texttt{rollout\_to} integrates the dynamics using Diffrax~\cite{kidger2021on}, and the saved trajectory is passed to a renderer.}
\label{fig:soromox_pcs_minimal}
\end{figure}

\subsection{API Overview}\label{sub:api_overview}

Continuum systems use the link-oriented pattern shown in \Cref{fig:soromox_pcs_minimal}; a link is a soft continuum segment, not a rigid body. Factory methods convert \texttt{LinkSpec} objects and, for \ac{GVS}, joint and strain-basis specifications into typed parameter PyTrees and static structure. Isotropic materials may be given by Young's modulus $E$, shear modulus $G$ (or Poisson's ratio $\nu$), and material damping coefficient $\eta_\mathrm{mat}$; alternatively, users provide full, potentially coupled stiffness and damping matrices in generalized coordinates. \ac{GVS} shares \texttt{LinkSpec} with \ac{PCS}; separate \texttt{JointSpec} objects connect links (or the first link to the base), with fixed, revolute, prismatic, and spherical joints as examples. A \texttt{GVSSegment} groups each link with its preceding joint, strain basis, and quadrature resolution, as shown in \Cref{fig:soromox_GVS_construction}.

\begin{figure}[t]
\begin{lstlisting}[style=soromoxpython]
import jax.numpy as jnp
from soromox.systems import (
    GVS,
    GVSSegment,
    JointSpec,
    LinkSpec,
    StrainBasisSpec,
)

# A GVS segment bundles link data, joint, basis,
# and the quadrature rule used for integration.
link = LinkSpec.rectangular(
    length=0.3,
    height=0.03,
    width=0.02,
    density=1000.0,
    young_modulus=1e6,
    poisson_ratio=0.45,
    material_damping_coefficient=1e4,
    reference_strain=[0, 0, 0, 1, 0, 0],
)
segment = GVSSegment(
    link=link,
    joint=JointSpec.revolute(axis="z"),
    basis=StrainBasisSpec(
        type="monomial",
        strain_selector=(
            "kappa_x", "kappa_y", "kappa_z", "sigma_x"
        ),
        basis_order=1,
    ),
    num_gauss_points=7,
)

# Omitting environment arguments selects upright +z
# mounting and default world -z gravity.
# The factory separates static structure from parameters.
robot = GVS.from_segments(
    [segment],
)
\end{lstlisting}
\caption{\textbf{Constructing a GVS robot.} \ac{PCS} and \ac{GVS} share the same \texttt{LinkSpec} vocabulary; here, a rectangular cross-section illustrates that the shared construction layer supports geometries beyond the circular \ac{PCS} example. The illustrated \ac{GVS} segment adds a revolute joint about the $z$ axis, a strain basis, and integration settings in a \texttt{GVSSegment}. With no environment arguments, the robot points along the positive world $z$ axis under gravity directed along the negative world $z$ axis.}
\label{fig:soromox_GVS_construction}
\end{figure}

\Cref{tab:soromox_api} summarizes the main methods exposed by robot objects. To make the interface explicit, the table lists representative call signatures together with the corresponding returned quantity. Kinematic routines are available both pointwise and in batched form along the backbone. Dynamic routines expose the individual terms of the equations of motion as well as fused forward-dynamics calls for simulation.

\begin{table*}[t]
\centering
\caption{Select methods from the \ac{SoRoMoX} user/API interface. Here $q\in\mathbb{R}^{n_q}$, $v\in\mathbb{R}^{n_v}$, $u\in\mathbb{R}^m$, $s$ is a backbone abscissa, $y=[q^\top,v^\top]^\top$, and $n_\mathrm{samp}$ denotes the number of sampled backbone points.}
\label{tab:soromox_api}
\small
\setlength{\tabcolsep}{4pt}
\begin{tabular}{>{\raggedright\arraybackslash}p{0.17\textwidth}>{\scriptsize\raggedright\arraybackslash}p{0.31\textwidth}>{\raggedright\arraybackslash}p{0.21\textwidth}>{\raggedright\arraybackslash}p{0.23\textwidth}}
\toprule
\textbf{Task} & \textbf{Representative call} & \textbf{Returns} & \textbf{Purpose} \\
\midrule
Kinematics &
\texttt{forward\_kinematics(q, s)}\newline
\texttt{forward\_kinematics\_abscissa\_batched(q, s\_ps)} &
Pose $g(q,s)$; batched shape $(n_\mathrm{samp},\cdot)$ &
Evaluate poses at arbitrary backbone points or sampled point sets. \\
Jacobians &
\texttt{jacobian(q, s)}\newline
\texttt{jacobian\_and\_time\_derivative(q, v, s)} &
$J(q,s)$ and optionally $\dot{J}(q,v,s)$ &
Map generalized velocities to inertial-frame twists. \\
Dynamical matrices &
\texttt{inertia\_matrix(q)}\newline
\texttt{coriolis\_matrix(q, v)}\newline
\texttt{damping\_matrix(q)} &
$M(q)$, $C(q,\dot{q})$, $D$ &
Expose matrix terms of the Euler-Lagrange equations. \\
Forces &
\texttt{gravitational\_force(q)}\newline
\texttt{elastic\_force(q)}\newline
\texttt{potential\_force(q)} &
$G(q)$, $K(q)$, $\tau_\mathcal{U}(q)$ &
Return conservative generalized forces. \\
Actuation &
\texttt{actuation\_matrix(q)}\newline
\texttt{actuation\_force(q, u)} &
$A(q)$ and $A(q)u$ &
Map actuator commands to generalized forces. \\
Open-loop simulation &
\texttt{forward\_dynamics(t, y, (u,))}\newline
\texttt{rollout\_to(state0, u, t1)} &
$\dot{y}$ or a trajectory \texttt{SystemState} &
Evaluate state derivatives and integrate open-loop dynamics. \\
Closed-loop simulation &
\texttt{rollout\_closed\_loop\_to(...)}\newline
\texttt{rollout\_discrete\_closed\_loop\_to(...)} &
Trajectory \texttt{SystemState} &
Integrate with continuous or sampled controller calls. \\
Parameter updates &
\texttt{update\_params(gravity=...)}\newline
\texttt{update\_link\_params(density=...)}\newline
\texttt{with\_isotropic\_material(material)} &
Updated robot PyTree &
Replace top-level or link-local values, or map material variables to canonical matrices. \\
\bottomrule
\end{tabular}
\end{table*}

For control design, \ac{SoRoMoX} additionally provides coordinate-transform utilities. \texttt{OperationalSpaceDynamics} projects the model to selected task-space points and exposes task-space inertia, Coriolis terms, forces, Jacobians, and dynamically consistent pseudoinverses. \texttt{ActuationSpaceDynamics} transforms systems with an integrable actuator-coordinate map into coordinates in which the actuation matrix has the collocated form $[I,0]^\top$, which is useful for underactuated soft robots. \Cref{fig:soromox_control_interface} shows how these transformations connect to packaged controllers and closed-loop rollouts; \Cref{sec:app:api_interfaces} specifies the underlying abstractions and interface contracts. In the example, \texttt{routing} denotes a previously configured \texttt{ThreadlikeRouting} object.

\begin{figure}[t]
\begin{lstlisting}[style=soromoxpython]
import jax.numpy as jnp
from soromox.actuation import ThreadlikeActuator
from soromox.control import (
    ConfigurationSpacePIDController,
    PIDControl,
    ReferenceTrajectory,
)
from soromox.coordinate_transformations import (
    ActuationSpaceDynamics,
    OperationalSpaceDynamics,
)

# Operational-space dynamics for a tip-position task.
osd = OperationalSpaceDynamics(
    robot=robot,
    s_ps=jnp.array([s_tip]),
    task_selector=jnp.array(
        [False, False, False, True, True, True]
    ),
)
Mx = osd.inertia_matrix(q)
cx = osd.coriolis_force(q, v)
tau_Ux = osd.gravitational_force(q) + osd.elastic_force(q)

# Actuation-space dynamics for a threadlike-actuated system.
threadlike_robot = PCS.from_links(
    links,
    actuators=ThreadlikeActuator.tendons(routing),
)
asd = ActuationSpaceDynamics(robot=threadlike_robot)
q_act = jnp.zeros(threadlike_robot.num_coordinates)
M_phi = asd.inertia_matrix(q_act)
tau_U_phi = asd.potential_force(q_act)
A_phi = asd.actuation_matrix(q_act)

# Packaged configuration-space PID controller.
ts = jnp.array([0.0, 1.0])
q_des_ts = jnp.stack([q, q])
v_des_ts = jnp.stack([v, v])
ref_traj = ReferenceTrajectory(
    ts=ts,
    x_des_ts=q_des_ts,
    xd_des_ts=v_des_ts,
)
pid = PIDControl(Kp=20.0, Ki=0.0, Kd=2.0)
builtin_controller = ConfigurationSpacePIDController(
    robot=robot,
    reference_trajectory=ref_traj,
    pid_control=pid,
)

# Rollouts call the controller during integration.
closed_loop = robot.rollout_closed_loop_to(
    initial_state=state0,
    controller=builtin_controller,
    t1=1.0,
    solver_dt=1e-4,
    save_dt=1e-2,
)
\end{lstlisting}
\caption{\textbf{Coordinate transformations and controllers.} Task-space and actuation-space dynamics expose projected model quantities, while packaged controllers use the same interface for closed-loop simulation.}
\label{fig:soromox_control_interface}
\end{figure}

Simulation is handled through Diffrax~\cite{kidger2021on}. The default solver is Tsitouras’ 5/4 (\texttt{Tsit5})~\cite{tsitouras2011runge}, but any compatible Diffrax solver and step-size controller can be passed to the rollout methods. The solver step size, controller sampling time, and saved output time grid are configured separately through arguments such as \texttt{solver\_dt}, \texttt{control\_dt}, \texttt{save\_dt}, and \texttt{save\_ts}.

\Cref{fig:soromox_jax_rendering} shows how the same interface connects to JAX transformations and gradient-based parameter optimization. Since the core methods are JAX-compatible, users can compile model evaluations with \texttt{jax.jit}, vectorize them with \texttt{jax.vmap}, and differentiate objectives with \texttt{jax.grad}, \texttt{jax.jacfwd}, or \texttt{jax.jacrev}. This enables differentiable workflows for parameter identification and material optimization.

\begin{figure}[t]
\begin{lstlisting}[style=soromoxpython]
import jax
from soromox.systems import IsotropicMaterialParams

# JIT caches optimized kernels after the first call.
fast_fk = jax.jit(robot.forward_kinematics)

# Vectorize forward dynamics over a leading batch axis.
n_b = 2
# y_batch shape: (n_b, state_size).
y_batch = jnp.repeat(state0.y[None, :], n_b, axis=0)
# u_batch shape: (n_b, num_actuators).
u_batch = jnp.zeros((n_b, robot.num_actuators))
yd_batch = jax.vmap(
    lambda y_i, u_i: robot.forward_dynamics(
        0.0, y_i, (u_i,)
    )
)(y_batch, u_batch)

# A loss can depend on kinematics, dynamics, or a rollout.
# measured_tip_positions has shape (T, 3).
def loss_for_robot(candidate_robot):
    traj = candidate_robot.rollout_to(
        initial_state=state0,
        u=u0,
        t1=1.0,
        solver_dt=1e-4,
        save_dt=1e-2,
    )
    q_pred_ts, _, _ = candidate_robot.split_state(traj.y)
    tip_pos_ts = jax.vmap(
        lambda q_i: candidate_robot.forward_kinematics(
            q_i, s_tip
        )[:3, 3]
    )(q_pred_ts)
    error = tip_pos_ts - measured_tip_positions
    return jnp.mean(error**2)

# Optimize only Young's modulus in log space.
material = IsotropicMaterialParams(
    young_modulus=jnp.array(1e6),
    poisson_ratio=jnp.array(0.45),
    material_damping_coefficient=jnp.array(362.0),
)
log_E = jnp.log(material.young_modulus)

def loss_for_log_E(candidate_log_E):
    candidate_material = material.replace(
        young_modulus=jnp.exp(candidate_log_E)
    )
    candidate_robot = robot.with_isotropic_material(
        candidate_material
    )
    return loss_for_robot(candidate_robot)

learning_rate = 1e-3
value_and_grad = jax.value_and_grad(loss_for_log_E)
for _ in range(100):
    _, grad = value_and_grad(log_E)
    log_E = log_E - learning_rate * grad
optimized_material = material.replace(
    young_modulus=jnp.exp(log_E)
)
optimized_robot = robot.with_isotropic_material(
    optimized_material
)
\end{lstlisting}
\caption{\textbf{JAX transformations and parameter optimization.} SoRoMoX methods can be compiled, vectorized, and differentiated. Here, a caller-owned \texttt{IsotropicMaterialParams} keeps the material description interpretable while the positive Young's modulus is optimized in log space and mapped to canonical per-link stiffness and damping matrices without changing the model structure.}
\label{fig:soromox_jax_rendering}
\end{figure}

\section{Computational Performance Evaluation}\label{sec:benchmarking}

\paragraph{Sequential Rollouts on the CPU}
We evaluate the accuracy and computational efficiency of SoRoMoX against two CPU-based libraries for simulating flexible rods: PyElastica~\cite{naughton2021elastica}, which implements a \ac{DCM}~\cite{gazzola2018forward}, and SoRoSim~\cite{mathew2022sorosim}, which implements reduced-order strain-based models.

Across four cases, we simulate \SI{3}{s} of soft-arm motion using equivalent physical parameters in all three simulators. The hardware, model parameters, and benchmark protocol are described in \Cref{sec:app:cpu_benchmark}. 

\begin{figure*}
    \centering    \includegraphics[width=0.9\linewidth]{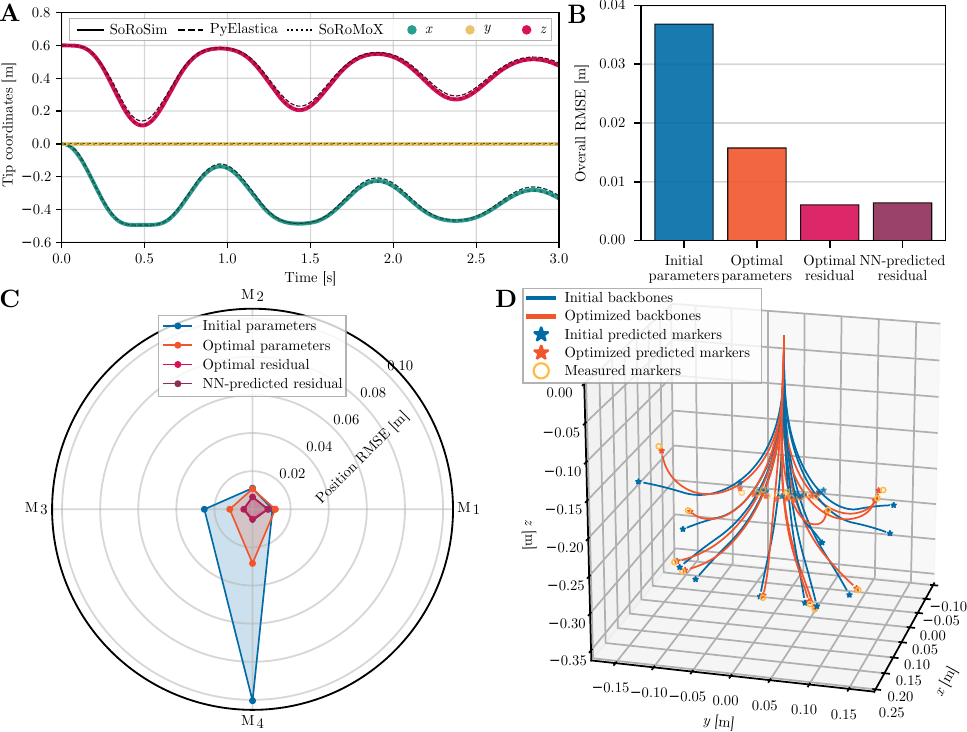}
    \caption{\textbf{Benchmarking Comparison and Static-Equilibrium Identification/Residual-Learning Case Study.} (\textbf{A}) CPU rollout benchmark of SoRoMoX against SoRoSim and PyElastica, showing the simulated tip response of a cylindrical soft beam initially aligned with the global $z$ axis and deflected in the $x$-$z$ plane by a distributed load of gravity-equivalent magnitude along the global $x$ direction. (\textbf{B}) Static-equilibrium system-identification and residual-learning application case study: overall soft robot model accuracy across four estimation procedures, quantified as the aggregate RMSE computed over all marker positions for pre-optimized physical parameters (blue), optimized physical parameters (orange), optimal residual generalized-force estimation (magenta), and NN-predicted residual generalized force (dark red).
    (\textbf{C}) Per-marker breakdown of the same accuracy evaluation, showing the individual position RMSE for each of the four markers across the four estimation procedures. (\textbf{D}) Backbone configurations from the same application case study, comparing pre-optimized physical parameters (blue lines) with optimal residual generalized-force estimation (orange lines).}\label{fig:benchmark_sysid}
\end{figure*}

\Cref{fig:benchmark_sysid}.A shows the tip response for a representative beam-deflection case, and \Cref{tab:comparison_speed} reports the runtimes. The trajectory comparison reported in \Cref{tab:app:rollout_agreement} gives SoRoSim--SoRoMoX tip-position RMSE values of \SIrange{0.86}{7.21}{mm}, corresponding to \SIrange{0.14}{1.20}{\percent} of the soft robot length. Across all four cases, SoRoMoX requires less computation time than SoRoSim, with speedups of approximately \(27.0\times\) for planar PCS, \(12.0\times\) for spatial PCS, \(5.2\times\) for spatial \ac{GVS}, and \(6.5\times\) for tendon-driven \ac{GVS}.



\begin{table}[t]
    \centering
    \caption{Computation time $\downarrow$ (s) required to simulate
    3\,s of motion using the Dormand–Prince \texttt{ode45} solver in SoRoSim, the Tsitouras \texttt{Tsit5} solver in SoRoMox and the second-order symplectic \texttt{PositionVerlet} solver in PyElastica.}
    \label{tab:comparison_speed}
    \begingroup
    \small
    \setlength{\tabcolsep}{5pt}
    \begin{tabular}{@{}llrrr@{}}
        \toprule
        Model & Case
        & Elastica~\cite{naughton2021elastica}
        & SoRoSim~\cite{mathew2022sorosim}
        & SoRoMoX \\
        \midrule
        \multirow{2}{*}{FEM/PCS}
        & Planar        & 2.25 & 75.73 & 2.81  \\
        & Spatial       & 2.45 & 78.65 & 6.53 \\
        \midrule
        \multirow{2}{*}{FEM/GVS}
        & Spatial       & --   & 55.54 & 10.63 \\
        & Tendons & 11.40 & 75.80 & 11.72 \\
        \bottomrule
    \end{tabular}
    \endgroup
\end{table}

\paragraph{Parallel Rollouts on the GPU}
As discussed in the introduction, parallel, batched rollouts of the (soft) robot models on the GPU are nowadays crucial for efficient training of RL policies~\cite{rudin2022learning}, batched optimization with differentiable simulation~\cite{spielberg2023advanced, du2021underwater, wang2023softzoo, wang2023diffusebot}, sampling-based planning methods (e.g., \ac{MPPI}~\cite{williams2016aggressive, pezzato2025sampling}), and related tasks.
To the best of our knowledge, \ac{SoRoMoX} is the first rod/strain-based simulator that can be readily deployed on modern GPUs. This deployment is straightforward in our package: a JAX vectorization operator (\texttt{jax.vmap}) can be wrapped around the simulation method (e.g., \texttt{robot.rollout\_to}) to evaluate $n_\mathrm{b}$ independent systems along a leading batch axis on the GPU, where the batch entries may correspond to robots, environments, trajectories, or controller initializations.
The key variable is the leading batch size $n_\mathrm{b}$: ideally, the total throughput scales linearly with $n_\mathrm{b}$ until the GPU capacity is reached. In practice, however, such linear scaling is only an upper theoretical bound, and the marginal performance gains from parallelization decrease as $n_\mathrm{b}$ grows. We quantify this scaling using the simulation-throughput factor $\Gamma_\mathrm{sim}$, defined as the total simulated time across all batched rollouts divided by the measured wall-clock runtime. Here, $\Gamma_\mathrm{sim}=1$ denotes parity between aggregate simulated time and wall-clock time. The full experimental protocol is given in \Cref{sec:app:gpu_batch_benchmark}.

\begin{figure}
    \centering
    \includegraphics[width=1.0\linewidth, trim={5 5 5 5}]{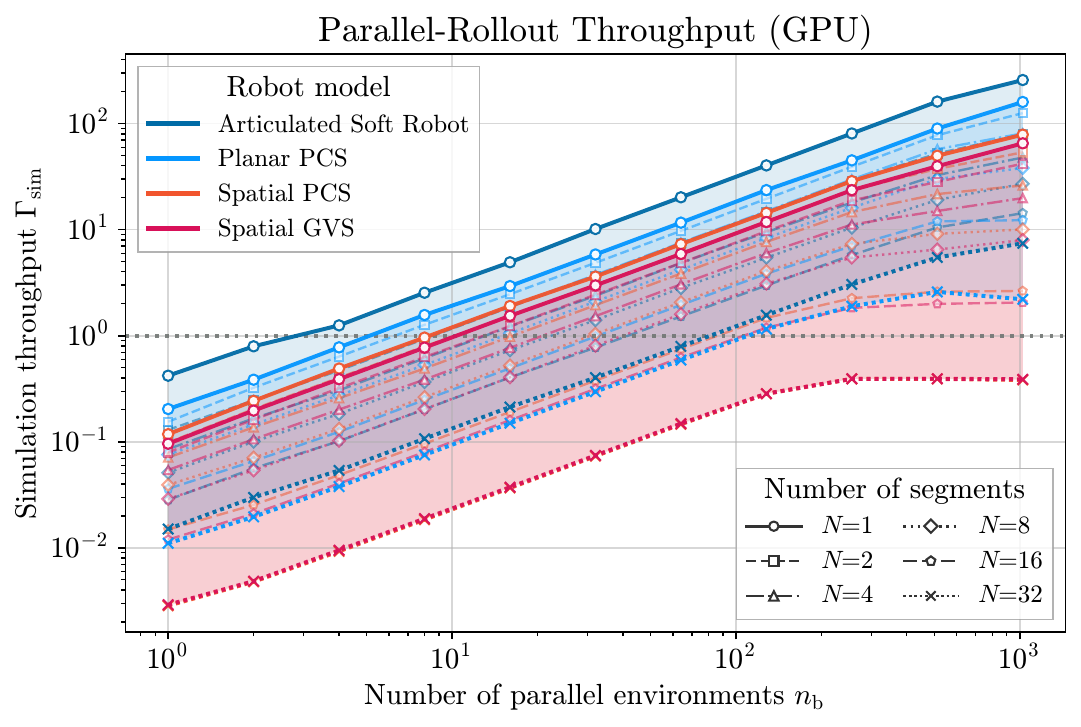}
    \caption{
        Benchmark of batched RTX 5090 GPU simulation throughput for $n_\mathrm{b}\in\{1,2,4,8,16,32,64,128,256,512,1024\}$ parallel environments and $N\in\{1,2,4,8,16,32\}$ robot links or segments. The simulation-throughput factor $\Gamma_\mathrm{sim}$ is the total simulated time across the batch divided by the wall-clock runtime. Full experimental details are given in \Cref{sec:app:gpu_batch_benchmark}.
    }\label{fig:benchmarking:batch_simulation_scaling}
\end{figure}

The GPU measurements in \Cref{fig:benchmarking:batch_simulation_scaling} demonstrate substantial throughput gains across all tested models and sizes. As discussed in \Cref{sub:execution_backends}, the \ac{PCS}- and \ac{GVS}-based systems leverage fused Warp kernels for parallel GPU execution. At \(n_\mathrm{b}=1024\), \ac{GVS} achieves \(679.7\times\) the single-environment throughput for \(N=1\) and \(133.6\times\) for \(N=32\). Aggregate throughput increases monotonically through \(n_\mathrm{b}=1024\) for the \ac{GVS} cases with \(N\in\{1,2,4,8,16\}\); the \(N=32\) case peaks at \(n_\mathrm{b}=512\) and is \(1.2\%\) lower at \(n_\mathrm{b}=1024\). At the largest tested batch size, the aggregate simulated-time throughput of all tested articulated and planar PCS configurations, including \(N=32\), meets or exceeds the wall-clock rate (\(\Gamma_\mathrm{sim}\geq1\)). The same holds for the tested spatial PCS and \ac{GVS} configurations with \(N\in\{1,2,4,8,16\}\).

\section{Application Case Studies}\label{sec:application_case_studies}
\begin{figure}
    \centering
    \includegraphics[width=1.0\linewidth]{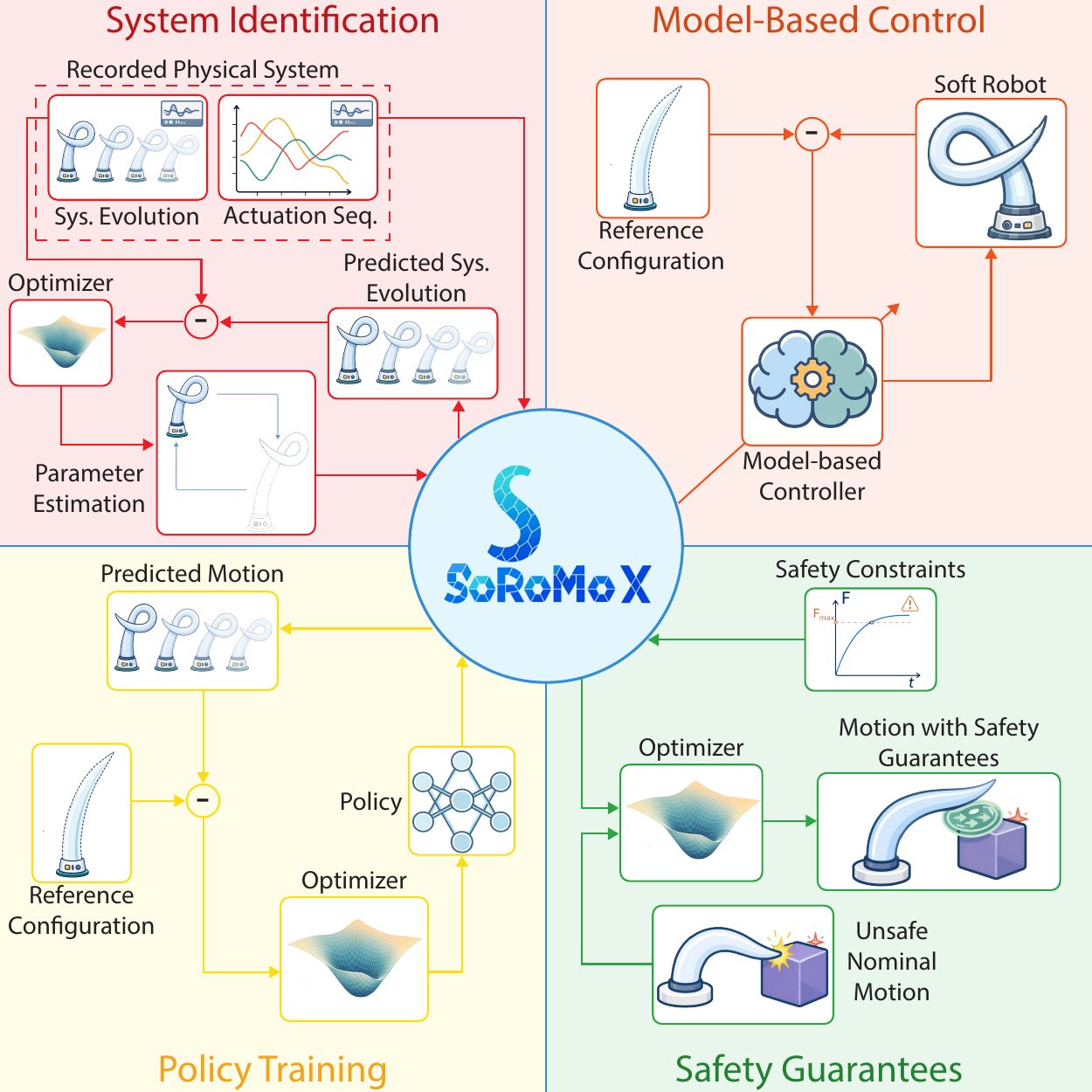}
    \caption{Application case studies for SoRoMoX: physical parameter identification and residual learning, model-based control and gain optimization, policy training, and safety-constrained control.}
    \label{fig:case_studies_blockscheme}
    \vspace{-0.1cm}
\end{figure}

This section summarizes downstream case studies that use SoRoMoX's control-oriented model interface together with its support for automatic differentiation and parallelization for different tasks: identifying physical parameters from static-equilibrium data, learning residual forces, deriving model-based controllers, optimizing controller gains, enforcing contact-aware control constraints, and training policies with parallel rollouts, as illustrated in \Cref{fig:case_studies_blockscheme}. The technical definitions of the robots, objectives, and controllers are given in \Cref{sec:usecases}.

\subsection{Static-Equilibrium System Identification}
Here, the Young's modulus, Poisson's ratio, and mass density of a conical tendon-driven continuum arm are identified from four-marker measurements collected at static equilibrium. Other geometric parameters, such as length, cross-section dimensions, and tendon routing, are measured directly from the hardware.

Efficient parameter identification is difficult for nonlinear soft robot dynamics, especially for full dynamical evolutions where closed-form linear least-squares formulations are generally unavailable. The same issue already appears in the nonlinear static-equilibrium fitting used here: each parameter update changes the elastic and nonlinear gravitational terms and therefore the predicted marker positions. \ac{SoRoMoX} addresses this by providing differentiable \ac{GVS} models, forward kinematics along the entire robot, static-equilibrium evaluation, and derivatives with respect to physical parameters within the same computation.

The arm is modeled with two \ac{GVS} links: the first uses a 7-\ac{DOF} strain basis with linear bending, constant torsion, and constant axial deformation, while the second uses a 2-\ac{DOF} constant-bending basis, yielding a 9-\ac{DOF} model. Quantitatively, the optimizer identifies $\theta^\star=[\SI{0.315}{MPa},\,0.45,\,\SI{1320}{kg \per m^3}]$ and reduces the overall static-equilibrium marker \ac{RMSE} from \SI{56.9}{mm} to \SI{19.3}{mm}, an improvement of about \SI{66}{\percent}, as shown in \Cref{fig:benchmark_sysid}B.

\subsection{Residual Learning}
Residual learning corrects model mismatch that remains after static-equilibrium system identification by adding input-dependent residual generalized forces. In this case study, these forces are fitted from equilibrium data and account for repeatable effects not captured by the nominal model~\cite{gao2024sim}.

The central algorithmic challenge is the efficient generation of residual-force targets. Each target is the solution of an optimization problem through the simulator or system model; without differentiability, generating such targets across many inputs becomes computationally expensive. \ac{SoRoMoX} supports differentiable static-equilibrium evaluation, allowing residual forces to be optimized through the same robot model used for marker prediction.

For each tendon input, we estimate an optimal residual generalized force and then use these optimized residuals as training targets for a neural network that maps tendon forces to residual-force predictions. Quantitatively, \Cref{fig:benchmark_sysid}B--D shows that the optimized residual and the neural-network residual reduce the overall static-equilibrium marker \ac{RMSE} to \SI{6.6}{mm} and \SI{6.9}{mm}, respectively, corresponding to improvements of about \SI{66}{\percent} and \SI{64}{\percent} relative to the model identified at static equilibrium.

\subsection{Model-Based Control}
Model-based control uses robot dynamics and kinematics to synthesize feedback laws for regulation and tracking, rather than treating the simulator as a black-box rollout engine. In this case study, we evaluate whether the same \ac{PCS} model implementation can support both strain-coordinate / shape control and operational-space pose tracking for fully actuated soft robots.

The central implementation challenge is that such controllers require many coupled model quantities to be available with consistent coordinates and conventions: inertial, Coriolis, gravitational, elastic, and damping terms, actuation maps, forward kinematics, Jacobians, and Jacobian derivatives~\cite{della2023model,pustina2024input}. \ac{SoRoMoX} exposes these quantities through one control-oriented API and provides closed-loop rollout utilities, allowing controller derivation, implementation, and simulation to share the same robot model.

\begin{figure*}[t]
    \centering
    \includegraphics[width=0.99\linewidth]{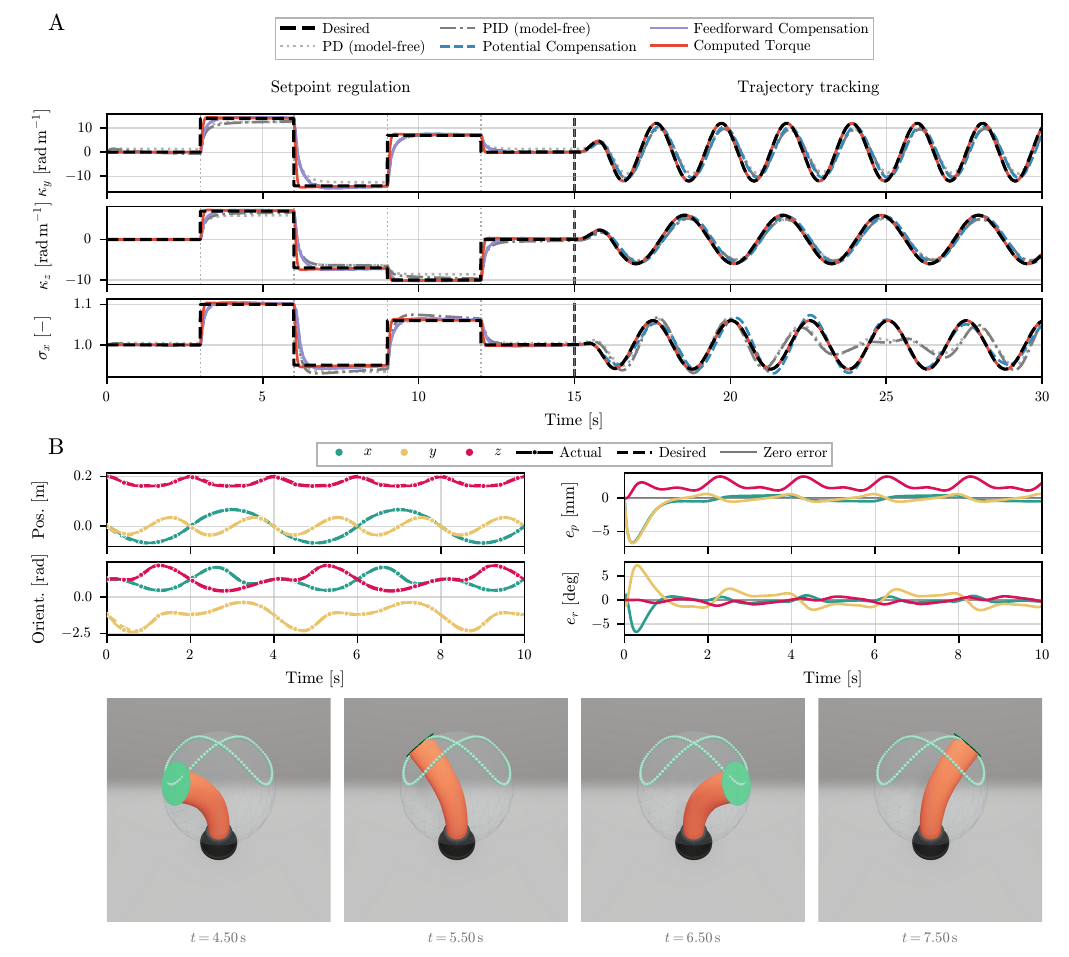}
    \caption{Model-based control with fully actuated spatial \ac{PCS} robots.
    (\textbf{A}) Configuration-space response of a one-segment robot during stepwise setpoint regulation ($0$--\SI{15}{s}) and trajectory tracking (\SI{15}{s}--\SI{30}{s}). The desired and actual commanded strains $\kappa_\mathrm{y}$, $\kappa_\mathrm{z}$, and $\sigma_\mathrm{x}$ are shown for model-free PD and PID control, potential compensation, feedforward compensation, and computed-torque control.
    (\textbf{B}) Operational-space impedance control of a two-segment robot tracking a spherical figure-eight pose trajectory. The plots compare actual and desired end-effector position and orientation coordinates and show their componentwise errors. In the snapshots, the coral body and dark base show the robot configuration, the translucent gray mesh is the target sphere, and the light-green beads trace the desired tip path. The green disk denotes the instantaneous target pose: its center gives the desired position, while its normal and crosshair give the outward-normal and tangent-frame axes, respectively.}
    \label{fig:model_based_control}
\end{figure*}

We evaluate two spatial \ac{PCS} examples based on the definitions in \Cref{sec:app:model_based_control_configuration,sec:app:model_based_control_operational}. First, a fully actuated one-segment robot is used to compare five configuration-space controllers under the same robot, reference, gains, and rollout: model-free proportional--derivative (PD), model-free proportional--integral--derivative (PID), potential compensation (PC), feedforward compensation (FF), and computed torque (CT). These controllers progressively add model terms, from no model compensation in PD/PID to static gravitational and elastic compensation, desired-trajectory inverse dynamics, and measured-state feedback linearization. Second, an operational-space impedance controller tracks a spherical figure-eight end-effector pose trajectory with a fully actuated two-segment robot. \Cref{fig:model_based_control} summarizes both closed-loop demonstrations. Quantitatively, exploiting model terms substantially improves performance over the model-free baselines: in terminal regulation, potential compensation reduces the errors in all six coordinates relative to both PD and PID, with coordinate-wise reductions across these comparisons ranging from \SI{43}{\percent} to \SI{98}{\percent}. During trajectory tracking, feedforward compensation reduces the \acp{RMSE} of all six coordinates by \SIrange{86.3}{99.7}{\percent} relative to potential compensation. Computed torque achieves the lowest \ac{RMSE} in every commanded and uncommanded generalized strain coordinate; during trajectory tracking, it further reduces the commanded-coordinate errors of $\kappa_\mathrm{y}$, $\kappa_\mathrm{z}$, and $\Delta\sigma_\mathrm{x}$ by \SI{62}{\percent}, \SI{79}{\percent}, and \SI{79}{\percent}, respectively, relative to feedforward compensation, as detailed in \Cref{tab:app:configuration_rmse}. For operational-space tracking, the Euclidean position-error norm has an \ac{RMSE} of \SI{2.84}{mm}, or \SI{2.11}{\percent} of the desired-position span, while the geodesic orientation-error norm has an \ac{RMSE} of \SI{2.29}{\degree}, or \SI{2.00}{\percent} of the desired-orientation span.

\subsection{Control Gain Optimization}
Control gain optimization selects controller parameters for desired soft robot closed-loop behavior. Manual gain tuning is time-consuming, inconsistent, and often leads to suboptimal transient responses.

The algorithmic difficulty is that the closed-loop response depends on the nonlinear coupling between the controller and the soft robot model. As a result, gradients of closed-loop performance metrics with respect to the gains are usually not available in closed form. \ac{SoRoMoX} makes this optimization tractable by differentiating through full closed-loop rollouts and by evaluating multiple gain initializations in parallel.

\begin{figure*}
    \centering
    \includegraphics[width=0.99\linewidth]{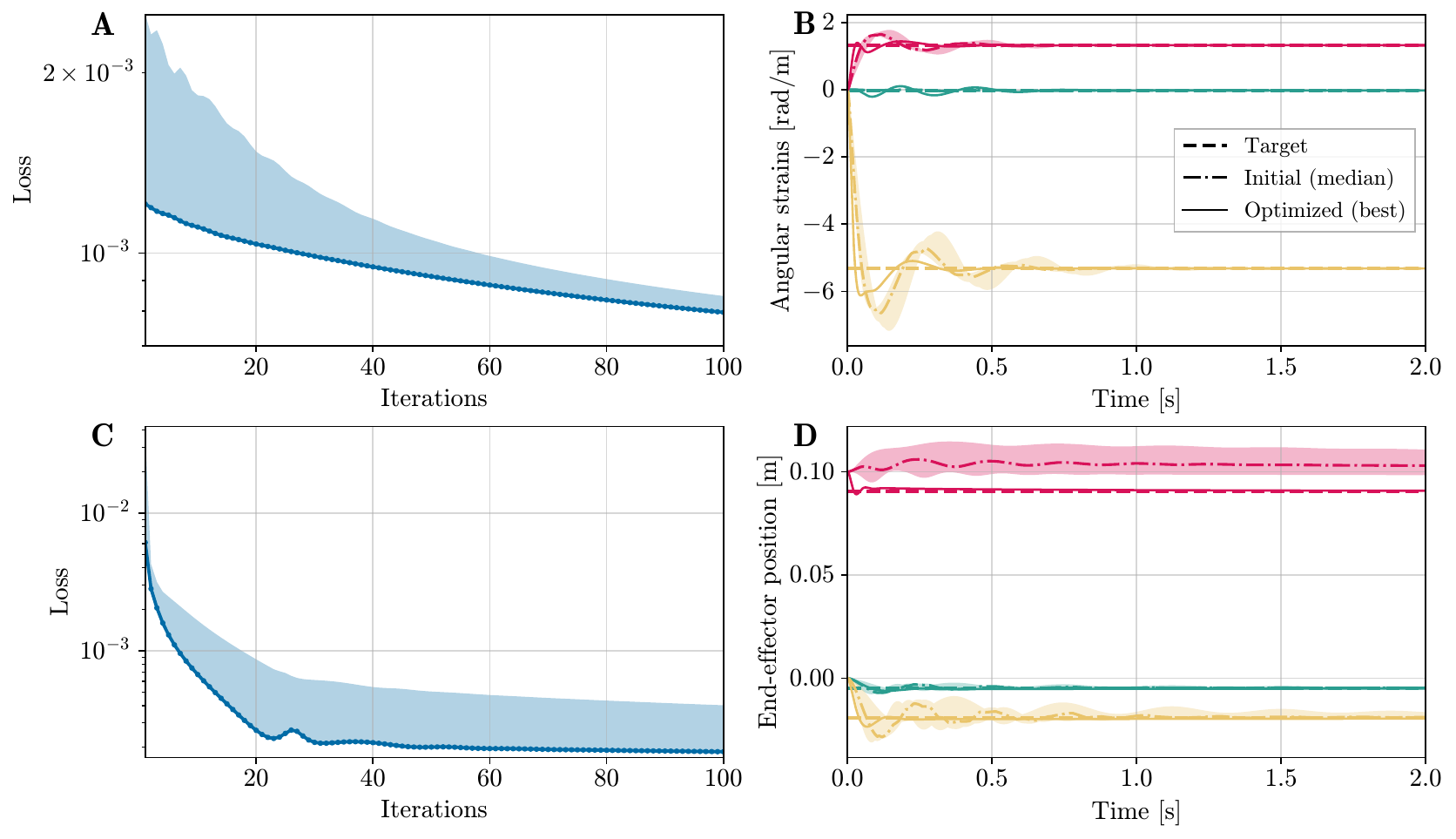}
    \caption{Results of the control gain optimization demo.
    (\textbf{A}) Loss function \eqref{eq:loss_fn} along the iterations of the system controlled in actuation coordinates via \eqref{eq:ctrl_as}; the system with minimum loss across the batch is shown in solid dotted line, while the shaded area represents the min-max range across the batch.
    (\textbf{B}) The first three configuration variables (coinciding with the angular strains) of the one-segment PCS system, showing in dashed line the target setpoint, in dash-dot line the median system across the batch at the first iteration, in solid line the median system across the batch at the last iteration; the shaded areas represent the min-max range across the batch of each strain.
    $\kappa_\mathrm{x},\kappa_\mathrm{y},\kappa_\mathrm{z}$ shown in yellow, green, and red, respectively.
    (\textbf{C}) Same as (\textbf{A}) but for the two-segment PCS system controlled via \eqref{eq:ctrl_os}.
    (\textbf{D}) Tip position of the two-segment PCS system controlled via \eqref{eq:ctrl_os}; $x,y,z$ coordinates shown in yellow, green, and red, respectively.
    }\label{fig:controlgainopt}
\end{figure*}

Two single-segment, tendon-driven \ac{PCS} robots, each with three linearly routed tendons, are considered: one is regulated in actuation coordinates using a potential-shaping controller~\cite{pustina2025analysis}, and the other in operational space using a synergistic controller~\cite{della2019exact, della2023model}. The optimized variables are the diagonal proportional, integral, and derivative gains, and the objective is the time integral of the squared closed-loop tracking error. \Cref{fig:controlgainopt} reports the loss over $100$ optimization iterations and the corresponding convergence of the configuration and tip position to their targets.

Quantitatively, the loss decreases by \SI{46}{\percent} from the initial median values to the optimized best values across the batch for the \emph{actuation-space control} case. The optimized gains reduce the settling time, defined by the \SI{5}{\percent} target band, by \SI{52}{\percent}, and the overshoot by \SI{29}{\percent}. In the \emph{operational-space control} case, the loss decreases by \SI{98}{\percent}, where the system shifted from non-converged within \SI{5} seconds, to reach the target with a settling time of \SI{900}{ms} and an overshoot of \SI{4} mm. The final cumulative tracking errors are \SI{0.20}{rad \per m} and \SI{3}{mm}, respectively.

\subsection{Safety-Constrained Control}
Safety-constrained control drives a soft robot toward a target while enforcing contact-force limits near obstacles. Compliance can reduce impact severity, but it does not by itself guarantee bounded forces in closed loop, so control-barrier-function methods are increasingly used to impose explicit force-safety constraints~\cite{ames2016control,dickson2025safe,wong2025contact}.

The central algorithmic challenge is formulating these constraints for accurate soft robot dynamics rather than only simplified template models. The contact-force barrier depends on task and contact quantities whose derivatives must be propagated through second-order soft robot dynamics. \ac{SoRoMoX} provides differentiable forward kinematics and dynamics for \ac{SOTA} rod-based soft robot models, enabling the derivation and implementation of high-order control Lyapunov and control barrier function (\acp{HOCLF}+\acp{HOCBF}) constraints used in this case study.

The simulated robot is a tendon-actuated \ac{PCS} continuum arm with two \SI{0.15}{m} segments, total backbone length \SI{0.30}{m}, and six straight tendons. In panels A and B of \Cref{fig:CBF_RL}, the arm is commanded toward a target while satisfying a prescribed contact-force bound of \SI{5}{N}. We compare a safety-unaware \ac{HOCLF} controller with a sequential-projection \ac{HOCLF}--\ac{HOCBF} controller that prioritizes the contact-force constraint.

Quantitatively, the nominal \ac{HOCLF} controller achieves the smallest terminal goal error, with final goal distance approximately \SI{0.003}{m}, but its maximum pairwise normal force reaches approximately \SI{33.5}{N}, exceeding the prescribed force bound by about \SI{28.5}{N}. In contrast, the \ac{HOCLF}--\ac{HOCBF} controller keeps the maximum pairwise normal force at or below the \SI{5}{N} safety limit throughout the simulation, while reaching a final goal distance of approximately \SI{0.032}{m}. Equivalently, the inclusion of the \ac{HOCBF} term reduces the maximum contact-force violation, normalized by the \SI{5}{N} bound, from approximately \SI{570}{\percent} to \SI{0}{\percent}.

\subsection{Parallel Reinforcement Learning}
\begin{figure*}
    \centering    \includegraphics[width=1.00\linewidth]{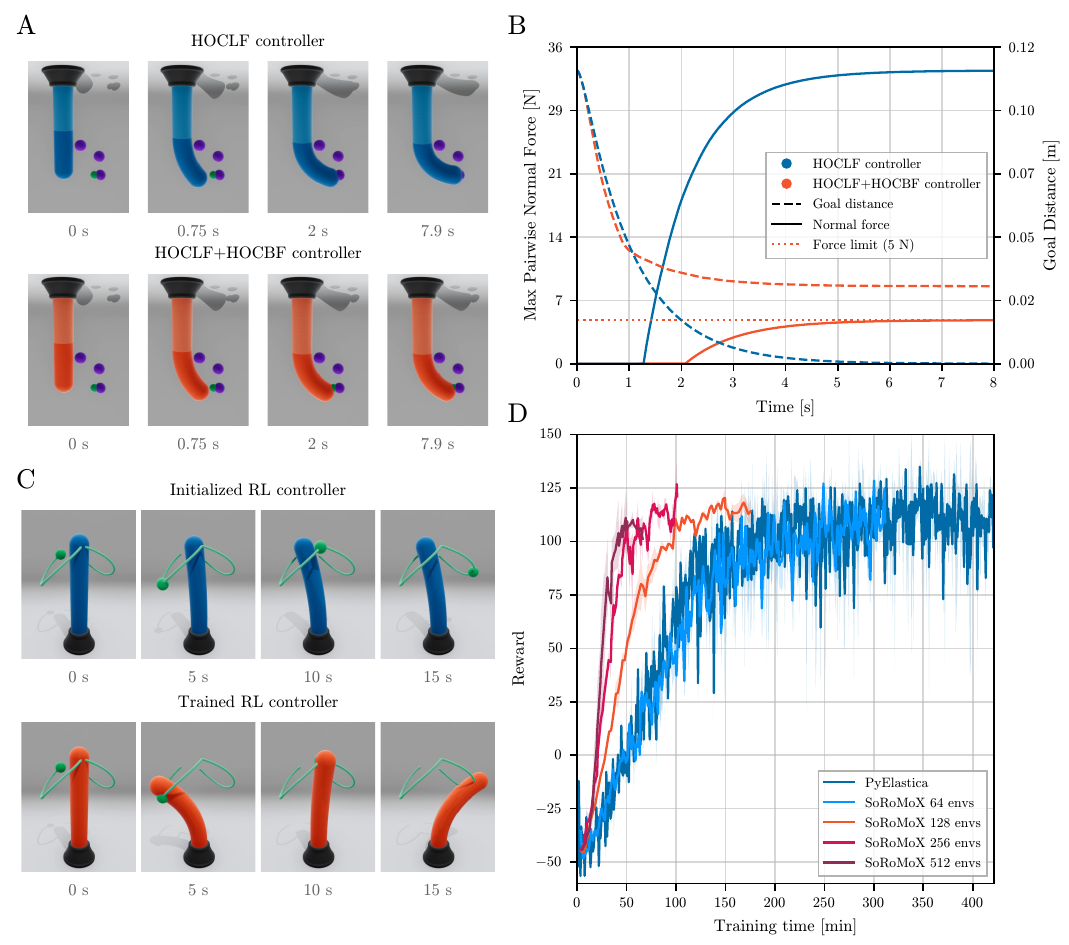}
    \caption{\textbf{Safety-constrained regulation under the HOCLF--
    HOCBF controller and parallel RL for goal tracking.}
    (\textbf{A}) Rendered shape evolution of a safety-unaware HOCLF controller (top) and a safety-constrained HOCLF+HOCBF controller (bottom) on the task end-effector regulation for a soft continuum arm (two \ac{PCS} segments). The end-effector goal is marked in green, and the spherical obstacles are rendered in purple. (\textbf{B}) Time evolution of the contact force and goal distance for both controllers during contact-force-constrained goal reaching, comparing the safety-unaware HOCLF and the safety-constrained HOCLF+HOCBF. The dotted orange line indicates the prescribed contact-force limit.
    (\textbf{C}) RL task description: a tendon-actuated soft continuum arm (one constant-strain segment) is trained to track a dynamically moving target constrained on a hemispherical surface. The target (green sphere) follows a smooth, velocity-limited trajectory (green line). The top panel shows a random policy that leads to poor tracking performance. The bottom panel shows a successful policy learned via reinforcement learning, where the arm accurately follows the target trajectory in 3D space. (\textbf{D}) Comparison of parallel training efficiency between SoRoMoX and PyElastica. The plot shows the evolution of the mean episode reward over time (in minutes) for different leading batch sizes $n_\mathrm{b}$. SoRoMoX demonstrates faster convergence and achieves higher rewards at earlier stages of training. The advantage becomes more pronounced with greater parallelization: using \(n_\mathrm{b}=256\) parallel environments yields an approximately 4$\times$ training speedup, while scaling to \(n_\mathrm{b}=512\) parallel environments achieves up to a 7$\times$ speedup relative to the CPU PyElastica DCM baseline.}
    \label{fig:CBF_RL}
\end{figure*}

Reinforcement learning trains a feedback policy through interactions with an environment or a simulation of that environment~\cite{sutton2018reinforcement}. In this case study, the environment is a tendon-actuated soft continuum arm tracking a moving target on a hemispherical surface, as shown in panel C of \Cref{fig:CBF_RL}.

The central computational challenge is the sequential nature of rollout-based policy improvement. When trajectories are collected one simulation at a time, each policy update is limited by the rate at which new interaction data can be generated, causing learning to progress slowly. This limitation has motivated massively parallel rollout pipelines for legged-robot \ac{RL}, where quadruped and biped locomotion policies can be trained in minutes rather than days~\cite{rudin2022learning,makoviychuk2isaac,grandia2024design}. \ac{SoRoMoX} uses JAX/GPU vectorization to run many \ac{PCS} rollout environments in parallel with the same robot model, increasing interaction throughput on a single GPU.

We train the policy with PPO~\cite{schulman2017proximal} using tendon-force actions for a single-segment \ac{PCS} continuum robot actuated by four tendons, and compare JIT-compiled, GPU-parallel \ac{SoRoMoX} training against a two-core CPU PyElastica baseline using a \ac{DCM} in panel D of \Cref{fig:CBF_RL}. Quantitatively, across \(n_\mathrm{b}=128\) parallel evaluation trajectories, the trained policy reaches a final mean tracking error of \SI{6.19}{mm} and a step-wise success rate of \SI{91.4}{\percent}, where success is defined as the end effector being within \SI{10}{mm} of the target. Relative to this baseline, \ac{SoRoMoX} reduces wall-clock training time by approximately \(4\times\) and \(7\times\) with \(n_\mathrm{b}=256\) and \(512\), respectively; these values jointly reflect model formulation, JIT compilation, hardware acceleration, and rollout parallelization.

\begin{soromoxbox}[Sidebar 1: What Makes a Soft Robot Model ``Control-Ready''?]
A rollout function is enough to simulate a soft robot, but too narrow to control
one. Model-based control, system identification, trajectory optimization, and
safety filtering all need intermediate model quantities, not just trajectories.
A control-ready implementation therefore exposes, with a documented contract, at
least: backbone kinematics $g(q,s)$; the Jacobian $J(q,s)$ and its time
derivative $\dot{J}(q,\dot q,s)$; the \ac{EOM} terms $M(q)$,
$C(q,\dot q)$, $G(q)$, $K(q)$, $D$; the actuation map $A(q)$; energies; and forward dynamics $\dot y = f_0(y) + f_u(y) \, u$, rather than only a black-box rollout.

Two design rules make this usable. First, every method states its mathematical
contract: input/output shapes, coordinate conventions, reference frames, units,
sign conventions, and valid domains. This matters acutely in soft robotics, where
poses may live in $\mathrm{SE}(2)$ or $\mathrm{SE}(3)$, Jacobians may be body- or
inertial-frame, actuation may mean tendon tensions, pressures, or generalized
forces, and the arclength $s$ may be a length, a normalized coordinate, or a link
index. Second, the same contract holds across model classes, so a controller,
optimizer, or renderer can call \texttt{inertia\_matrix} or
\texttt{actuation\_matrix} without knowing whether the underlying model is \ac{PCS},
\ac{GVS}, or articulated or which compatible actuator components it uses. The payoff is reuse: modeling
assumptions, controllers, coordinate transformations, and benchmarks all compose.
Looking ahead, the field would benefit from a standardized robot-description
format for continuum soft robots, a soft robot analogue of URDF, to make
models portable across simulators and control stacks.
Further details about the API interface can be found in \Cref{sub:api_overview} and the corresponding code blocks.
\end{soromoxbox}

\begin{soromoxbox}[Sidebar 2: Structure-Aware Evaluation]
\ac{SoRoMoX} exposes model structure explicitly instead of routing every strain-based robot through the most general formulation.
Dedicated planar \ac{PCS}, spatial \ac{PCS}, and \ac{GVS} systems use, respectively, lower-dimensional $SE(2)$ operations, closed-form constant-strain segment maps, and numerical variable-strain integration.
This specialization is complementary to \ac{GVS}-centered implementations such as SoRoSim~\cite{mathew2022sorosim}: the general formulation remains available, while simpler model assumptions can be evaluated with lower-dimensional algebra and fewer integration operations.

A second important distinction is the use of a static runtime layout together with early reduction to active coordinates.
Heterogeneous \ac{GVS} segment descriptions are converted into fixed-shape arrays for compilation, but the generalized state remains restricted to active coordinates through selector maps, gather indices, and masks.
Consequently, inactive strain components are not merely discarded after assembling full dynamical matrices; they are removed before propagating Jacobians, Jacobian derivatives, and dynamic integrands whenever possible.
This makes reduced parametrizations, such as \ac{PCC}~\cite{webster2010design} models with only bending strains, cheaper than a full \ac{PCS}~\cite{renda2018discrete} parametrization even when both share much of the same implementation path.

The propagation of state-dependent quantities follows the serial structure of the robot body.
Forward kinematics, Jacobians, and Jacobian derivatives are advanced from proximal to distal segments with \texttt{lax.scan}, whereas independent Gaussian-quadrature contributions within a segment are evaluated in parallel using batched array operations such as \texttt{vmap}.
State-independent data, including quadrature grids, basis evaluations, local mass matrices, and active stiffness and damping blocks, are cached at model construction time rather than reconstructed during every dynamics evaluation.
Related constant-strain operators are also evaluated together when they are needed at the same strain and arclength: for example, the adjoint, tangent map, and tangent derivative reuse the same trigonometric coefficients and Lie-algebra matrix powers rather than recomputing them independently.
Finally, the implementation distinguishes between query types: routines that only require potential forces, gravity, or inertia avoid computing Jacobian time derivatives or convective-force terms, whereas forward dynamics uses a fused path that reuses shared kinematic quantities and directly evaluates $C(q,\dot{q})\dot{q}$.
Further details on efficient evaluation are given in \Cref{sec:app:computational_efficiency}.
\end{soromoxbox}

\begin{soromoxbox}[Sidebar 3: Singularities Are Not Failures]
Closed-form strain-based models often contain expressions that have well-defined
limits but appear numerically as $0/0$ in their raw algebraic form. For example,
Lie-group exponentials and constant-strain tangent maps contain the coefficient
$c(\vartheta)=(1-\cos\vartheta)/\vartheta^2$, whose analytical limit at
$\vartheta=0$ is $c(0)=1/2$.
For a constant strain propagated along a segment, the Lie-algebra argument is
$\chi=s\xi$, with rotational part $\chi_\mathrm{rot}=(s\xi)_\mathrm{rot}$.
The scalar $\vartheta=\|\chi_\mathrm{rot}\|=|s|\,\lVert\xi_{\mathrm{rot}}\rVert$ vanishes
not only for straight or zero-bending shapes but also at zero arclength,
$s\to0$. These are removable singularities: the underlying maps have finite
limits; for example, $\lim_{s\to0}\exp_{\mathrm{SE(3)}}(s\xi)=I_4$. The error is
never the model, only a naive implementation that evaluates the indeterminate
ratio.

The general remedy is to select an analytical limit or Taylor approximation near
$\vartheta=0$ and the closed form away from it. Importantly, the singular closed
form should not be evaluated when the regularized branch is selected: computing
both expressions and only masking the invalid output can still expose automatic
differentiation to undefined intermediate values.\footnote{Some array-programming
and compiler transformations replace conditional execution by elementwise
selection. In JAX, this can occur when \texttt{vmap} vectorizes a scalar
conditional; the additional input guards used in \ac{SoRoMoX} are described in
\Cref{sec:app:singularities}.} Done properly, both the function values and their
gradients stay well defined for straight reference shapes and other
zero-rotational-strain cases, as well as at zero arclength; these are precisely
the regimes where controllers and optimizers tend to operate.
Further details on singularity-safe evaluation are given in \Cref{sec:app:singularities}.
\end{soromoxbox}

\begin{soromoxbox}[Sidebar 4: Trust, but Verify: Correctness as a Tested Property]
Soft robot kinematics and dynamics are too long to validate by manual inspection:
even a single constant-strain segment produces intricate dynamic expressions.
Correctness is therefore best treated as a property that is continuously tested,
not checked once. We use a layered strategy, run automatically in a
continuous-integration (CI) pipeline so it survives every refactor, performance
tweak, and extension:

\begin{enumerate}
  \item \emph{Primitives.} Verify the $\mathrm{SE}(2)$/$\mathrm{SE}(3)$
        Lie-algebra operations, namely hat maps, exponentials and logarithms,
        adjoints, and tangent maps, against their definitions and known block
        structure.
  \item \emph{Special cases.} Check kinematics and dynamics where closed-form
        answers are easy: straight reference shapes, pure axial-strain displacement, zero velocity,
        single planar segments.
  \item \emph{Internal consistency.} Cross-check independent routes to the same
        quantity: pointwise versus batched evaluation; forward and inverse
        kinematic loops; body- versus inertial-frame Jacobians; fused versus
        unfused forward dynamics; and explicit expressions against identities
        obtained by automatic differentiation, for example
        $\dot J=\mathrm{AD}_q(J)\dot q$, or conservative forces as gradients of
        potential energy.
  \item \emph{Cross-model coherence.} Embed specialized models in more general
        ones, for example planar PCS in a \ac{GVS} implementation configured with constant strain bases, and require agreement on matched
        geometric and material parameters, reference strains, and generalized configurations.
\end{enumerate}

Finally, randomized tests sample near singular configurations and assert finite
outputs and well-defined forward- and reverse-mode gradients. Targeted tests also
differentiate vectorized kinematics in reverse mode at the straight configuration
and compare bundled constant-strain operators with their separately evaluated
counterparts. To our knowledge, this kind of systematic correctness testing for
soft robot models has not been discussed in the literature; yet it is what lets
the package be refactored and optimized without silent regressions.
Further verification details are given in \Cref{sec:app:verification}.
\end{soromoxbox}

\section{Conclusion}\label{sec:conclusion}
This article presented \ac{SoRoMoX}, a fast, differentiable, and parallelizable JAX/Python framework for control-oriented soft robot models, with Warp kernels accelerating \ac{PCS}- and \ac{GVS}-based continuum models. By providing fully numerical and \ac{JIT}-compatible implementations of articulated and rod-based models, including \ac{PCS} and \ac{GVS} formulations, \ac{SoRoMoX} makes key model quantities accessible for simulation, control, optimization, learning, and safety-constrained decision making. The benchmarking results and application case studies show that modern software design choices such as automatic differentiation, batched execution, and accelerator compatibility can substantially broaden the practical use of soft robot models.

Future development will extend \ac{SoRoMoX} along several complementary directions. A faster planar \ac{VS} setting would support rapid prototyping of controllers and algorithms that use higher-order geometric shape parametrizations before moving to full spatial models. Existing and future systems could also be extended with kinematic trees for branched soft robot architectures and closed-chain kinematics for systems such as parallel soft robots. The actuation interface can also be expanded beyond the currently implemented modalities to include shape-memory alloys, dielectric elastomer actuators, electrohydraulic actuators, hydrogels, jamming-based mechanisms, and other emerging soft-actuation technologies~\cite{yasa2023overview}. Finally, providing ready-to-use model instantiations for specific soft robots, especially open-source designs, would make the framework easier to benchmark, reproduce, and adopt across the community. More broadly, a standardized robot-description protocol for continuum and soft robots could help the community share robot designs across simulators and model implementations, reducing the effort required to turn a physical or open-source design into an executable computational model.

\ifanonymoussubmission
\else
\section*{Acknowledgments}
 The work by M. Stölzle was supported under the European Union's Horizon Europe Program from Project EMERGE - Grant Agreement No. 101070918. The work by S. Gribonval was supported by the Erasmus+ program of the European Union. The work by D. Feliu-Talegon was supported by the Dutch Research Foundation (NWO) through the VENI grant ROSES 20297. The work of V.D. Perfetta and C. Della Santina was supported by the European Union's ERC grant RIPLEy No. 101165078. The work by K. Wong was supported by The Hong Kong Jockey Club Scholarships. This work by M. Tarnini, A. Teejo Mathew, and F. Renda was funded by the Center for Autonomous Robotic Systems, Khalifa University of Science and Technology (KU-CARS). The work by D. Rus and M. Stölzle was supported in part by the Singapore MIT Alliance on Research and Technology (SMART) under the Mens, Manus, et Machina (M3S) program, by the BARI EMERGE program under grant N00014-26-1-2304, and by the MIT-GIST collaboration.
The work of C. Della Santina was further supported by the GRAIL project under Grant Agreement No.~101298421.

We thank Runze Zuo and Daniel Bruder from the University of Michigan for contributing the model for the UMArm~\cite{zuo2025umarm} and Daniele Caradonna from the Biorobotics Institute, School of Advanced Studies Sant'Anna for his contributions to the I-Support~\cite{arleo2021isupport} model implementation and the identification of system parameters.
\fi

\bibliographystyle{IEEEtran}
\bibliography{main.bib}



\clearpage
\appendices
\crefalias{section}{appendix}
\crefalias{subsection}{appendix}
\crefalias{subsubsection}{appendix}

\section{Core Abstractions and Interface Contracts}\label{sec:app:api_interfaces}
\Cref{sec:getting_started} presents the principal user-facing operations. This appendix specifies the abstractions that support model construction, differentiation, simulation, control, and rendering without cataloguing the complete \ac{API}.

\subsection{Parameters and Structure}
System definitions distinguish user-facing construction specifications, dynamic parameter PyTrees, and static structure. For continuum systems, \texttt{LinkSpec}, \texttt{JointSpec}, \texttt{GVSSegment}, and \texttt{StrainBasisSpec} describe physical components and topology during construction. Factory methods convert them into \texttt{PCSParams}, \texttt{PlanarPCSParams}, or \texttt{GVSParams} together with the corresponding static structure. Runtime link arrays---including length, density, cross-section coefficients, reference strain, and canonical generalized stiffness and damping matrices---live under \texttt{params.link}; \ac{GVS} additionally stores canonical joint stiffness and damping under \texttt{params.joint}. Isotropic Young's modulus, shear modulus or Poisson's ratio, and material damping remain caller-owned \texttt{IsotropicMaterialParams}. Structure records choices that determine array layout or compiled control flow, including quadrature counts, active-strain masks, joint, basis, and cross-section types, padding limits, and actuator topology.

\texttt{with\_params(params)} replaces the complete parameter object, while \texttt{update\_params(...)} replaces top-level fields such as \texttt{link}, \texttt{joint}, or \texttt{gravity}. The narrower \texttt{update\_link\_params(...)} method replaces nested link values, and \ac{GVS} additionally provides \texttt{update\_joint\_params(...)}. Fixed mounting is instead passed as the constructor's \texttt{base\_pose}, exposed as \texttt{fixed\_base\_pose}, and changed with \texttt{with\_fixed\_base\_pose(...)}. To update interpretable isotropic material variables, \texttt{with\_isotropic\_material(material)} explicitly maps the caller-owned material PyTree to canonical link stiffness and damping matrices. All methods return a new system, check compatibility with its structure, and refresh dependent caches. Compatible numerical parameters can be updated inside compiled JAX transformations without reconstructing the model, whereas structural changes require reconstruction and recompilation. Eager updates validate values and structure, while compiled updates enforce structural compatibility.

\subsection{System Interfaces and State}
\texttt{SoftRobot} exposes separate dimensions \texttt{num\_internal\_dofs}, \texttt{num\_coordinates}, \texttt{num\_velocities}, and \texttt{state\_size}. Its plant state and derivative have the generic form
\begin{equation}
    \begin{aligned}
        y&=[q^\top,v^\top,\chi^\top]^\top,\\
        \dot y&=f\!\left(t,y;(u,\tau_\mathrm{ext})\right)
        =[\left(N(q)v\right)^\top,\dot v^\top,\dot\chi^\top]^\top,
    \end{aligned}
    \label{eq:app:forward_dynamics_contract}
\end{equation}
where \(q\in\mathbb{R}^{n_q}\), \(v\in\mathbb{R}^{n_v}\), and \(\chi\) is optional auxiliary state. For fixed systems, \(N(q)=I\) and \(v=\dot q\); floating systems can have \(n_q\neq n_v\). The helpers \texttt{pack\_configuration}, \texttt{pack\_velocity}, and \texttt{pack\_state}, with their corresponding \texttt{split\_*} methods, preserve this layout and arbitrary leading batch axes. \texttt{forward\_dynamics(t, y, actuation\_args)} returns an array with the shape of \(y\); the optional actuation tuple conventionally contains the input \(u\) and external generalized force \(\tau_\mathrm{ext}\). The base class supplies open-loop, continuous closed-loop, and sampled-data rollouts.

For a spatial floating system, the leading base coordinates are \([q_x,q_y,q_z,q_w,p_x,p_y,p_z]\), using a scalar-last Hamilton quaternion, and the leading generalized velocities are \([\omega_x,\omega_y,\omega_z,v_x,v_y,v_z]\). Both angular and linear base velocities are expressed in the world frame, with the linear velocity taken at the base-frame origin. A compact state construction is:
\begin{lstlisting}[style=soromoxpython,numbers=none,aboveskip=6pt,belowskip=6pt]
robot = PCS.from_links(links, floating_base=True)

q = robot.pack_configuration(
    jnp.zeros(robot.num_internal_dofs),
    base_pose=jnp.array([0., 0., 0., 1., 0., 0., 0.]),
)
v = robot.pack_velocity(
    jnp.zeros(robot.num_internal_dofs),
    base_velocity=jnp.zeros(6),
)
y = robot.pack_state(q, v)

q, v, auxiliary = robot.split_state(y)
base_pose, q_internal = robot.split_configuration(q)
base_velocity, v_internal = robot.split_velocity(v)
\end{lstlisting}

\texttt{SystemState} is the public JAX PyTree exchanged with rollouts, controllers, and environment models. It contains time \(t\), plant state \(y\), and optional fields for the applied input \(u\), controller state, and environment state. The latter two may be arbitrary JAX-compatible PyTrees and are integrated alongside the plant when present. A returned trajectory uses the same object with a leading time dimension on every populated leaf. \texttt{SoftRobot} extends \texttt{DynamicalSystem} with arc-length kinematics and Jacobians, dynamical and energy terms, cross-section geometry, and composed actuation.

\subsection{Analytical Derivative Paths}
Selected \texttt{SoftRobot} methods define custom Jacobian--vector products (JVPs). Their public signatures are unchanged: the derivative rule calls an analytical model hook when provided and otherwise differentiates the protected primal implementation. Representative identities are
\begin{equation}
    \begin{aligned}
        \nabla_q \mathcal{U}(q)
        &=\tau_\mathcal{U}(q)=G(q)+K(q),\\
        \operatorname{JVP}_{q}[J](q,s;\dot{q})
        &=\dot{J}(q,\dot{q},s).
    \end{aligned}
    \label{eq:app:custom_jvp_identities}
\end{equation}
Analytical arc-length derivatives similarly reuse pose and Jacobian recurrences. \ac{PCS} provides closed-form recurrences for selected kinematic paths and analytical force--energy relations. Other tangent combinations use \ac{AD} when an analytical rule is unavailable or not advantageous. Consequently, the same \texttt{jax.grad}, \texttt{jax.jvp}, \texttt{jax.jacfwd}, and \texttt{jax.jacrev} calls apply to all models.

\subsection{Task- and Actuation-Space Transformations}
\texttt{OperationalSpaceDynamics} defines a task by backbone coordinates \texttt{s\_ps} and a Boolean \texttt{task\_selector}. A single point yields an end-effector task; multiple points yield a shape task. The selector may be common to all points or point-specific and defaults to every twist component. Spatial tasks select three angular and three linear components; planar tasks select one angular and two linear components. Omitting angular components gives a position task. Planar orientation is selected independently, whereas spatial orientation includes either all three angular components or none at each point.

\texttt{rotation\_representation} selects rotation vectors, scalar-last quaternions, or continuous 6D coordinates for spatial poses and references; planar orientation is scalar. Jacobians and dynamics remain twist-based. Pose and velocity dimensions may therefore differ for quaternion or 6D representations. Full poses are retained for geometric orientation errors, while the selector determines task velocities, Jacobians, and forces.

For a full-row-rank task Jacobian, the adapter uses~\cite{khatib2003unified}
\begin{equation}
    \Lambda=(J M^{-1}J^\top)^{-1},\qquad
    J_\mathrm{M}^+=M^{-1}J^\top\Lambda,
    \label{eq:app:operational_space_map}
\end{equation}
and re-exposes the system matrix and force methods in task coordinates. \texttt{ActuationSpaceDynamics} augments the installed transmission coordinates \(\varphi_\mathrm{a}(q)\) with unactuated coordinates \(H_\mathrm{u}q\) to form the complete actuation-space coordinates \(\varphi\)~\cite{pustina2024input}:
\begin{equation}
    \varphi=\begin{bmatrix}\varphi_\mathrm{a}(q)\\H_\mathrm{u}q\end{bmatrix},\qquad
    A_\varphi=\begin{bmatrix}I\\0\end{bmatrix}.
    \label{eq:app:actuation_space_map}
\end{equation}
The user may specify \(H_\mathrm{u}\) directly. Otherwise, the constructor evaluates \(A(0)\), verifies \(\operatorname{rank}A(0)=m\), and computes a complete QR factorization. If \(Q_\perp\) contains the final \(n-m\) columns of \(Q\), it sets \(H_\mathrm{u}=Q_\perp^\top\); its rows therefore form an orthonormal basis of \(\ker A(0)^\top\). For \(m=n\), \(H_\mathrm{u}\) is empty. The resulting coordinate Jacobian \(J_\varphi(q)=\bigl[A(q)\;\;H_\mathrm{u}^\top\bigr]^\top\) must remain nonsingular. The transformed interface is collocated and retains the configuration-space matrix and force vocabulary.

\subsection{Reference Trajectories and the Controller Contract}
\texttt{ReferenceTrajectory} accepts time samples or continuous functions for desired position or pose, velocity, and acceleration. It generates the complementary representation by linear interpolation or vectorized evaluation. Missing velocity and acceleration functions are obtained by differentiating the available position and velocity functions. Pose references instead convert orientation derivatives to geometric twists using the declared rotation representation, task-point count, and planar or spatial convention.

Feedback controllers generally consume position and velocity; tracking feedforward and inverse-dynamics terms also require acceleration. Derivatives generated from sampled positions correspond to the piecewise-linear interpolant. Smooth feedforward acceleration therefore requires explicit derivative data or a smooth continuous reference. Actuation-space conversion maps position through the coordinate map, velocity through its Jacobian, and acceleration through the Jacobian and its time derivative. Operational-space controllers retain full poses for geometric errors before velocity-space selection.

The controller contract is the JAX-compatible map \texttt{SystemState} \(\mapsto(u,\dot z)\), with optional controller state \(z\). \texttt{ClosedFormModelBasedController} decomposes \(u\) into model-based and feedback terms and combines their state derivatives. Continuous rollouts evaluate this map at the integrator time and integrate \(z\) with the plant. Sampled-data rollouts evaluate it every \texttt{control\_dt}, hold \(u\) between updates, and advance \(z\) at the sampling boundary.

\subsection{Composable Actuation Models}\label{sec:app:actuation_api}
\texttt{Transmission} defines actuator coordinates \(\varphi_\mathrm{a}(q)\); \texttt{EffortModel} maps the control input to work-conjugate effort \(e\); \texttt{Actuator} composes both. \texttt{PassiveElement} contributes conservative forces, damping, and energy without a control channel. Compatibility is checked during system construction.

\begin{equation}
    A(q)=\left(\frac{\partial \varphi_\mathrm{a}(q)}{\partial q}\right)^\top,
    \qquad
    \dot{\varphi}_\mathrm{a}=A(q)^\top\dot{q},
    \qquad
    \tau_u=A(q)e,
    \label{eq:app:actuation_api}
\end{equation}
defines the common work-conjugate interface and preserves power, \(\dot{q}^\top\tau_u=\dot{\varphi}_\mathrm{a}^\top e\). \texttt{DirectEffort} sets \(e=u\); other effort laws may depend on actuator velocity or state. Body, actuator, and passive-element coefficients use immutable full, selective, or indexed updates. Component type, channel count, and routing topology remain structural.

\section{Additional Computational Performance Evaluation Details}\label{sec:app:benchmarking}
\subsection{CPU Sequential-Rollout Benchmark}\label{sec:app:cpu_benchmark}
For the CPU rollout comparison in \Cref{tab:comparison_speed}, we consider a cylindrical beam with a length of \SI{0.6}{m}, a radius of \SI{3}{cm}, a density of \SI{1000}{kg \per m^3}, a Young's modulus of \SI{1}{MPa}, a Poisson ratio of $0.5$, and a damping coefficient of $10000$. The benchmark was run on a PC with an Intel(R) Core(TM) Ultra 7 165H CPU (1.40 GHz) and \SI{16}{GB} RAM. We use two piecewise elements for SoRoMoX and SoRoSim, and $10-100$ elements for PyElastica for every scenario. The number of elements was selected to achieve comparable simulation accuracy across the three libraries. The simulation is performed over \SI{3}{s}.

SoRoSim and SoRoMoX use explicit Runge--Kutta 5(4) methods: the Dormand--Prince method implemented in MATLAB's \texttt{ode45} and the Tsitouras method implemented in \texttt{Tsit5}~\cite{tsitouras2011runge}, respectively. Both pair a fifth-order solution with an embedded fourth-order estimate that enables adaptive step-size control. In this benchmark, \texttt{ode45} uses adaptive internal steps, while SoRoMoX runs \texttt{Tsit5} with a fixed integration time step of $\delta t = \SI{0.1}{ms}$. PyElastica uses the second-order symplectic \texttt{PositionVerlet} integrator with the same fixed step size.

Three different scenarios are simulated, where the system is initially oriented upward along the global $Z$ direction. First, a beam is subjected to a distributed force with magnitude equal to gravity, applied along the global $X$ direction, resulting in deformation in the $X-Z$ plane. Second, a beam is subjected to distributed forces with magnitude equal to gravity, applied along both the global $X$ and $Y$ directions, resulting in fully three-dimensional deformation. Third, a beam is subjected to a distributed force with magnitude equal to gravity applied along the global $Z$ direction, combined with two straight tendons running from the base to the tip. These tendons are positioned \SI{2}{cm} from the center and separated by \SI{90}{\degree}, inducing deformation in multiple directions. The second scenario is simulated in SoRoSim and SoRoMoX using both spatial PCS and \ac{GVS} formulations to compare computational efficiency. In PyElastica, the second scenario remains identical, as PyElastica does not distinguish between strain-basis implementations because it relies on an \ac{FEM} formulation.

We quantify the equal-time tip-translation agreement between SoRoSim and SoRoMoX in \Cref{tab:app:rollout_agreement}. Let $p_\mathrm{X}(t_k)$ and $p_\mathrm{S}(t_k)$ denote the respective tip positions at a saved timestamp $t_k$ common to both outputs. For $N$ corresponding samples, we use
\begin{equation}
\begin{aligned}
d_k &= \lVert p_\mathrm{X}(t_k)-p_\mathrm{S}(t_k)\rVert_2,\\
\bar d &= \frac{1}{N}\sum_{k=1}^{N}d_k,
& \operatorname{RMSE}_p &= \sqrt{\frac{1}{N}\sum_{k=1}^{N}d_k^2},\\
d_\mathrm{max} &= \max_{k=1,\ldots,N}d_k.
\end{aligned}
\end{equation}
No temporal or phase alignment is applied, so the metrics compare the two rollouts at identical physical times.

\begin{table}[t]
\centering
\caption{\textbf{Equal-time tip-translation agreement between SoRoSim and SoRoMoX.}
The table reports the mean, root-mean-square error (RMSE), beam-length-normalized RMSE, and maximum Euclidean position-error magnitude over each rollout.}
\label{tab:app:rollout_agreement}
\begingroup
\small
\setlength{\tabcolsep}{3pt}
\begin{tabular}{@{}lrrrr@{}}
\toprule
\textbf{Case}
& \makecell{\textbf{Mean}\\$[\si{mm}]$}
& \makecell{\textbf{RMSE}\\$[\si{mm}]$}
& \makecell{\textbf{RMSE}/\boldmath$L$\unboldmath\\$[\si{\percent}]$}
& \makecell{\textbf{Maximum}\\$[\si{mm}]$} \\
\midrule
Planar PCS        & 2.36 & 2.75 & 0.46 & 5.13 \\
Spatial PCS       & 6.31 & 7.21 & 1.20 & 14.14 \\
Complex GVS       & 0.71 & 0.86 & 0.14 & 1.82 \\
Tendon-driven GVS & 4.81 & 5.82 & 0.97 & 10.40 \\
\bottomrule
\end{tabular}
\endgroup
\end{table}

The tip-position RMSE values remain below \SI{1}{cm} for all four cases and correspond to \SIrange{0.14}{1.20}{\percent} of the \SI{0.6}{m} beam length. The complex GVS case shows the closest agreement, with an RMSE of \SI{0.86}{mm} and a maximum error of \SI{1.82}{mm}. The planar PCS, tendon-driven GVS, and spatial PCS cases yield progressively larger RMSE values of \SI{2.75}{mm}, \SI{5.82}{mm}, and \SI{7.21}{mm}, respectively; the largest instantaneous discrepancy is \SI{14.14}{mm} for spatial PCS.

\subsection{GPU Batch-Scaling Benchmark}\label{sec:app:gpu_batch_benchmark}
For the GPU batch-scaling benchmark in \Cref{fig:benchmarking:batch_simulation_scaling}, we define the simulation-throughput factor $\Gamma_\mathrm{sim}$ as the ratio between the total simulated time and the wall-clock time:
\begin{equation}
\Gamma_\mathrm{sim} = \frac{n_\mathrm{b} \, T_\mathrm{sim}}{T_\mathrm{wall}},
\end{equation}
where $n_\mathrm{b}$ is the leading batch size, $T_\mathrm{sim} \in \mathbb{R}_+$ is the simulated time per batched rollout, and $T_\mathrm{wall} \in \mathbb{R}_+$ is the measured wall-clock runtime. Here, $\Gamma_\mathrm{sim}=1$ denotes parity between aggregate simulated time and wall-clock time.

We run the batch-scaling benchmark on a desktop workstation equipped with an NVIDIA RTX 5090 GPU with \SI{32}{GB} of video memory, an Intel(R) Core(TM) Ultra 9 285K CPU with $24$ cores, and \SI{128}{GB} of RAM. In the benchmark, we vary the leading batch size as $n_\mathrm{b} \in \{1, 2, 4, 8, 16, 32, 64, 128, 256, 512, 1024\}$ and the number of model links/segments as $N \in \{1, 2, 4, 8, 16, 32\}$. For each setting, we apply constant zero actuation and simulate the soft robot for $T_\mathrm{sim} = \SI{1}{s}$ using Tsitouras' 5/4 solver~\cite{tsitouras2011runge} with a time step of $\delta t = \SI{0.1}{ms}$. Simulation data, such as the robot configurations, are saved every \SI{0.01}{s}. For all systems that rely on numerical spatial integration, namely planar PCS, spatial PCS, and \ac{GVS}, we use $5$ Gaussian quadrature points. For \ac{GVS}, we use a zero-order basis to isolate the inherent computational penalty of the algorithm that enables a more general functional parametrization of spatial strain compared with PCS. Each configuration is evaluated after one discarded warm-up run, and the reported wall-clock runtime is the arithmetic mean of three timed executions. Independent zero-mean Gaussian perturbations with a standard deviation of $10^{-3}$ are applied to the initial generalized coordinates and velocities across the batch.
At the largest tested batch size, $n_\mathrm{b}=1024$, the aggregate simulated-time throughput of all tested articulated and planar PCS configurations, including $N=32$, meets or exceeds the wall-clock rate ($\Gamma_\mathrm{sim}\geq1$). The same holds for the tested spatial PCS and \ac{GVS} configurations with $N\in\{1,2,4,8,16\}$. These results apply only to the stated setup: the boundary depends on the hardware, simulation time step, integration solver and tolerances, spatial quadrature resolution, GVS strain-basis order, and number of segments.

\section{Additional Controller Details}\label{sec:app:controllers}
\Cref{tab:implemented_model_based_controllers} summarizes the implemented controller families. In the table caption, $A$, $M$, $C$, $G$, $K$, and $D$ denote the actuation, inertial, Coriolis, gravitational, elastic, and damping terms, respectively.

\begin{table*}[!t]
\centering
\caption{\textbf{Implemented model-based controllers in \ac{SoRoMoX}.} For a generic controlled coordinate $\zeta$, $\operatorname{PID}_\zeta=K_\mathrm{p}(\zeta^\mathrm{d}-\zeta)+K_\mathrm{i}\int\operatorname{sat}_\Gamma(\zeta^\mathrm{d}-\zeta)dt+K_\mathrm{d}(\dot{\zeta}^\mathrm{d}-\dot{\zeta})$, where $\operatorname{sat}_\Gamma(e)=\Gamma^{-1}\tanh(\Gamma e)$ acts componentwise, and $[\cdot]_\mathrm{a}$ extracts the actuated rows. For the impedance row, $F_\mathrm{x}^\mathrm{d}=(J_\mathrm{M}^+)^\top \left ( K(q)+D\dot{q} \right )+C_\mathrm{x}N_\mathrm{x}\dot{q}+\Lambda\ddot{x}^\mathrm{d}+K_\mathrm{x} e_\mathrm{x}+D_\mathrm{x}\dot{e}_\mathrm{x}$, and $N_\mathrm{x}=I-J_\mathrm{M}^+J$, with $J_\mathrm{M}^+$ the dynamically consistent pseudoinverse and $C_\mathrm{x}$ the operational-space Coriolis matrix.}
\label{tab:implemented_model_based_controllers}
\begin{adjustbox}{max width=\textwidth,max totalheight=0.79\textheight,center}
\fontsize{6}{6.6}\selectfont
\setlength{\tabcolsep}{0pt}
\renewcommand{\arraystretch}{0.74}
\newcommand{\pidycompact}{$\operatorname{PID}_{\varphi_\mathrm{a}}$}
\begin{tabularx}{\textwidth}{@{}>{\raggedright\arraybackslash}p{0.15\textwidth}@{\hspace{1.5pt}}>{\raggedright\arraybackslash}p{0.065\textwidth}@{\hspace{1.5pt}}>{\raggedright\arraybackslash}p{0.24\textwidth}@{\hspace{5pt}}>{\raggedright\arraybackslash}p{0.12\textwidth}@{\hspace{5pt}}>{\raggedright\arraybackslash}p{0.22\textwidth}@{\hspace{2pt}}>{\raggedright\arraybackslash}X@{}}
\toprule
\textbf{Controller} & \textbf{Control Task} & \textbf{Model-Based Term} & \textbf{Error-Based Feedback Term} & \textbf{Important Assumptions} & \textbf{Best Used When} \\
\midrule
\multicolumn{6}{@{}l}{\textsc{Configuration-space controllers}} \\
\cmidrule(lr){1-6}
\makecell[l]{PID\\\cite{khan2020best}} &
Setpoint Regulation &
-- &
$A^\top(q)\operatorname{PID}_q$ &
Feedback authority limited by actuator map $A(q)$; potential forces need to be compensated by integral term; quasi-static objective. &
Model uncertainty is high, and baseline regulation is sufficient. \\
\makecell[l]{Potential compensation\\ \cite{kelly1994pd,della2023model,pustina2025analysis}} &
Setpoint Regulation &
$A^{-1}(q) \left ( G(q^\mathrm{d}) + K(q^\mathrm{d}) \right )$ &
$A^\top(q)\operatorname{PID}_q$ &
Full actuation; $A(q)$ invertible; feasible setpoint; quasi-static objective. &
Fully actuated setpoint regulation with mostly convex potential forces or limited model accuracy. \\
\makecell[l]{Gravity cancellation\\ \cite{della2020model,della2023model}} &
Setpoint Regulation &
$A^{-1}(q)\!\left(G(q)+K(q^\mathrm{d})\right)$ &
$A^\top(q)\operatorname{PID}_q$ &
Full actuation; $A(q)$ invertible; feasible equilibrium; accurate gravity-force model; reliable $q$ estimate; quasi-static objective. &
Fully actuated regulation when gravity dominates, and the gravity model and $q$ are reliable. \\
\makecell[l]{Potential cancellation\\ \cite{della2020model,pustina2025analysis}} &
Setpoint Regulation &
$A^{-1}(q) \left ( G(q) + K(q) \right )$ &
$A^\top(q)\operatorname{PID}_q$ &
Full actuation; $A(q)$ invertible; accurate potential-force model; reliable $q$ estimate; quasi-static objective. &
Fully actuated regulation when potential forces are only locally convex, and the potential model and $q$ are reliable. \\
\makecell[l]{Feedforward compensation\\\cite{kelly1994pd,della2023model}} &
Trajectory Tracking &
\makecell[l]{$A^{-1}(q)\!\left(M(q^\mathrm{d})\ddot{q}^\mathrm{d}+C(q^\mathrm{d},\dot{q}^\mathrm{d})\dot{q}^\mathrm{d}\right.$\\$\left.+G(q^\mathrm{d})+K(q^\mathrm{d})+D\dot{q}^\mathrm{d}\right)$} &
$A^\top(q)\operatorname{PID}_q$ &
Full actuation; $A(q)$ invertible; feasible reference. &
Fully actuated tracking when a planned trajectory is available, but model accuracy is limited. \\
\makecell[l]{Mixed-state feedback\\ \cite{della2020model}} &
Trajectory Tracking &
\makecell[l]{$A^{-1}(q)\!\left(M(q)\ddot{q}^\mathrm{d}+C(q,\dot{q})\dot{q}^\mathrm{d}\right.$\\$\left.+G(q)+K(q^\mathrm{d})+D\dot{q}^\mathrm{d}\right)$} &
$A^\top(q)\operatorname{PID}_q$ &
Full actuation; $A(q)$ invertible; Coriolis consistency; accurate dynamical model; reliable $q$ estimate. &
Fully actuated tracking when current-state gravity/dynamics should be compensated. \\
\makecell[l]{Computed torque\\ \cite{della2020model}} &
Trajectory Tracking &
\makecell[l]{$A^{-1}(q)\!\left(M(q)\ddot{q}^\mathrm{d}+C(q,\dot{q})\dot{q}\right.$\\$\left.+G(q)+K(q)+D \dot{q}\right)$} &
\makecell[l]{$A^{-1}(q)M(q)\operatorname{PID}_q$} &
Full actuation; $A(q)$ invertible; accurate dynamical model; reliable state estimates. &
Fully actuated tracking with an accurate full model and reliable $q,\dot{q}$. \\
\addlinespace[0.12em]
\multicolumn{6}{@{}l}{\textsc{Actuation-space controllers}} \\
\cmidrule(lr){1-6}
\makecell[l]{PID\\ \cite{stolzle2024experimental,pustina2024input}} &
Setpoint Regulation &
-- &
\pidycompact &
Integrable actuation map; stable zero dynamics; potential forces need to be compensated by integral term; $\varphi_\mathrm{a}$ measured/estimated. &
Underactuated baseline when the actuation map is integrable. \\
\makecell[l]{Potential compensation\\ \cite{stolzle2024experimental,pustina2025analysis}} &
Setpoint Regulation &
$\left[G_\varphi(q^\mathrm{d}) + K_\varphi(q^\mathrm{d})\right]_\mathrm{a}$ &
\pidycompact &
Integrable actuation map; stable zero dynamics; feasible $q^\mathrm{d}$ and $\varphi_\mathrm{a}^\mathrm{d}$. &
Underactuated setpoint regulation with integrable actuation map and mostly convex potential forces. \\
\makecell[l]{Gravity cancellation\\ \cite{pustina2024input,pustina2025analysis,stolzle2025phdthesis}} &
Setpoint Regulation &
$\left[G_\varphi(q)+K_{\varphi}(q^\mathrm{d})\right]_\mathrm{a}$ &
\pidycompact &
Integrable actuation map; stable zero dynamics; accurate gravity-force model; feasible $q^\mathrm{d}$ and $\varphi_\mathrm{a}^\mathrm{d}$. &
Underactuated regulation when gravity dominates and the gravity model and $q$ are reliable. \\
\makecell[l]{Potential cancellation\\ \cite{pustina2025analysis,stolzle2025phdthesis}} &
Setpoint Regulation &
$\left[G_\varphi(q) + K_\varphi(q) \right]_\mathrm{a}$ &
\pidycompact &
Integrable actuation map; stable zero dynamics; accurate potential-force model; reliable $q$ estimate. &
Underactuated regulation when potential forces are only locally convex and the potential model and $q$ are reliable. \\
\makecell[l]{Feedforward compensation\\ \cite{pustina2024input,pustina2025analysis}} &
Trajectory Tracking &
\makecell[l]{$\left[M_\varphi(q^\mathrm{d})\ddot{\varphi}^\mathrm{d}+C_\varphi(q^\mathrm{d},\dot{q}^\mathrm{d})\dot{\varphi}^\mathrm{d}\right.$\\$\left.+G_\varphi(q^\mathrm{d})+K_\varphi(q^\mathrm{d})+D_\varphi(q^\mathrm{d})\dot{\varphi}^\mathrm{d}\right]_\mathrm{a}$} &
\pidycompact &
Integrable actuation map; stable zero dynamics; feasible reference. &
Underactuated tracking when an integrable actuation map and feasible trajectory are available. \\
\makecell[l]{Mixed-state feedback\\ \cite{pustina2024input,pustina2025analysis}} &
Trajectory Tracking &
\makecell[l]{$\left[M_\varphi(q)\ddot{\varphi}^\mathrm{d}+C_\varphi(q,\dot{q})\dot{\varphi}^\mathrm{d}\right.$\\$\left.+G_\varphi(q)+K_\varphi(q^\mathrm{d})+D_\varphi(q)\dot{\varphi}^\mathrm{d}\right]_\mathrm{a}$} &
\pidycompact &
Integrable actuation map; stable zero dynamics; feasible reference; Coriolis consistency; accurate dynamical model; reliable $q$ estimate. &
Underactuated tracking when current-state gravity/dynamics should be compensated. \\
\addlinespace[0.12em]
\multicolumn{6}{@{}l}{\textsc{Operational-space controllers}} \\
\cmidrule(lr){1-6}
\makecell[l]{Synergistic\\control~\cite{della2019exact,della2023model}} &
Setpoint Regulation &
$P_{M,A}=(JM^{-1}A)^{-1}JM^{-1}$ &
\makecell[l]{$P_{M,A}J^\top \operatorname{PID}_\mathrm{x}$} &
Task and actuator dimensions match; $JM^{-1}A$ full rank; stable internal dynamics, quasi-static objective. &
Underactuated task regulation when no integrable actuation map is available but task and actuator dimensions match. \\
\makecell[l]{Impedance control\\ \cite{khatib2003unified,della2020model,stolzle2024guiding}} &
Trajectory Tracking &
$A^{-1}(q)\!\left(J^\top F_\mathrm{x}^\mathrm{d} +G(q)\right)$ & &
Full actuation; $A(q)$ invertible; task Jacobian full row rank; stable null dynamics. &
Fully actuated task-space tracking or interaction with accurate full model and reliable $q,\dot{q}$. \\
\bottomrule
\end{tabularx}
\end{adjustbox}
\end{table*}

\section{Rendering and Visualization}\label{sec:app:rendering}
\begin{figure*}[t]
    \centering
    \includegraphics[width=\textwidth]{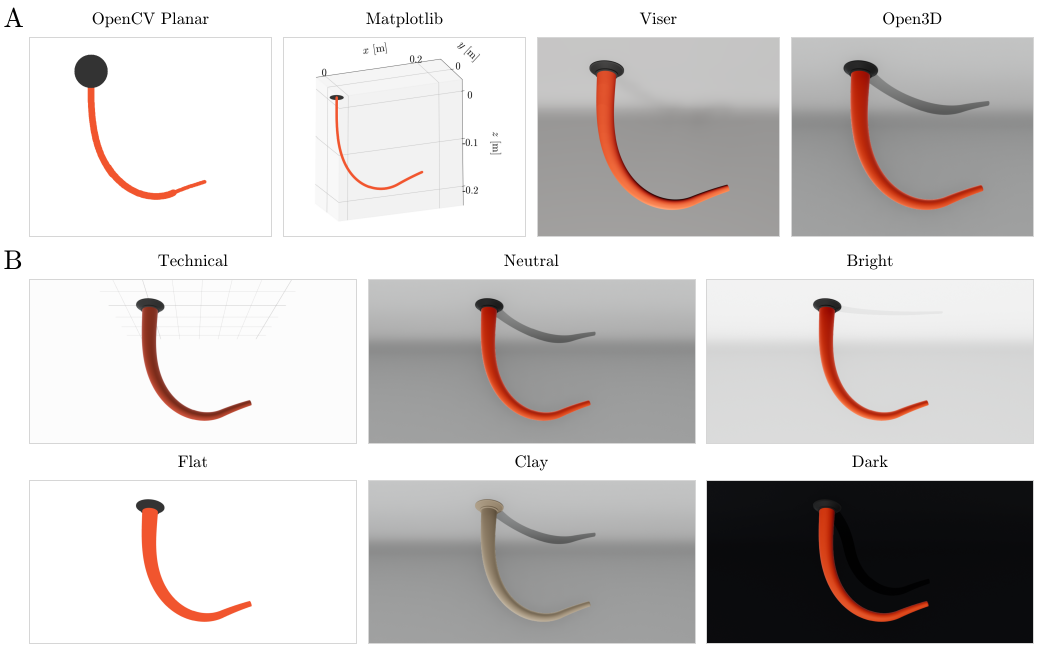}
    \caption{\textbf{Rendering backends and visual presets in \ac{SoRoMoX}.} (\textbf{A}) Example renderings of a hanging soft arm using OpenCV Planar, Matplotlib, Viser, and Open3D. (\textbf{B}) The technical, neutral, bright, flat, clay, and dark presets provide polished visual styles out of the box through coordinated lighting, materials, and backgrounds. The robot configuration is held fixed across presets to highlight the appearance changes.}
    \label{fig:renderers_and_presets}
\end{figure*}

\noindent\textbf{Geometry reconstruction.}
The rendering layer samples the robot's forward kinematics to obtain backbone positions and material frames. Open3D and Viser use these frames and the model's cross-section geometry to display discrete markers or construct continuous surfaces between sampled cross sections. Circular, elliptical, and rectangular sections can have dimensions that vary along the backbone, allowing tapered and noncircular bodies to be visualized. Supported actuator paths show routing alongside the body. The numbers of backbone and contour samples control visual detail independently of the model discretization.

\noindent\textbf{Rendering backends.}
\Cref{fig:renderers_and_presets}A compares the four backends. All render configurations and trajectories; their display environments and strengths differ:
\begin{center}
\small
\setlength{\tabcolsep}{3pt}
\begin{tabularx}{\columnwidth}{@{}l>{\raggedright\arraybackslash}X@{}}
\toprule
\textbf{Backend} & \textbf{Visualization approach} \\
\midrule
OpenCV & Planar drawing on an image canvas for lightweight animations and video generation \\
Matplotlib & Planar and spatial plots with axes, integrated into figures and notebooks \\
Open3D~\cite{zhou2018open3d} & Spatial mesh scenes in a desktop viewer, with offscreen image and video rendering \\
Viser~\cite{yi2025viser} & Spatial mesh scenes in a web browser, with interactive controls and live state streaming \\
\bottomrule
\end{tabularx}
\end{center}
Shared settings cover camera, colors, geometry, scene, and output. Six editable presets provide polished visual styles out of the box for Open3D and Viser, with coordinated lighting, materials, and backgrounds (\Cref{fig:renderers_and_presets}B). Appearance varies by backend and viewing mode; Matplotlib and OpenCV use white backgrounds without scene lighting.

\begin{samepage}
\noindent\textbf{Visualizing robot motion.}
Individual configurations support shape inspection, saved trajectories support playback, and Viser additionally accepts live configuration updates. Configurations may come from simulations, prescribed motions, or measured or estimated states of a physical robot. Matplotlib, Open3D, and Viser can display multiple configurations or trajectories together, enabling comparisons between motions, including predicted and observed behavior. Consistent camera settings and distinct robot colors aid these comparisons, while image and video export provide records for analysis and presentation.

\end{samepage}

\section{Application Case Studies}\label{sec:usecases}
\subsection{Static-Equilibrium System Identification}
Here, the Young's modulus, Poisson's ratio, and mass density are identified from four-marker measurements collected at static equilibrium, as even slight variations in stiffness or gravity can substantially change the resulting configurations.

In this case, we consider a continuum soft robot with a conical shape, characterized by a base radius of $r_\mathrm{base} = \SI{15.4}{mm}$, a tip radius of $r_\mathrm{tip} = \SI{4.8}{mm}$, and a total length of $L = \SI{360}{mm}$. Two tendons, routed straight along the robot up to $\SI{305}{mm}$ from the base, are displaced at angles of $\SI{30}{\degree}$ and $\SI{150}{\degree}$ from the horizontal axis of the cross-section. By actuating each tendon independently, the robot can reach points on a spherical surface patch covering nearly half a hemisphere. 

Due to the actuation discontinuity, the robot is decomposed into two distinct links, each modeled using the \ac{GVS} formulation with a \ac{VS} strain parametrization. Specifically, the first link is characterized by $7$ \acp{DOF}, enabling linear bending, constant torsion, and constant axial deformation, while the second is restricted to $2$ \acp{DOF}, allowing only constant bending along each axis. The resulting $9$-\acp{DOF} full-body dynamics is governed by Eq.~\ref{eq:system_dynamics}.
Here, the stiffness matrix $S$ is a function of the Young's modulus $E$ and Poisson's ratio $\nu$, which describe the material’s resistance to deformation. On the other side, the resulting gravitational force $G(q)$ is directly influenced by the mass density $\rho$. Variations in these material parameters therefore modify both the elastic and gravitational contributions to the model, ultimately affecting the configurations that the robot can attain at steady state. 

Building on this information, the physical parameters can be identified by minimizing the discrepancy between the robot's measured equilibrium configurations and the corresponding model predictions. Four markers are attached to its body as detailed in Table~\ref{tab:markers}, each providing three-dimensional Cartesian coordinates obtained through a motion capture system.
\begin{table}[t]
    \centering
    \caption{Positions of markers on the soft robot for static-equilibrium system identification.}
    \label{tab:markers}
    \begingroup
    \setlength{\tabcolsep}{4pt} 
    \begin{tabular}{ccc}
        \toprule
        \textbf{Marker ID} & \textbf{Distance from base} & \textbf{Distances from backbone} \\
        & $d_{i,x}$~[mm] & $(d_{i,y}, d_{i,z})$~[mm] \\
        \midrule
        1 & $129.1$ & $(0, 25)$  \\
        2 & $219.5$ & $(0, -21)$ \\
        3 & $280.0$ & $(0, 20)$  \\
        4 & $368.0$ & $(0, 0)$   \\
        \bottomrule
    \end{tabular}
    \endgroup
\end{table}

Let $\theta = \left[ E, \nu, \rho \right]$ denote the vector of optimization variables. The optimal $\theta^\star$ that best fits the static-equilibrium marker data can be computed by solving the following optimization problem:
\begin{equation}
\begin{aligned}
\min_{\theta} \quad & \sum_{i=1}^{n_\mathrm{e}}\sum_{j=1}^{n_\mathrm{m}}\|p_{ij}(q_i)-p_{ij}^\mathrm{meas}\|_2^2 \\
\text{s.t.} \quad & G(q_i,\rho) + S(E,\nu)\,q_i = A(q_i) \, u_i , \quad i = 1, \dots, n_\mathrm{e} \\
                 & E_\mathrm{min} \leq E \leq E_\mathrm{max}\\
                 & \nu_\mathrm{min} \leq \nu \leq \nu_\mathrm{max}\\
                 & \rho_\mathrm{min} \leq \rho \leq \rho_\mathrm{max}
\end{aligned}
\label{eq:sysid_optimization}
\end{equation}
where $n_\mathrm{e}$ is the number of experiments, $n_\mathrm{m}$ the number of markers, $p_{ij}^\mathrm{meas}$ is the measured position of the $j$-th marker in the $i$-th experiment, and $p_{ij}$ the corresponding expected value according to the model. Here, the first constraint enforces the static equilibrium condition that $q(\theta^\star)$ has to satisfy. From an implementation standpoint, this nonlinear constraint is embedded directly into the cost function evaluation when computing $p_{ij}(q_i)$. As a result, the lower and upper bounds on the optimization variables are the only explicit constraints that need to be provided to the solver. The bounds for the Young's modulus and Poisson's ratio are selected based on typical values reported in the literature for silicone-based materials, with $E_\mathrm{min}=\SI{0.01}{MPa}$, $E_\mathrm{max}=\SI{1}{MPa}$, $\nu_\mathrm{min}=\num{0.4}$, and $\nu_\mathrm{max}=\num{0.5}$.
The considered manipulator has a volume of \SI{1.416e-4}{m^3} and a mass of \SI{0.185}{kg}, yielding a theoretical material density of $\rho = m_\mathrm{robot}/V_\mathrm{robot} = \SI{1310}{kg \per m^3}$. However, material inhomogeneities and manufacturing imperfections cause the actual value to deviate from this estimate. For this reason, the mass density is also included as an optimization variable, with the theoretical value used solely to define its physical bounds, $\rho_\mathrm{min}=\SI{1000}{kg \per m^3}$ and $\rho_\mathrm{max}=\SI{2000}{kg \per m^3}$.
Finally, the initial condition $\theta_0$ was assigned as the mean value of the lower and upper bounds of each physical parameter, i.e. $\theta_0=\left[\SI{0.505}{MPa},\,0.45,\,\SI{1500}{kg \per m^3} \right]$.

The introduced optimization problem can be solved using a gradient-based approach by exploiting the fully differentiable structure of the SoRoMoX implementation in JAX. By leveraging \ac{AD}, the gradient of the loss function with respect to the physical parameters can be computed efficiently and exactly. This enables the use of first-order optimization algorithms, such as Adam, which iteratively update the parameter vector based on gradient information, leading to an efficient solution of the nonlinear optimization problem.

Using the Adam algorithm provided by the \emph{optimistix} package~\cite{optimistix2024} with a learning rate of $0.01$, the optimizer converges to the optimal solution $\theta^\star=\left[ \SI{0.315}{MPa},\,0.45,\, \SI{1320}{kg \per m^3} \right]$.

Fig.~\ref{fig:benchmark_sysid}B illustrates how the mathematical model of the soft robot becomes significantly more accurate at static equilibrium with the identified parameters,
reducing the overall static-equilibrium RMS marker error from $56.9$~mm to $19.3$~mm, yielding an improvement of about \SI{66}{\percent}.
Specifically, although the error on the first two markers slightly increases, the accuracy on the last two is drastically improved, resulting in a more homogeneous and accurate fit across all markers.
\subsection{Static-Equilibrium Identification and Residual Learning}
Even after fitting physical parameters to equilibrium data, residual modeling errors persist, as anticipated in Fig.~\ref{fig:benchmark_sysid}B. Rather than discarding these discrepancies, they can be represented by a residual external force $\tau_\mathrm{ext}$ \cite{gao2024sim}, which absorbs unmodeled effects and parametric uncertainty. Under this assumption, the system dynamics of Eq.~\ref{eq:system_dynamics} can be reformulated as:
\begin{equation}
    M(q) \, \Ddot{q} + C(q,\dot{q}) \, \dot{q} + G(q) + K(q) + D \, \dot{q} = A(q) \, u + \tau_\mathrm{ext}.
    \label{eq:system_dynamics_tauext}
\end{equation}
This residual external force can be estimated through an optimization procedure analogous to static-equilibrium system identification. However, this approach yields a fixed vector that struggles to minimize the error across the entire workspace. To improve generalizability, we instead model the external force as a function of the applied control input, since neglected manipulator effects may vary with the tendon pulling forces. Solving the resulting optimization problem for each control input $u_i$ produces a discrete dataset of optimal estimates $\tau_{\mathrm{ext},i}^\star(u_i)$. To recover the continuous representation required for the system dynamics, \cite{gao2024sim} proposed training a neural network on this dataset to learn a smooth mapping $\tau_{\mathrm{ext,NN}}(u)$.

Following this direction, the $i$-th optimal estimate $\tau_{\mathrm{ext},i}^\star$ can be obtained by solving the following optimization problem:
\begin{equation}
\begin{aligned}
\min_{\tau_{\mathrm{ext},i}} \quad & \sum_{j=1}^{n_\mathrm{m}}\|p_{ij}(q_i)-p_{ij}^\mathrm{meas}\|_2^2 + \lambda \|\tau_{\mathrm{ext},i}\|^2_2 \\
\text{s.t.} \quad & G(q_i) + S\,q_i = A(q_i) \, u_i + \tau_{\mathrm{ext},i} \\
                 & \underline{\tau}_{\mathrm{ext},i} \leq \tau_{\mathrm{ext},i} \leq \overline{\tau}_{\mathrm{ext},i}
\end{aligned}
\label{eq:sysid_optimization_resid}
\end{equation}
where the first constraint enforces static equilibrium under the applied external force $\tau_{\mathrm{ext},i}$, and the second ensures physically admissible configurations of the soft robot. From empirical considerations, force bounds of $\pm 1$~N and $\pm 0.01$~N are assigned to the first and second links, respectively.  

As in the static-equilibrium identification step, this problem is solved via a gradient-based approach that exploits the fully differentiable structure of the SoRoMoX implementation, allowing exact gradient computation through automatic differentiation and efficient optimization via the Adam algorithm. Choosing $\lambda = 10^{-8}$, Fig.~\ref{fig:benchmark_sysid}B,C illustrates how the mathematical model accuracy significantly improves thanks to the identified optimal residual forces $\tau_{\mathrm{ext}}^\star$. Fig.~\ref{fig:benchmark_sysid}D further shows that backbone configurations obtained with residual-force estimation closely match their measured counterparts across all inputs $u_i$, while pre-optimized results exhibit substantially larger errors.

To train the Neural Network, the resulting dataset of $n_\mathrm{e} = 22$ input-output pairs $\{(u_i,\, \tau_{\mathrm{ext},i}^\star)\}$ is split into a training set ($90\%$ corresponding to $M$ samples) and a test set ($10\%$). 
A fully-connected feedforward neural network with $4$ hidden layers of $512$ units each and ReLU activations learns the smooth mapping $\tau_{\mathrm{ext,NN}}(u)$ by minimizing the mean squared error between predicted and optimal generalized forces, regularized by an L2 penalty on the network weights $\Theta$:
\begin{equation}
    \mathcal{L}(\Theta) = \frac{1}{2M} \sum_{i=1}^{M} \|\tau_{\mathrm{ext,NN}}(u_i, \Theta) - \tau_{\mathrm{ext},i}^\star\|^2_2 + \frac{\gamma}{2}\|\Theta\|^2_2
\end{equation}
with $\gamma = 10^{-5}$. Training was performed for $12 \cdot 10^3$ steps using the Adam optimizer, with an initial learning rate of $10^{-3}$ decayed via a cosine schedule to $5\%$ of its initial value, and gradient clipping with unit norm threshold to stabilize convergence.

Fig.~\ref{fig:benchmark_sysid}B,C highlights how the trained neural network accurately approximates the optimal residual external force for both the training and testing dataset.

Overall, both the optimized residual and the learned residual improve the agreement between predicted and measured equilibrium shapes, with overall \ac{RMSE} errors of $\SI{6.6}{mm}$ and $\SI{6.9}{mm}$, respectively. This yields an improvement of about \SI{66}{\percent} and \SI{64}{\percent} relative to the model identified at static equilibrium.

\subsection{Model-Based Control}

The model-based control case study evaluates whether the decomposed dynamics and kinematics exposed by SoRoMoX can be used directly for feedback design.
It compares progressively richer model compensation in configuration space and demonstrates simultaneous end-effector position and orientation tracking in operational space.

The following simulation cases use the spatial \ac{PCS} strain convention
\begin{equation}
    \begin{aligned}
        \xi_i(q_i)
        &=
        \xi_\mathrm{ref}+q_i,\\
        \xi_i(q_i)
        &=
        \begin{bmatrix}
            \kappa_{\mathrm{x},i}&
            \kappa_{\mathrm{y},i}&
            \kappa_{\mathrm{z},i}&
            \sigma_{\mathrm{x},i}&
            \sigma_{\mathrm{y},i}&
            \sigma_{\mathrm{z},i}
        \end{bmatrix}^{\top},\\
        \xi_\mathrm{ref}&=
        \begin{bmatrix}
            0&0&0&1&0&0
        \end{bmatrix}^{\top},\\
        q_i
        &=
        \begin{bmatrix}
            \kappa_{\mathrm{x},i}&
            \kappa_{\mathrm{y},i}&
            \kappa_{\mathrm{z},i}&
            \Delta\sigma_{\mathrm{x},i}&
            \sigma_{\mathrm{y},i}&
            \sigma_{\mathrm{z},i}
        \end{bmatrix}^{\top}.
    \end{aligned}
    \label{eq:app:model_based_control_strain}
\end{equation}
Thus, $\kappa_{\mathrm{x},i}$ denotes torsion, $\kappa_{\mathrm{y},i}$ and $\kappa_{\mathrm{z},i}$ bending, $\Delta\sigma_{\mathrm{x},i}=\sigma_{\mathrm{x},i}-1$ the axial-strain displacement from the reference value, and $\sigma_{\mathrm{y},i}$ and $\sigma_{\mathrm{z},i}$ transverse shear. The full axial strain is therefore $\sigma_{\mathrm{x},i}=1+\Delta\sigma_{\mathrm{x},i}$.
In both cases, all six generalized strain coordinates are active in every segment; no coordinate is deactivated.
All segments also share the geometry and material parameters $L_i=\SI{0.1}{m}$, $r_i=\SI{0.02}{m}$, $E_i=\SI{2}{kPa}$, $G_{\mathrm{s},i}=\SI{1}{kPa}$, and $\rho_i=\SI{1070}{kg \per m^3}$.

\emph{Shared implementation.}
Both cases assemble their controllers directly from the exposed model terms $M$, $C$, $G$, $K$, $D$, $A$, $g$, $J$, and $\dot{J}$, without model-specific dynamics code.
Desired velocities and accelerations are generated from the continuous reference maps by \ac{AD} and the built-in closed-loop rollout jointly integrates the controller and robot dynamics.

\subsubsection{Configuration-Space Regulation and Tracking}\label{sec:app:model_based_control_configuration}

The configuration-space study isolates the effect of increasingly complete model compensation under identical feedback, reference, and rollout conditions.
The control objective is to drive the generalized strain-displacement configuration $q(t)$ to $q^\mathrm{d}(t)$ during one uninterrupted benchmark comprising stepwise setpoint regulation and time-varying trajectory tracking.
The regulation phase tests equilibrium accuracy, whereas the tracking phase additionally tests compensation of the velocity- and acceleration-dependent dynamics.
The resulting commanded-coordinate responses for all five controllers are shown in \Cref{fig:model_based_control}A.
We consider a horizontally mounted, fully actuated, one-segment spatial \ac{PCS} robot with $q\in\mathbb{R}^{6}$, $A(q)=I_6$, and $q(0)=\dot{q}(0)=0$.
Only $\kappa_\mathrm{y}$, $\kappa_\mathrm{z}$, and $\Delta\sigma_\mathrm{x}$ receive nonzero references, whereas $\kappa_\mathrm{x}$, $\sigma_\mathrm{y}$, and $\sigma_\mathrm{z}$ remain active and are regulated to zero.
The integration and saved-sample periods are $\delta t=\SI{5e-5}{s}$ and $\delta t_\mathrm{s}=\SI{0.01}{s}$, respectively.

Specifically, setpoint regulation is performed for $t\in[0,\SI{15}{s})$, followed by trajectory tracking for $t\in[\SI{15}{s},\SI{30}{s}]$.
During regulation, the desired physical strains $(\kappa_\mathrm{y}^\mathrm{d},\kappa_\mathrm{z}^\mathrm{d},\sigma_\mathrm{x}^\mathrm{d})$ are
\begin{equation}
    \begin{cases}
        (0,0,1.00), & 0\leq t<\SI{3}{s},\\
        (14,7,1.10), & \SI{3}{s}\leq t<\SI{6}{s},\\
        (-14,-7,0.95), & \SI{6}{s}\leq t<\SI{9}{s},\\
        (7,-10,1.06), & \SI{9}{s}\leq t<\SI{12}{s},\\
        (0,0,1.00), & \SI{12}{s}\leq t<\SI{15}{s},
    \end{cases}
    \label{eq:app:configuration_regulation_reference}
\end{equation}
where the bending strains are expressed in $\si{m^{-1}}$ and the axial strain is dimensionless.
For the tracking phase, let $\tau=t-\SI{15}{s}$,
\begin{equation}
    s(\tau)=\operatorname{clip}\left(\frac{\tau}{\SI{1.5}{s}},0,1\right),
    \qquad
    h(s)=10s^3-15s^4+6s^5.
\end{equation}
The desired trajectory is
\begin{equation}
    \begin{aligned}
        \kappa_\mathrm{y}^\mathrm{d}(t)
        &=12h\bigl(s(\tau)\bigr)\sin\left(3\tau\right),\\
        \kappa_\mathrm{z}^\mathrm{d}(t)
        &=6h\bigl(s(\tau)\bigr)\sin\left(2\tau+\frac{\pi}{4}\right),\\
        \sigma_\mathrm{x}^\mathrm{d}(t)
        &=1+0.06h\bigl(s(\tau)\bigr)\sin\left(2.5\tau+\frac{\pi}{2}\right),
    \end{aligned}
    \label{eq:app:configuration_tracking_reference}
\end{equation}
with angular frequencies $3$, $2$, and $\SI{2.5}{rad \per s}$, respectively.

Proportional--derivative (PD) and proportional--integral--derivative (PID) feedback provide model-free baselines.
Potential compensation (PC), feedforward compensation (FF), and computed torque (CT) then introduce, respectively, static-force compensation, desired-trajectory inverse dynamics, and measured-state inverse-dynamics cancellation~\cite{kelly1994pd,della2020model,della2023model}.
All five implementations share the robot, reference, feedback gains, and rollout; only the model-compensation term changes.
Let $e_q=q^\mathrm{d}-q$. The integral state $\eta_q$ evolves according to the unit-preserving saturation
\begin{equation}
    \begin{aligned}
        \dot{\eta}_q
        &=\operatorname{sat}_{\Gamma}(e_q),
        \qquad \eta_q(0)=0,\\
        \operatorname{sat}_{\Gamma}(e)
        &=\Gamma^{-1}\tanh\!\left(\Gamma e\right),
    \end{aligned}
    \label{eq:app:configuration_integral_saturation}
\end{equation}
where $\tanh(\cdot)$ acts componentwise and $\Gamma=\operatorname{diag}(e_\mathrm{sat})^{-1}$. For the coordinate ordering in \eqref{eq:app:model_based_control_strain}, the physical error scales are
\begin{equation}
    e_\mathrm{sat}
    =
    \begin{bmatrix}
        10 & 10 & 10 & 0.1 & 0.1 & 0.1
    \end{bmatrix}^{\top},
    \label{eq:app:configuration_integral_scales}
\end{equation}
where the first three entries are in \si{m^{-1}} and the final three are dimensionless. The rotational scale corresponds to one radian of angular error across the \SI{0.1}{m} segment, and the linear scale corresponds to a \SI{10}{\percent} extension or shear-strain error. This saturation preserves the units of the integrated error and has unit slope at the origin, leaving the nominal small-error integral gain unchanged.
The shared feedback term is
\begin{equation}
    \operatorname{PID}_{q}
    =
    K_\mathrm{p}e_q
    +K_\mathrm{i}\eta_q
    +K_\mathrm{d}(\dot{q}^\mathrm{d}-\dot{q}),
    \label{eq:app:configuration_pid}
\end{equation}
The controller $\operatorname{PD}_{q}$ follows by setting $K_\mathrm{i}=0$.
For $A(q)=I_6$, the five control laws are
\begin{equation}
    \begin{aligned}
        u_\mathrm{PD}
        &=
        \operatorname{PD}_{q},\\
        u_\mathrm{PID}
        &=
        \operatorname{PID}_{q},\\
        u_\mathrm{PC}
        &=
        G(q^\mathrm{d})+K(q^\mathrm{d})+\operatorname{PID}_{q},\\
        u_\mathrm{FF}
        &=
        M(q^\mathrm{d})\ddot{q}^\mathrm{d}
        +C(q^\mathrm{d},\dot{q}^\mathrm{d})\dot{q}^\mathrm{d}\\
        &\quad
        +G(q^\mathrm{d})+K(q^\mathrm{d})
        +D\dot{q}^\mathrm{d}+\operatorname{PID}_{q},\\
        u_\mathrm{CT}
        &=
        M(q)\left(\ddot{q}^\mathrm{d}
        +\operatorname{PID}_{q}^\mathrm{CT}\right)\\
        &\quad
        +C(q,\dot{q})\dot{q}+G(q)+K(q)+D\dot{q},
    \end{aligned}
    \label{eq:app:configuration_controllers}
\end{equation}
where $\operatorname{PID}_{q}^\mathrm{CT}=M(0)^{-1}\operatorname{PID}_{q}$.
The matched force-domain gains are
\begin{equation}
    K_\mathrm{p}=\omega_n^2M(0),\quad
    K_\mathrm{d}=2\zeta\omega_nM(0),\quad
    K_\mathrm{i}=\omega_i\omega_n^2M(0).
    \label{eq:app:configuration_gains}
\end{equation}
with $\omega_n=\SI{30}{rad \per s}$, $\zeta=0.9$, and $\omega_i=\SI{0.75}{rad \per s}$.

Metrics are evaluated separately over the regulation and tracking phases.
Let $\mathcal{Q}=\{\kappa_\mathrm{x},\kappa_\mathrm{y},\kappa_\mathrm{z},\Delta\sigma_\mathrm{x},\sigma_\mathrm{y},\sigma_\mathrm{z}\}$ contain all active generalized strain coordinates. For phase $\mathcal{P}$, saved-sample set $\mathcal{I}_\mathcal{P}$, and $j\in\mathcal{Q}$, let $e_{q,j}[k]=q_j^\mathrm{d}[k]-q_j[k]$ and $N_\mathcal{P}=|\mathcal{I}_\mathcal{P}|$. The reported metric is
\begin{equation}
    \operatorname{RMSE}_{j}^{\mathcal{P}}
    =
    \sqrt{\frac{1}{N_\mathcal{P}}
    \sum_{k\in\mathcal{I}_\mathcal{P}}e_{q,j}^2[k]}.
    \label{eq:app:configuration_metrics}
\end{equation}
The metric remains coordinate-wise because angular and linear strain errors have different units. Reporting all six coordinates also exposes motion induced through dynamic coupling in the uncommanded torsion $\kappa_\mathrm{x}$ and shear coordinates $\sigma_\mathrm{y}$ and $\sigma_\mathrm{z}$, whose references remain zero throughout the experiment.
For setpoint regulation, we additionally quantify the terminal tracking error over the final $10\%$ of each of the four post-change constant-reference intervals $\mathcal{I}_{m}^{\mathrm{term}}$, $m\in\{1,\ldots,4\}$, corresponding to the final \SI{0.3}{s} of each \SI{3}{s} interval. The initial zero-reference interval is excluded. Weighting the four setpoints equally gives
\begin{equation}
    \operatorname{MAE}_{j}^{\mathrm{term}}
    =
    \frac{1}{4}\sum_{m=1}^{4}
    \frac{1}{N_m}\sum_{k\in\mathcal{I}_{m}^{\mathrm{term}}}
    \left|e_{q,j}[k]\right|,
    \qquad N_m=|\mathcal{I}_{m}^{\mathrm{term}}|.
    \label{eq:app:configuration_terminal_metrics}
\end{equation}
\Cref{tab:app:configuration_rmse} reports the full-phase regulation and trajectory-tracking \acp{RMSE} together with the terminal-regulation mean absolute errors.

\begin{table*}[t]
\centering
\scriptsize
\setlength{\tabcolsep}{2.2pt}
\renewcommand{\arraystretch}{1.12}
\caption{\textbf{Configuration-space regulation and tracking errors.}
The angular-strain errors $\kappa_\mathrm{x}$, $\kappa_\mathrm{y}$, and $\kappa_\mathrm{z}$ are reported in \si{rad \per m}; the axial-strain-displacement error $\Delta\sigma_\mathrm{x}$ and transverse-shear errors $\sigma_\mathrm{y}$ and $\sigma_\mathrm{z}$ are dimensionless. Only $\kappa_\mathrm{y}$, $\kappa_\mathrm{z}$, and $\Delta\sigma_\mathrm{x}$ receive nonzero references. The terminal-regulation mean absolute error uses the final $10\%$ of each post-change setpoint interval and excludes the initial zero-reference interval. Bold values indicate the lowest error for each coordinate within each block; $\approx0$ denotes an error below $10^{-12}$.}
\label{tab:app:configuration_rmse}
\begin{tabularx}{\textwidth}{@{}
>{\raggedright\arraybackslash}X
*{6}{>{\centering\arraybackslash}p{0.105\textwidth}}
@{}}
\toprule
\textbf{Controller}
& \makecell{\boldmath$\kappa_\mathrm{x}$\unboldmath\\$[\si{rad \per m}]$}
& \makecell{\boldmath$\kappa_\mathrm{y}$\unboldmath\\$[\si{rad \per m}]$}
& \makecell{\boldmath$\kappa_\mathrm{z}$\unboldmath\\$[\si{rad \per m}]$}
& \makecell{\boldmath$\Delta\sigma_\mathrm{x}$\unboldmath\\$[-]$}
& \makecell{\boldmath$\sigma_\mathrm{y}$\unboldmath\\$[-]$}
& \makecell{\boldmath$\sigma_\mathrm{z}$\unboldmath\\$[-]$} \\
\midrule
\multicolumn{7}{@{}l}{\textsc{Setpoint regulation---RMSE}} \\
\cmidrule(lr){1-7}
PD (model-free)
& 2.817 & 3.067 & 1.573
& $1.806\times10^{-2}$ & $3.704\times10^{-2}$ & $1.133\times10^{-1}$ \\
PID (model-free)
& 2.472 & 3.052 & 1.531
& $1.846\times10^{-2}$ & $3.606\times10^{-2}$ & $6.747\times10^{-2}$ \\
Potential compensation
& $2.097\times10^{-1}$ & 2.965 & 1.408
& $1.832\times10^{-2}$ & $2.111\times10^{-2}$ & $4.591\times10^{-2}$ \\
Feedforward compensation
& $2.097\times10^{-1}$ & 2.965 & 1.408
& $1.832\times10^{-2}$ & $2.111\times10^{-2}$ & $4.591\times10^{-2}$ \\
Computed torque
& \boldmath$\approx0$\unboldmath & \textbf{2.088} & \textbf{1.026}
& \boldmath$1.198\times10^{-2}$\unboldmath & \boldmath$\approx0$\unboldmath & \boldmath$\approx0$\unboldmath \\
\addlinespace[0.12em]
\multicolumn{7}{@{}l}{\textsc{Setpoint regulation---terminal-window MAE}} \\
\cmidrule(lr){1-7}
PD (model-free)
& 2.827 & 1.081 & 0.7853
& $5.828\times10^{-3}$ & $3.180\times10^{-2}$ & $1.058\times10^{-1}$ \\
PID (model-free)
& 0.9285 & 0.4410 & 0.3845
& $4.565\times10^{-3}$ & $1.446\times10^{-2}$ & $2.315\times10^{-2}$ \\
Potential compensation
& $4.579\times10^{-2}$ & 0.2532 & 0.1546
& $1.924\times10^{-3}$ & $3.524\times10^{-3}$ & $4.549\times10^{-3}$ \\
Feedforward compensation
& $4.579\times10^{-2}$ & 0.2532 & 0.1546
& $1.924\times10^{-3}$ & $3.524\times10^{-3}$ & $4.549\times10^{-3}$ \\
Computed torque
& \boldmath$\approx0$\unboldmath
& \boldmath$5.963\times10^{-2}$\unboldmath
& \boldmath$3.998\times10^{-2}$\unboldmath
& \boldmath$4.466\times10^{-4}$\unboldmath
& \boldmath$\approx0$\unboldmath
& \boldmath$\approx0$\unboldmath \\
\addlinespace[0.12em]
\multicolumn{7}{@{}l}{\textsc{Trajectory tracking---RMSE}} \\
\cmidrule(lr){1-7}
PD (model-free)
& 1.768 & 2.688 & 0.9485
& $2.501\times10^{-2}$ & $2.936\times10^{-2}$ & $1.204\times10^{-1}$ \\
PID (model-free)
& 2.233 & 2.603 & 0.9200
& $2.652\times10^{-2}$ & $3.357\times10^{-2}$ & $7.504\times10^{-2}$ \\
Potential compensation
& 0.1369 & 2.549 & 0.9085
& $9.478\times10^{-3}$ & $2.715\times10^{-2}$ & $7.290\times10^{-2}$ \\
Feedforward compensation
& $1.870\times10^{-2}$ & $1.395\times10^{-2}$ & $3.960\times10^{-2}$
& $2.309\times10^{-4}$ & $1.099\times10^{-3}$ & $1.959\times10^{-4}$ \\
Computed torque
& \boldmath$\approx0$\unboldmath
& \boldmath$5.309\times10^{-3}$\unboldmath
& \boldmath$8.326\times10^{-3}$\unboldmath
& \boldmath$4.881\times10^{-5}$\unboldmath
& \boldmath$\approx0$\unboldmath
& \boldmath$\approx0$\unboldmath \\
\bottomrule
\end{tabularx}
\end{table*}

The time histories in \Cref{fig:model_based_control}A and the quantitative errors in \Cref{tab:app:configuration_rmse} highlight three effects of model compensation. First, the full regulation-phase \acp{RMSE} include the transients following each setpoint change, whereas the terminal-window mean absolute errors isolate the residual tracking error. In these terminal windows, potential compensation reduces the errors in all six coordinates relative to both model-free controllers, with reductions across the two baselines and six coordinates ranging from \SI{43}{\percent} to \SI{98}{\percent}. Potential and feedforward compensation coincide in regulation because the desired velocities and accelerations vanish between setpoint changes. Second, during trajectory tracking, adding the desired-trajectory inertial, Coriolis, and damping terms in the feedforward controller reduces the \acp{RMSE} of all six coordinates by \SIrange{86.3}{99.7}{\percent} relative to potential compensation. Third, full feedback linearization with computed torque further reduces the tracking \acp{RMSE} of $\kappa_\mathrm{y}$, $\kappa_\mathrm{z}$, and $\Delta\sigma_\mathrm{x}$ by \SI{62}{\percent}, \SI{79}{\percent}, and \SI{79}{\percent}, respectively, relative to feedforward compensation, and suppresses the three uncommanded-coordinate errors to numerical precision. For terminal regulation, computed torque reduces the mean absolute errors of these three commanded coordinates by \SI{76}{\percent}, \SI{74}{\percent}, and \SI{77}{\percent}, respectively, relative to potential or feedforward compensation.

\subsubsection{Operational-Space Position and Orientation Tracking}\label{sec:app:model_based_control_operational}

The operational-space study evaluates simultaneous position and orientation tracking using the complete translational and rotational task-space dynamics.
The control objective is to track a spherical surface-following figure eight: the desired position traverses a period-$T=\SI{4}{s}$ figure eight on the sphere
$\mathcal{S}=\{p\in\mathbb{R}^{3}\mid\|p-c\|=r_\mathrm{s}\}$ of radius $r_\mathrm{s}=\SI{0.08}{m}$, while the first end-effector axis follows the outward surface normal.
This coupled pose task exercises all six coordinates of the $SE(3)$ operational-space formulation.
The operational-space case uses a fully actuated two-segment spatial \ac{PCS} robot with $q=[q_1^\top,q_2^\top]^\top\in\mathbb{R}^{12}$, $A(q)=I_{12}$, and $q(0)=\dot{q}(0)=0$.
The simulation interval is $t\in[0,\SI{10}{s}]$, with integration period $\delta t=\SI{2e-5}{s}$ and saved-sample period $\delta t_\mathrm{s}=\SI{0.01}{s}$.
The controlled output is the complete end-effector pose $x(t)=(R(t),p(t))\in SE(3)$ at $s=L_1+L_2$, with all angular and translational task coordinates selected.
The corresponding pose tracking, componentwise errors, and representative robot configurations are shown in \Cref{fig:model_based_control}B.

To construct the desired pose, let $(R_0,p_0)$ be the initial end-effector pose, let $e_1,e_2,e_3$ denote the canonical basis of $\mathbb{R}^3$, and define
\begin{equation}
    \begin{aligned}
        n_0&=R_0e_1,&
        t_1&=R_0e_2,\\
        t_2&=R_0e_3,&
        c&=p_0-r_\mathrm{s}n_0,
    \end{aligned}
    \label{eq:app:operational_surface_geometry}
\end{equation}
where $c$ is the sphere center.
With $\phi(t)=2\pi t/T$, define the figure-eight displacement in the tangent plane as
\begin{equation}
    p_\mathrm{tan}^\mathrm{d}(t)
    =
    \SI{0.08}{m}\sin\phi(t)\,t_1
    +\SI{0.04}{m}\sin\bigl(2\phi(t)\bigr)\,t_2.
\end{equation}
The desired position on the sphere is
\begin{equation}
    \begin{aligned}
        p^\mathrm{d}(t)
        ={}&
        c+r_\mathrm{s}\Big[
            \cos\left(
                \frac{\|p_\mathrm{tan}^\mathrm{d}(t)\|}{r_\mathrm{s}}
            \right)n_0\\
        &\quad
            +\operatorname{sinc}\left(
                \frac{\|p_\mathrm{tan}^\mathrm{d}(t)\|}{r_\mathrm{s}}
            \right)
            \frac{p_\mathrm{tan}^\mathrm{d}(t)}{r_\mathrm{s}}
        \Big],
    \end{aligned}
    \label{eq:app:operational_position_reference}
\end{equation}
where $\operatorname{sinc}(z)=\sin(z)/z$ for $z\neq0$ and $\operatorname{sinc}(0)=1$.
The surface-following reference frame is defined by
\begin{equation}
    \begin{aligned}
        n^\mathrm{d}(t)
        &=
        \frac{p^\mathrm{d}(t)-c}{r_\mathrm{s}},\\
        y^\mathrm{d}(t)
        &=
        \frac{\left(I-n^\mathrm{d}(t)n^\mathrm{d}(t)^\top\right)t_1}
        {\left\|\left(I-n^\mathrm{d}(t)n^\mathrm{d}(t)^\top\right)t_1\right\|},\\
        z^\mathrm{d}(t)
        &=
        n^\mathrm{d}(t)\times y^\mathrm{d}(t),\\
        R^\mathrm{d}(t)
        &=
        \begin{bmatrix}
            n^\mathrm{d}(t)&y^\mathrm{d}(t)&z^\mathrm{d}(t)
        \end{bmatrix}.
    \end{aligned}
    \label{eq:app:operational_orientation_reference}
\end{equation}
Consequently, the end-effector position follows the spherical figure eight while its first axis follows the outward surface normal.

The geometric pose and velocity errors are
\begin{equation}
    \begin{aligned}
        e_\mathrm{x}
        &=
        \begin{bmatrix}
            e_\mathrm{R}\\e_\mathrm{p}
        \end{bmatrix},&
        e_\mathrm{R}
        &=
        \log_{SO(3)}\left(R^\mathrm{d}R^\top\right)^\vee,\\
        e_\mathrm{p}
        &=p^\mathrm{d}-p,&
        \dot{e}_\mathrm{x}
        &=\dot{x}^\mathrm{d}-J\dot{q}.
    \end{aligned}
    \label{eq:app:operational_errors}
\end{equation}
Using $\Lambda$ and $J_\mathrm{M}^{+}$ from \eqref{eq:app:operational_space_map} and $N_\mathrm{x}=I-J_\mathrm{M}^{+}J$, the partial-feedback-linearization impedance controller is~\cite{della2020model,stolzle2025phdthesis}
\begin{equation}
    \begin{aligned}
        u
        &=
        J^\top F_\mathrm{x}^\mathrm{d}+G(q),\\
        F_\mathrm{x}^\mathrm{d}
        ={}&
        (J_\mathrm{M}^{+})^\top\left(K(q)+D\dot{q}\right)
        +C_\mathrm{x}N_\mathrm{x}\dot{q}
        +\Lambda\ddot{x}^\mathrm{d}\\
        &+K_\mathrm{x}e_\mathrm{x}
        +D_\mathrm{x}\dot{e}_\mathrm{x}.
    \end{aligned}
    \label{eq:app:operational_impedance_control}
\end{equation}
The term $C_\mathrm{x}N_\mathrm{x}\dot{q}$ cancels the Coriolis coupling induced by the null-space velocity.

The operational-space adapter derives $J$, $\Lambda$, $J_\mathrm{M}^{+}$, $C_\mathrm{x}$, and $N_\mathrm{x}$ from the robot model, including the rotational task dynamics; the pose-reference utilities evaluate twist-consistent geometric errors.

Let $\lambda_i=[\Lambda(0)]_{ii}$, let $\bar{\lambda}_\mathrm{p}$ be the mean of the three translational entries, and let $k_\mathrm{p}=\SI{1}{N \per m}$ denote each translational stiffness entry.
The common target natural frequency, rotational stiffnesses, and critically damped gains are
\begin{equation}
    \omega_n=\sqrt{\frac{k_\mathrm{p}}{\bar{\lambda}_\mathrm{p}}},
    \qquad
    k_{\mathrm{R},i}=\lambda_{\mathrm{R},i}\omega_n^2,
    \qquad
    d_i=2\sqrt{k_i\lambda_i}.
    \label{eq:app:operational_impedance_gains}
\end{equation}
Thus, the rotational and translational task coordinates have matched local modal bandwidths.

To quantify position and orientation separately without dependence on the selected coordinate axes, define the Euclidean position-error magnitude $\varepsilon_\mathrm{p}[k]=\|e_\mathrm{p}[k]\|_2$ and the geodesic orientation-error magnitude $\varepsilon_\mathrm{R}[k]=\|e_\mathrm{R}[k]\|_2$. The corresponding reference spans are the maximum pairwise Euclidean and geodesic distances along the desired trajectory,
\begin{equation}
    \begin{aligned}
        S_\mathrm{p}
        &=\max_{k,\ell}\|p^\mathrm{d}[k]-p^\mathrm{d}[\ell]\|_2,\\
        S_\mathrm{R}
        &=\max_{k,\ell}
        \left\|
        \log_{SO(3)}\left(R^\mathrm{d}[k]R^\mathrm{d}[\ell]^\top\right)^\vee
        \right\|_2.
    \end{aligned}
    \label{eq:app:operational_reference_spans}
\end{equation}
For $a\in\{\mathrm{p},\mathrm{R}\}$, the \ac{RMSE} and its range-normalized counterpart are
\begin{equation}
    \operatorname{RMSE}_{a}
    =
    \sqrt{\frac{1}{N}\sum_{k=1}^{N}\varepsilon_a^2[k]},
    \qquad
    \operatorname{NRMSE}_{a}
    =
    \frac{\operatorname{RMSE}_{a}}{S_a}.
    \label{eq:app:operational_metrics}
\end{equation}
For the desired trajectory, $S_\mathrm{p}=\SI{134.6}{mm}$ and $S_\mathrm{R}=\SI{114.6}{\degree}$. Across the complete \SI{10}{s} rollout, the position-error norm has $\operatorname{RMSE}_\mathrm{p}=\SI{2.84}{mm}$ and $\operatorname{NRMSE}_\mathrm{p}=\SI{2.11}{\percent}$. The corresponding geodesic orientation errors are $\operatorname{RMSE}_\mathrm{R}=\SI{2.29}{\degree}$ and $\operatorname{NRMSE}_\mathrm{R}=\SI{2.00}{\percent}$.

\subsection{Control Gain Optimization}
The dynamic performance of a closed-loop system is strongly dependent on the controller, whose parameters often require careful tuning in order to attain the desired transient characteristics.
However, manually selecting the control parameters can be non-trivial, and it may require extensive trial-and-error~\cite{khan2020best}.
This section exemplifies how the SoRoMoX package allows for automatically tuning the gains of model-based feedback controllers by gradient-based optimization of the closed-loop dynamics of soft robots.

The soft robot is parametrized as a single-segment \ac{PCS} and actuated by three linearly routed tendons aligned with its backbone.
Two cases are investigated for setpoint regulation: one using a model-based regulator in actuation coordinates and the other using a synergistic controller in operational space.
The two controllers are implemented and optimized for their respective cases, as detailed below.
Both systems are underactuated, since the number of actuators $m = 3$ is smaller than the number of degrees of freedom $n = 6$.

\subsubsection{Model-Based Controllers}
\paragraph{Potential Shaping Actuation Space Controller}
It has been demonstrated in literature~\cite{pustina2024input} that it is often possible to reformulate the robot dynamics in actuation coordinates $\varphi = [\varphi_\mathrm{a}^\top, \varphi_\mathrm{u}^\top]^\top \in \mathbb{R}^n$, where $\varphi_\mathrm{a} \in \mathbb{R}^{m}, \varphi_\mathrm{u} \in \mathbb{R}^{n - m}$ are denoted as actuated and unactuated coordinates, respectively.
If the integrability assumption is met, a transformation exists that enables the change of coordinates as $h_{\mathrm{a}}(\cdot) : \mathbb{R}^n \rightarrow \mathbb{R}^m$, such that $\partial h_{\mathrm{a}} / \partial q = A^T(q)$ \cite{pustina2024input}.
In the case of underactuation, to complete the change of coordinates, one must set the last $n - m$ variables, which are usually set as the remaining configuration variables, defining the transformation as~\cite{stolzle2025phdthesis}
\begin{equation}
    \varphi = h(q) = \begin{bmatrix} h_{\mathrm{a}}(q) \\ 0_{n - m} \end{bmatrix} + 
    \begin{bmatrix}
        0_{m \times m} & 0_{m \times (n - m)} \\
        0_{(n - m) \times m} & I_{n - m}
    \end{bmatrix} q.
\end{equation}
The SoRoMoX package natively implements the transformation of coordinates $h(\cdot) : \mathbb{R}^n \rightarrow \mathbb{R}^n$ from configuration to actuation space.

In this example, the desired actuated variables derive from the target configuration as $\varphi_\mathrm{a}^\mathrm{d} = h_{\mathrm{a}}(q^\mathrm{d}) \in \mathbb{R}^m$.
The model-based setpoint regulation controller with potential compensation in actuation coordinates is~\cite{pustina2025analysis}
\begin{equation}
    u(t) = A_{\varphi,\mathrm{a}}^{-1}(q^\mathrm{d}) \bigl( G_{\varphi,\mathrm{a}}(q^\mathrm{d}) + K_{\varphi,\mathrm{a}}(q^\mathrm{d}) \bigr) + \text{PID}_{\varphi_\mathrm{a}},
    \label{eq:ctrl_as}
\end{equation}
where $A_{\varphi,\mathrm{a}} \in \mathbb{R}^{m \times m}$, $G_{\varphi,\mathrm{a}} \in \mathbb{R}^m$, $K_{\varphi,\mathrm{a}} \in \mathbb{R}^m$ are the first $m$ rows of the actuation matrix, the gravity force vector, and the elastic force vector, respectively, of the actuation-space dynamics \cite{pustina2024input}. Moreover,
\begin{equation}
\begin{split}
    \mathrm{PID}_{\varphi_\mathrm{a}}= \,
    & K_\mathrm{p} (\varphi_\mathrm{a}^\mathrm{d} - \varphi_\mathrm{a}(t)) + K_\mathrm{i} \eta_{\varphi_\mathrm{a}} - K_\mathrm{d} \, \dot{\varphi}_\mathrm{a}(t)\\
    \end{split}
\end{equation}
is the feedback term in actuation space, with $\eta_{\varphi_\mathrm{a}}$ according to \ref{eq:app:configuration_integral_saturation}, and $K_\mathrm{p}, K_\mathrm{i}, K_\mathrm{d} \in \mathbb{R}^{m \times m}$ being the proportional, integral, and derivative gains of an integral-saturated PID, respectively.

\paragraph{Synergistic Operational Space Controller}
In this case, the target is defined as a position $x^\mathrm{d} \in \mathbb{R}^3$ that the tip of the soft arm is supposed to reach.
The tip position of the robot is defined as $x(t) = \operatorname{FK}(q(t)) \in \mathbb{R}^{n_\mathrm{x}}$, where $n_\mathrm{x} = 3$ is the dimension of the operational (or task) space and $\operatorname{FK}(\cdot) : \mathbb{R}^n \rightarrow \mathbb{R}^{n_\mathrm{x}}$ is the forward kinematics that maps the configuration to the tip position~\cite{della2023model}.

Established controllers in operational space developed for the fully actuated case~\cite{della2020model, stolzle2024guiding} can be transferred to the underactuated one by means of the \emph{synergistic projector}, firstly introduced in~\cite{della2019exact}, that maps forces from task space directly into the actuation space under the assumptions that (i) the actuation space has the same dimension of the task space, thus $n_\mathrm{x} = m$, and (ii) the matrix $J(q) M^{-1}(q) A(q)$ is full rank.
Then, the synergistic projector is defined as
\begin{equation}
    P_\mathrm{M,A}(q) = \left( J(q) M^{-1}(q) A(q) \right)^{-1} J(q) M^{-1}(q),
\end{equation}
where $J = \partial \operatorname{FK} / \partial q \in \mathbb{R}^{n_\mathrm{x} \times n}$ is the Jacobian of the forward kinematics, $M(q) \in \mathbb{R}^{n \times n}$ is the inertia matrix, and $A(q) \in \mathbb{R}^{n \times m}$ is the actuation matrix.
The resulting control law
\begin{equation}
    u(t) = P_\mathrm{M,A}(q(t)) \, J^\top(q(t)) \, F_\mathrm{x} (t),
    \label{eq:ctrl_os}
\end{equation}
generates a fully actuated dynamics in operational space, given a generic force $F_\mathrm{x} \in \mathbb{R}^{n_\mathrm{x}}$~\cite{della2023model}.
In this example, the target force is generated from a PID controller in task space
\begin{equation}
    \begin{split}
    F_\mathrm{x}(t) = \text{PID}_\mathrm{x} = \,
    & K_\mathrm{p} (x^\mathrm{d} - x(t)) + K_\mathrm{i} \eta_{\varphi_x} - K_\mathrm{d} \, \dot{x}(t) \\
\end{split}
\end{equation}
where $K_\mathrm{p}, K_\mathrm{i}, K_\mathrm{d} \in \mathbb{R}^{n_\mathrm{x} \times n_\mathrm{x}}$ are the proportional, integral, and derivative gains, respectively.

\subsubsection{Feedback Control Gain Optimization}
We automatically tune the gains of the PID feedback controllers by optimizing the closed-loop behavior using gradient descent, combining the robot models and controllers from the SoRoMoX package with the Optax~\cite{deepmind2020jax} optimization package.
In this approach, we retrieve the gradients of the cost function encoding the performance of the closed-loop system response with respect to the chosen parameters by performing a forward and backward pass on the closed-loop simulation.
Considering the diagonal terms of the control gain matrices $k_\mathrm{p} = \mathrm{diag}(K_\mathrm{p})$, $k_\mathrm{i} = \mathrm{diag}(K_\mathrm{i})$, and $k_\mathrm{d} = \mathrm{diag}(K_\mathrm{d})$, the optimization variables are defined as the vector $\phi = [k_\mathrm{p}^\top, k_\mathrm{i}^\top, k_\mathrm{d}^\top]^\top$ that is updated via gradient descent
\begin{equation}
    \phi^{(\mathrm{j} + 1)} = \phi^{(\mathrm{j})} - \alpha_\mathrm{opt}^{(\mathrm{j})}\nabla_{\phi} l^{(\mathrm{j})},
\end{equation}
where $\alpha_\mathrm{opt}$ is the adaptive learning step of the Yogi optimizer~\cite{zaheer2018adaptive}, $\mathrm{j}$ denotes the $j$th iteration of the optimization loop, and the cost function is defined as
\begin{equation}
    l = \frac{1}{T} \int_{0}^{T} e(t')^\top \, \hat{W} \, e(t') dt',
    \label{eq:loss_fn}
\end{equation}
where the error $e(t)$ and the scaling matrix $\hat{W}$ are defined as $\hat{W} =\operatorname{diag}(L^2,L^2,L^2,1,1,1), \, e(t) =q^\mathrm{d} - q(t)$ in the \emph{actuation-space control} case and $\hat{W} = L^{-2} I_3, \, e(t) = x^\mathrm{d} - x(t)$ in the \emph{operational-space control} case, $L$ is the length of segment, and $T$ is the simulation horizon.

In order to improve the robustness of the optimization and guard against ``getting stuck'' in local minima, we perform batch optimization for both cases.
A batch is generated by randomly sampling a set of $n_\mathrm{b}$ initial values of $\phi$ within valid intervals with uniform distribution, yielding the batched collection $\Phi = (\phi_1, ..., \phi_{n_\mathrm{b}})$.
We run $n_\mathrm{b}$ parallel optimizations, one for each $\phi_i \in \Phi$, associated with the corresponding loss.

\subsubsection{Results}
\Cref{fig:controlgainopt} summarizes the results for both gain-optimization cases. The left panels (\Cref{fig:controlgainopt}(a),(c)) show the loss over $100$ optimization iterations, while the right panels show the corresponding convergence of the regulated configuration variables (\Cref{fig:controlgainopt}(b)) and tip position (\Cref{fig:controlgainopt}(d)) toward their targets. The optimized gains reduce the loss by \SI{46}{\percent} for the \emph{actuation-space control} case and by \SI{98}{\percent} for the \emph{operational-space control} case, measured from the initial median value across the batch to the best optimized value.

Additional control-performance metrics provide a more detailed view of the resulting trade-offs. For the \emph{actuation-space control} case, the cumulative tracking error decreases to \SI{0.20}{rad \per m}, corresponding to a \SI{34}{\percent} reduction relative to the initial median value; the settling time, defined as the first time at which the response enters the \SI{5}{\percent} target band, improves by \SI{52}{\percent}; and the overshoot decreases by \SI{29}{\percent}. For the \emph{operational-space control} case, the cumulative task-space tracking error reaches \SI{3}{mm}, corresponding to a \SI{95}{\percent} reduction; the final settling time becomes \SI{900}{ms}, with an overshoot of \SI{4}{mm}.
The reduced batch spread, shown by the shaded regions, further suggests that the randomly initialized gain sets converge to similar best-performing solutions. This is supported by a batch-wise loss-variance reduction of \SI{99}{\percent} for both \emph{actuation-space control} and \emph{operational-space control} cases.

\subsection{High-Order Control Lyapunov and Barrier Function (HOCLF+HOCBF) Control}

\ac{SoRoMoX} enables the implementation of high-order CLF-CBF controllers for soft robotic systems by leveraging its fully differentiable dynamics.
The goal of this section is to demonstrate that safety constraints can be incorporated in a modular fashion while simultaneously driving the soft robot end-effector toward a desired target.

\subsubsection{Approach}

\paragraph{High-Order CLF-CBF Controller}

The soft robot dynamics are written in control-affine form as
\begin{equation}\label{eq:soft_robot_dynamics_control_affine}
    \dot{y} = f_0(y) + f_u(y) \, u,
\end{equation}
where
\[
y = [q^\top,\dot{q}^\top]^\top,
\]
denotes the robot state and $u$ is the vector of tendon-tension inputs.
The drift and input vector fields $f_0(y)$ and $f_u(y)$ are obtained directly from the differentiable \ac{SoRoMoX} forward dynamics.

To encourage end-effector tracking, we define the control Lyapunov function
\begin{equation}
    V(y)
    =
    \|x(q)-x^\mathrm{d}\|,
\end{equation}
where $x(q)=\operatorname{FK}(q)$ is the end-effector position and $x^\mathrm{d}$ is the desired target position.

\paragraph{Distance-Based Contact Barrier}

We discretize the robot backbone into sampled points indexed by $i$.
For each sampled robot point and spherical obstacle $\mathcal{O}_j$, we define the signed clearance
\begin{equation}
    d_{ij}(q)
    =
    \|p_i(q)-c_j\|
    -
    \left(r_{\mathrm{obs},j}+r_\mathrm{robot}\right),
\end{equation}
where $p_i(q)$ is the position of the $i$-th backbone point, $c_j$ is the obstacle center, and $r_{\mathrm{obs},j}$ and $r_\mathrm{robot}$ are the obstacle and robot radii, respectively.

The minimum pairwise clearance is converted into a force-like barrier quantity:
\begin{equation}\label{eq:force_like_barrier}
    b(y)
    =
    F_{\max}
    +
    k_\mathrm{c}
    \min_{i,j} d_{ij}(q),
\end{equation}
where $F_{\max}$ is the maximally acceptable contact force and $k_\mathrm{c}$ is the stiffness of a penetration-based linear-elastic contact model that we assume in this case.
Thus, satisfying
\[
b(y)\ge 0,
\]
corresponds to respecting the prescribed contact-force bound.

Since the barrier depends on the robot configuration $q$, while the soft robot dynamics are second-order, the resulting safety constraint has relative degree two with respect to the control input.
We therefore impose the high-order CBF condition
\begin{equation}\label{eq:hocbf_contact_force}
    \ddot{b}(y,u)
    +
    \alpha_1 \, \dot{b}(y)
    +
    \alpha_0 \, b(y)
    \ge 0,
\end{equation}
where $\alpha_0,\alpha_1>0$ are HOCBF gains.
\paragraph{Projection-based controller implementation}

At each control step, we first compute a task-stabilizing nominal control input
using the HOCLF condition. Specifically, the HOCLF admissible set is defined as
\begin{equation}
\mathcal{C}_{V}(y)
=
\left\{
u :
\ddot{V}(y,u)
+
\gamma_1 \dot{V}(y)
+
\gamma_0 V(y)
\le 0
\right\}.
\end{equation}
Since $\mathcal{C}_{V}(y)$ is generally not a singleton, the HOCLF condition
does not uniquely determine a control input. We therefore select the
HOCLF-based nominal input as the minimum-norm control input satisfying the
HOCLF condition, equivalently the Euclidean projection of the origin onto
$\mathcal{C}_{V}(y)$:
\begin{equation}
\label{eq:clf_nominal_control}
\bar u(y)
:=
u_{\mathrm{CLF}}(y)
=
\Pi_{\mathcal C_V(y)}(0)
=
\arg\min_{u\in \mathcal C_V(y)}
\frac{1}{2}\|u\|^2 .
\end{equation}
This nominal input promotes convergence toward the desired trajectory through
the HOCLF decrease condition, but does not by itself encode the hard
contact-force safety requirement.

Safety is then enforced by projecting the HOCLF-based nominal input onto the
affine halfspace induced by the HOCBF condition:
\begin{equation}
\label{eq:cbf_projection}
u^\star
=
\Pi_{\mathcal{C}_{b}(y)}
\left(
\bar{u}(y)
\right),
\end{equation}
where
\begin{equation}
\mathcal{C}_{b}(y)
=
\left\{
u :
\ddot{b}(y,u)
+
\alpha_1 \dot{b}(y)
+
\alpha_0 b(y)
\ge 0
\right\}.
\end{equation}

Here,
$\Pi_{\mathcal{C}}(\cdot)$ denotes the Euclidean projection onto the set
$\mathcal{C}$. For control-affine dynamics, both the HOCLF and HOCBF
conditions are affine in the control input and therefore define halfspaces in
the control space. Specifically, for a halfspace of the form
\begin{equation}
\mathcal{H}
=
\left\{
u :
w^\top u \ge \beta
\right\},
\end{equation}
the corresponding closed-form projection is
\begin{equation}
\Pi_{\mathcal{H}}(v)
=
v
+
\frac{
\max\!\left(
0,\,
\beta-w^\top v
\right)
}{
\|w\|^2
}
w.
\end{equation}
Similarly, for a halfspace of the form
\begin{equation}
\mathcal{H}
=
\left\{
u :
w^\top u \le \beta
\right\},
\end{equation}
the Euclidean projection is
\begin{equation}
\Pi_{\mathcal{H}}(v)
=
v
-
\frac{
\max\!\left(
0,\,
w^\top v-\beta
\right)
}{
\|w\|^2
}
w.
\end{equation}

The HOCLF-based nominal controller promotes convergence toward the desired
trajectory, while the HOCBF projection enforces the contact-force safety
requirement. This construction gives priority to safety: the final control
input $u^\star$ satisfies the HOCBF constraint whenever the HOCBF admissible
halfspace is nonempty, even if doing so requires modifying the nominal HOCLF
input $\bar u(y)$ and potentially sacrificing the HOCLF decrease condition.


\subsubsection{Simulation Setup}
The simulation was performed using a tendon-actuated \ac{PCS} model consisting of two constant-strain segments, each of length \SI{0.15}{m}, resulting in a total backbone length of \SI{0.30}{m}. Each segment has radius \SI{36}{mm}, density \SI{1070}{kg \per m^3}, Young's modulus \SI{20}{kPa}, and shear modulus \SI{20}{kPa}. Gravity acts along the positive $z$ direction. A diagonal strain damping matrix is used, with translational strain damping weighted by $1$ and rotational/shear strain damping weighted by $10^3$, scaled by the segment length.

The robot is actuated by six straight tendons, with three tendons attached to each segment. The tendon attachment points are placed around the segment cross-sections slightly inside the outer radius, at a radial offset of $r_\mathrm{robot}-\SI{5}{mm}$. The tendon routings are modeled as linear routings with zero slope along both local transverse directions. The initial condition is the straight reference configuration with zero strain displacement and zero velocity.

The goal position for the end effector is set to
\[
x^\mathrm{d} = [0.10,\;0.05,\;0.32]^\top \si{m}.
\]
The safety environment contains three spherical obstacles with centers
\[
[0.10,0.08,0.24]^\top, \:
[0.12,0.06,0.32]^\top, \:
[0.04,0.055,0.20]^\top \,\mathrm{m},
\]
each with radius \SI{20}{mm}. For evaluating the barrier function, the robot backbone is sampled at $20$ points uniformly along its length. The clearance is computed as the pairwise distance between each sampled robot point and each obstacle, minus the sum of the robot and obstacle radii. The contact force proxy is modeled as a linear penalty after penetration, with stiffness \SI{1000}{N \per m}, and the prescribed force safety limit is \SI{5}{N}.

Two closed-loop rollouts are compared: a safety-unaware HOCLF-only controller and the safety-constrained HOCLF-HOCBF controller~\cite{wong2025contact}. The controller is implemented as a closed-form sequence of half-space projections~\cite{wong2026closed}: the nominal zero input is first projected onto the HOCLF constraint and then projected onto the HOCBF constraint. The closed-loop simulations are run for \SI{8}{s} the \texttt{robot.rollout\_closed\_loop\_to} with the Tsitouras 5/4 Runge--Kutta integrator~\cite{tsitouras2011runge}. The solver step size and the data saving interval are both set to $10^{-3}\si{s}$.

\subsubsection{Results}
The results are shown in Fig.~\ref{fig:CBF_RL}. The proposed CLF-CBF controller drives the robot toward the goal while enforcing the prescribed force safety constraint. With the CBF, the goal distance decreases from approximately 0.11\,m to 0.03\,m over the 8\,s simulation horizon. Meanwhile, the maximum pairwise normal force increases as the robot approaches the obstacle but converges to the prescribed safety limit of 5\,N without exceeding it. In contrast, the controller without the CBF reaches a smaller goal distance but produces unsafe contact forces, with the maximum pairwise normal force rising above 30\,N. This comparison highlights the safety-performance tradeoff introduced by the CBF and confirms that the proposed controller preserves the force constraint over the extended rollout.

\subsection{Parallel Reinforcement Learning}

To evaluate the efficiency of SoRoMoX for parallel reinforcement learning, we construct a representative RL benchmark and compare it with PyElastica. We consider a 3D tracking task where a tendon-actuated soft continuum arm tracks a dynamically moving target.
The arm is modeled using a Cosserat-rod formulation in PyElastica and one \ac{PCS} segment in SoRoMoX, and is actuated by $m=4$ tendons.
The RL scenario setting is illustrated in \Cref{fig:CBF_RL}B. The end-effector position target moves along a hemispherical surface, following a smooth trajectory with a randomly varying direction and a bounded velocity.
We train a policy $\pi(a_t\mid o_t)$ with \ac{PPO}~\cite{schulman2017proximal} on this task. 
Here, the agent observes a vector
$
o_t = [x^\top, v^\top, u_t^\top, (x^\mathrm{d})^\top, (v^\mathrm{d})^\top]^\top \in \mathbb{R}^{16}
$,
where $x \in \mathbb{R}^{3}$ and $v \in \mathbb{R}^{3}$ denote the end-effector position and velocity, $u_t \in \mathbb{R}^{4}$ represents the current tendon-force input, and $x^\mathrm{d} \in \mathbb{R}^{3}, v^\mathrm{d} \in \mathbb{R}^{3}$ correspond to the desired target position and velocity.
The action space is defined as a four-dimensional continuous vector $a_t \in [-1,1]^4$, representing incremental changes in tendon forces. The applied tendon-force vector is updated as
$
u_{t+1} = \mathrm{clip}(u_t + \Delta u_{\max} a_t, 0, u_{\max})
$,
where $\Delta u_{\max}$ is the maximum input-increment scale and $u_{\max}$ is the maximum allowable tendon-force input.
%
The objective is to minimize both the positional tracking error and undesirable motion away from the target. Let
$e_\mathrm{x} = x - x^\mathrm{d}$, $\quad \varepsilon_\mathrm{x} = \|e_\mathrm{x}\|
$.
We define the relative velocity projected along the error direction as
$\dot{\varepsilon}_\mathrm{x} = (v - v^\mathrm{d})^\top \frac{e_\mathrm{x}}{\varepsilon_\mathrm{x} + \epsilon_\mathrm{RL}}$,
where $\epsilon_\mathrm{RL}>0$ is a small numerical regularization constant.
The reward function is given by
\begin{equation}
\mathcal{R} = -\left( \frac{\varepsilon_\mathrm{x}}{0.2} + \frac{\max(\dot{\varepsilon}_\mathrm{x}, 0)}{0.3} \right)
+ \mathbb{I}(\varepsilon_\mathrm{x} < 0.01)\, 1.5.
\end{equation}
The first term penalizes tracking error, while the second term penalizes motion that increases the distance to the target. A success bonus is added when the arm end is within \SI{0.01}{m} of the target. 

%
We evaluate SoRoMoX and PyElastica on a single NVIDIA Tesla A100 GPU and two Intel Xeon Gold 6448Y CPU cores. For all experiments, the total training budget is fixed at 1 million environment steps, and the episode length is 105. 
To ensure a fair comparison, all PPO training runs use the same rollout size of 12,800 samples and a fixed PPO minibatch size of 428. All other training hyperparameters are set to their default values as implemented in Stable Baselines3~\cite{raffin2021stable}. As shown earlier in \Cref{tab:comparison_speed}, PyElastica achieves lower simulation latency than SoRoMoX in the single-environment setting due to differences in the underlying simulation models. In our experimental setup, the evaluated PyElastica configuration used two CPU cores and two parallel environments. In contrast, SoRoMoX runs entirely on the GPU, enabling substantially larger levels of parallelization. We therefore evaluate SoRoMoX with $n_\mathrm{b} \in \{64, 128, 256, 512\}$ parallel environments on a single GPU.

The training performance and wall-clock efficiency are summarized in Fig.~\ref{fig:CBF_RL}C. 
We simulated a soft robotic arm with a length of \SI{250}{mm} and a radius of \SI{25}{mm} to track a moving target ball. Tracking success was defined as the end-effector being within a \SI{10}{mm} radius of the ball position. Across $n_\mathrm{b}=128$ parallel evaluation trajectories, the mean tracking error at the final evaluated step was \SI{6.19}{mm}, with a step-wise success  rate of 91.4\%.
Despite its slower single-environment simulation speed, SoRoMoX rapidly closes the performance gap as $n_\mathrm{b}$ increases and ultimately surpasses the PyElastica baseline in overall training efficiency. This advantage becomes increasingly pronounced at larger scales, where SoRoMoX converges faster and achieves higher rewards earlier in training. Notably, the configurations with $n_\mathrm{b}=256$ and $n_\mathrm{b}=512$ parallel environments achieve approximately 4$\times$ and 7$\times$ reductions in wall-clock training time, respectively, while maintaining comparable or better learning performance. These values reflect an end-to-end comparison that combines the different model formulations and reductions (one-segment PCS in SoRoMoX versus a DCM in PyElastica), JIT execution stacks, hardware, and degree of rollout batching; they do not isolate the contribution of GPU batching. We did not evaluate a sequential CPU SoRoMoX or SoRoSim baseline on this RL task. Nevertheless, the CPU rollout results in \Cref{tab:comparison_speed} and the GPU batch-scaling results in \Cref{fig:benchmarking:batch_simulation_scaling} suggest that comparison against a sequential reduced-order CPU baseline could yield a larger relative wall-clock advantage. These results highlight the importance of GPU-native parallelism for RL workloads, where the ability to efficiently scale the number of simultaneous environments can outweigh per-environment simulation speed.


\section{Best Practices for Soft Robot Model Implementation}\label{sec:app:best_practices}
This appendix expands on the implementation best practices summarized in Sidebars 2--4, covering efficient evaluation, singularity-safe expressions, and verification of mathematical correctness.

\subsection{Implementation Principles for Efficient Evaluation}\label{sec:app:computational_efficiency}

The computational cost of strain-based soft robot models is dominated by repeated evaluations of kinematics, Jacobians, and integral quantities along the robot body.
In \ac{SoRoMoX}, we therefore structure the implementation around the dependency pattern of the underlying mechanics: quantities are propagated serially from the proximal to the distal end of the robot, while independent quadrature-point contributions are evaluated in batch.
This avoids large symbolic expressions, exposes parallelism, and keeps the implementation compatible with \ac{JIT} compilation and \ac{AD}.

A first design choice is to use numerical spatial integration instead of deriving and evaluating closed-form symbolic expressions for the full dynamics.
This is particularly important for inertial, gravitational, and Coriolis/convective terms, whose symbolic expressions quickly become unwieldy even for simple soft robots.
At the same time, we do not use a single maximally general implementation for all models.
Although the \ac{VS} approximation~\cite{renda2020geometric, boyer2020dynamics, mathew2025reduced} is expressive enough to represent \ac{PCS} models~\cite{renda2018discrete}, we provide specialized implementations for planar \ac{PCS}, spatial \ac{PCS}, and \ac{GVS}.
The planar \ac{PCS} implementation uses $SE(2)$ operations and smaller matrix blocks, which are cheaper than their $SE(3)$ counterparts.
The spatial \ac{PCS} implementation exploits constant-strain closed-form expressions for segment poses, adjoint maps, tangent maps, and their derivatives.
The \ac{GVS} implementation is used when variable strain distributions are required, and it then pays the additional cost of integrating the strain field along each segment.

A second principle is to reduce the computation to active coordinates as early as possible.
Inactive strain components are not merely removed from the final dynamical matrices; instead, Jacobians and Jacobian derivatives are assembled directly with respect to the active generalized coordinates whenever possible.
As a result, a reduced parametrization such as \ac{PCC}~\cite{webster2010design} can be cheaper than a full \ac{PCS} parametrization even when both share the same implementation path.
In the \ac{GVS} model, this idea is combined with a fixed-shape runtime layout: heterogeneous segment descriptions are converted into padded arrays with fixed maximum numbers of degrees of freedom and quadrature points.
Active-coordinate maps, gather indices, and masks allow the state to remain compact while still presenting static array shapes to JAX, which is essential for efficient compilation and batching.
JAX \ac{GVS} dynamics assembly further groups ordered segments into bounded active-\ac{DOF} prefix buckets, so each reduction uses only the largest causal coordinate prefix needed within that bucket.

State-independent quantities are precomputed and cached.
For \ac{PCS}, this includes the quadrature grid, per-segment spatial mass matrices, and full and active stiffness and damping matrices.
For \ac{GVS}, the cached data additionally include length-scaled interior quadrature weights, local mass/stiffness/damping matrices at quadrature nodes, and strain-basis matrices evaluated at the quadrature and Magnus integration points.
This avoids repeatedly evaluating cross-section geometry, material matrices, and basis functions at every time step.

The remaining state-dependent quantities are propagated using scan loops.
Forward kinematics, body-frame Jacobians, and their time derivatives are advanced from segment to segment using \texttt{lax.scan}, matching the serial structure of an open kinematic chain.
Within each segment, independent quadrature contributions are evaluated with batched operations such as \texttt{vmap}.
Endpoint nodes are retained where useful for pose and Jacobian propagation, but omitted from integral assembly whenever their quadrature weights are zero.

Forward dynamics uses a fused assembly path.
Rather than computing the inertia matrix, gravitational force, and Coriolis matrix through separate public calls, the implementation evaluates shared kinematic quantities once and assembles the terms needed for the \ac{EOM} from \eqref{eq:system_dynamics} in a single routine.
In particular, the convective force $C(q,\dot{q})\dot{q}$ is evaluated directly, avoiding materialization of the full Coriolis matrix in the common forward-dynamics path.
Likewise, when the convective force requires only the product $\dot{J}(q,\dot{q},s)\dot{q}$, the kinematic recurrence propagates this contraction directly instead of constructing the complete Jacobian derivative; the full $\dot{J}$ remains available through the public kinematics interface.
Contracted $SE(3)$ Lie-algebra actions are evaluated directly through vector cross products rather than materializing the corresponding $6\times6$ small-adjoint matrix.
The acceleration is then obtained by solving the linear system for $\ddot{q}$ rather than explicitly forming $M(q)^{-1}$.

The implementation also avoids \ac{AD} when compact analytical expressions are available.
Analytical recursions are used for Jacobians, Jacobian time derivatives, constant-strain tangent-map derivatives, and convective-force terms.
When several related constant-strain quantities are required at the same point, the adjoint and inverse adjoint, tangent map, and tangent derivative are assembled as a bundle that shares the powers of the Lie-algebra matrix and the stabilized scalar coefficients.
Single-quantity queries retain separate paths so that this fusion does not require constructing unused operators.
At the same time, cheaper paths are kept for queries that do not require all quantities: for example, inertia and gravity can be assembled from pose/Jacobian data, while Jacobian-derivative quantities are computed only when needed for convective terms or explicit Jacobian time derivatives.

Analytical matrix-free directional products directly evaluate \(J(q,s)v\), the tangent pushforward from generalized velocity to Cartesian twist, and \(\sum_s J(q,s)^\top w_s\), the cotangent pullback from Cartesian wrenches \(w_s\) to generalized forces.

Finally, because these optimizations can otherwise lead to duplicated code, shared helper functions are used for the recurring computational kernels: segment propagation, constant-strain step terms, \ac{GVS} Magnus updates, Jacobian/Jacobian-derivative propagation, and individual dynamics integrands.
This keeps the fused implementation maintainable while preserving the performance benefits of avoiding redundant kinematic and dynamic evaluations.

\subsection{Well-Defined Evaluation of Singular Closed-Form Expressions}\label{sec:app:singularities}
It is well known in the soft robotics literature that many rod- and strain-based models exhibit apparent singularities near the straight or zero-bending configuration~\cite{godage2016dynamics, deutschmann2020modeling, della2020improved, caasenbrood2024design, stolzle2025phdthesis}.
For \ac{PCS} and \ac{GVS} models, these singularities typically arise in closed-form Lie-group and constant-strain expressions, where trigonometric coefficients contain terms such as
\begin{equation}
    \frac{\sin \vartheta}{\vartheta}, \qquad
    \frac{1-\cos \vartheta}{\vartheta^2}, \qquad
    \frac{\vartheta-\sin \vartheta}{\vartheta^3},
\end{equation}
or higher-order analogues in tangent maps and their derivatives.
Here, $\vartheta$ denotes the magnitude of the rotational part of the Lie-algebra argument.
For a constant strain
\begin{equation}
    \xi =
    \begin{bmatrix}
        \xi_{\mathrm{rot}} \\
        \xi_{\mathrm{trans}}
    \end{bmatrix}
    \in \mathbb{R}^6,
    \qquad
    \xi_{\mathrm{rot}},\xi_{\mathrm{trans}}\in\mathbb{R}^3,
\end{equation}
propagated along a segment as
\begin{equation}
    g(s) = g_0 \exp_{SE(3)}(s\,\xi),
\end{equation}
the relevant scalar is
\begin{equation}
    \vartheta = \|(s\,\xi)_{\mathrm{rot}}\|
           = |s|\,\|\xi_{\mathrm{rot}}\|.
\end{equation}
Equivalently, using the matrix representation of the twist,
\begin{equation}
    \exp_{SE(3)}(s\,\xi) = \exp(s\,\widehat{\xi}),
\end{equation}
where the widehat operator maps the vector $\xi\in\mathbb{R}^6$ to its corresponding matrix in the Lie algebra $\mathfrak{se}(3)$.
Thus, the same apparent singularity is reached either when the local rotational strain tends to zero, $\xi_{\mathrm{rot}}\rightarrow 0$, or when the arclength argument tends to zero, $s\rightarrow 0$.

These singularities are removable at the level of the underlying maps.
For example,
\begin{equation}
    \lim_{s\rightarrow 0} \exp_{SE(3)}(s\,\xi) = I_4,
\end{equation}
and, for fixed $s$ and translational strain $\xi_{\mathrm{trans}}$,
\begin{equation}
    \lim_{\xi_{\mathrm{rot}}\rightarrow 0}
    \exp_{SE(3)}(s\,\xi)
    =
    \begin{bmatrix}
        I_3 & s\,\xi_{\mathrm{trans}} \\
        0 & 1
    \end{bmatrix}.
\end{equation}
Similarly, the derivative with respect to the arclength parameter is well defined:
\begin{equation}
    \frac{\partial}{\partial s}\exp(s\,\widehat{\xi})
    =
    \exp(s\,\widehat{\xi})\,\widehat{\xi},
\end{equation}
and therefore
\begin{equation}
    \left.
    \frac{\partial}{\partial s}\exp(s\,\widehat{\xi})
    \right|_{s=0}
    =
    \widehat{\xi}.
\end{equation}
For the full pose $g(s)=g_0\exp(s\,\widehat{\xi})$, this gives $g'(0)=g_0\widehat{\xi}$.
At the mathematical complete-strain limit $\xi=0$, which is distinct from the zero-displacement configuration $q=0$ whenever $\xi_\mathrm{ref}\neq0$, the derivative with respect to $s$ vanishes as well.
If only the rotational strain vanishes, $\xi_{\mathrm{rot}}=0$, the derivative remains finite and corresponds to a pure translational twist.
The problem is therefore not the model itself, but a naive implementation that evaluates indeterminate closed-form ratios such as $0/0$.

In \ac{SoRoMoX}, these removable singularities are handled in the low-level Lie-algebra and constant-strain routines.
To make the implementation pattern concrete, consider again
\begin{equation}
    c(\vartheta)=\frac{1-\cos\vartheta}{\vartheta^2},
    \qquad
    \lim_{\vartheta\rightarrow0}c(\vartheta)=\frac{1}{2}.
\end{equation}
A scalar implementation can use a true conditional to evaluate a Taylor approximation near the origin and the closed form elsewhere.
In JAX, \texttt{lax.cond} provides this behavior for a scalar predicate.
However, when \texttt{vmap} vectorizes that predicate, the conditional can be lowered to an elementwise selection for which both candidate expressions are evaluated.
The unused closed-form value may then contain a NaN, which can also contaminate reverse-mode derivatives.

The solution is to guard the \emph{input} of every potentially singular operation before selecting its output.
For the Lie-algebra argument $\chi=s\,\xi$, let
$\chi_\mathrm{rot}=(s\,\xi)_\mathrm{rot}=s\,\xi_\mathrm{rot}$ and
$\varrho=\chi_\mathrm{rot}^\top\chi_\mathrm{rot}=\|\chi_\mathrm{rot}\|^2$ denote its squared rotational magnitude.
With a small-angle threshold $\varepsilon>0$, the implementation follows the schematic construction
\begin{align}
    \varrho_\mathrm{safe}
        &=
        \begin{cases}
            1, & \text{if } \varrho\leq\varepsilon^2, \\
            \varrho, & \text{otherwise},
        \end{cases} \\
    \vartheta_\mathrm{safe}
        &= \sqrt{\varrho_\mathrm{safe}}, \\
    c_\mathrm{Taylor}(\varrho)
        &= \frac{1}{2}-\frac{\varrho}{24}
           +\frac{\varrho^2}{720}-\cdots, \\
    c_\mathrm{closed}(\vartheta_\mathrm{safe})
        &= \frac{1-\cos\vartheta_\mathrm{safe}}
                 {\vartheta_\mathrm{safe}^2}, \\
    c_\varepsilon(\chi)
        &=
        \begin{cases}
            c_\mathrm{Taylor}(\varrho),
                & \text{if } \varrho\leq\varepsilon^2, \\
            c_\mathrm{closed}(\vartheta_\mathrm{safe}),
                & \text{otherwise}.
        \end{cases}
\end{align}
The five lines respectively guard the squared magnitude, recover a finite angle, evaluate the Taylor approximation, evaluate the closed form, and select the appropriate result.
The value $1$ is only a finite placeholder in the discarded closed-form branch.
Consequently, both candidate computations remain finite even if vectorization evaluates both of them, and the final selection recovers the desired Taylor or closed-form value.
The same input-before-output guarding is used for the related Lie coefficients and rotational magnitude, while constant-strain tangent derivatives use explicit analytical directional derivatives.
The construction covers both zero arclength and zero rotational strain and remains well defined under batched forward- and reverse-mode differentiation.
The goal is not to change the mathematical model, but to evaluate its removable singularities through the correct analytical limits in the transformed numerical program as well.

\subsection{Strategies for Verifying the Mathematical Correctness of the Implementation}\label{sec:app:verification}
The (mathematical) complexity of most soft robot models makes it essential to verify implementation correctness.
In contrast to articulated rigid systems, where for low-\ac{DOF} open serial chains the kinematics and dynamics can often be derived ``by hand’’ or symbolically and then used to validate a numerical implementation, this is far less straightforward for soft robots: even for a single \ac{CS}~\cite{renda2018discrete} segment, the expressions—especially for the dynamics—quickly become long and intricate.
This calls for alternative (unit-)testing strategies that, for example, leverage modern techniques such as \ac{AD}.
Despite its importance and nontriviality, this topic has, to the best of our knowledge, not yet been systematically discussed in the soft robotics literature.

Therefore, we implemented extensive unit testing that verifies the correctness of the implementation of the soft robot models. Specifically, we test via \ac{CI} the following aspects: (i) the correctness of the SE(2) and SE(3) lie algebra underlying most models, (ii) the outputs of, for example, the forward kinematics at configurations for which closed-form solutions are easy to obtain (e.g., no rotational strains, single planar constant strain robots, etc), (iii) consistencies via the methods included in the same model, such as the forward-inverse kinematics loop, batched vs. pointwise functions (e.g., forward kinematics, Jacobians), bodyframe vs. inertia frame Jacobians, forward dynamics in the fused version vs. single calls of the dynamical matrices, etc., (iv) consistency of the explicit solutions to Jacobians $J$, Jacobian time derivatives $\dot{J}$, Coriolis matrix $C(q,\dot{q})$, and the gravitational forces vector $G(q)$ with the results obtained using \ac{AD} on the forward kinematics, the Jacobian, Christoffel symbols of the inertia matrix, and of the kinetic and gravitational potential energy, respectively, (v) the coherence in between the various models (e.g., planar \ac{PCS} vs. 3D \ac{PCS} vs. \ac{GVS}) for the same, randomized system parameters and configurations, (vi) the differentiability of the system kinematics and dynamics for forward-mode and reverse-mode \ac{AD} - particularly around singularities such as the straight configuration~\cite{della2020improved} - ensuring that the gradient obtained via \ac{AD} is always well-defined. In the following, we will discuss strategies (ii), (iii), (iv), and (v) in more detail.
For example, targeted regressions differentiate vectorized forward kinematics in reverse mode at the straight configuration and require bundled constant-strain operators to agree with the corresponding individually evaluated expressions. Automated tests also verify numerical equivalence between the JAX and Warp kernels.

\subsubsection{Evaluation of Lie Algebra in Special Cases}
In our unit tests for the $\mathfrak{se}(3)$ utilities\footnote{Please note that for conciseness, we focus on the testing procedure for the spatial/3D, but all outlined strategies naturally transfer to the planar case.}, we first verify that the hat operator implements the standard twist--matrix correspondence. For $\xi = (\omega, v)^\top \in \mathbb{R}^6$ with $\omega, v \in \mathbb{R}^3$ and $[\omega]_\times \in \mathfrak{so}(3)$, we check that
\begin{equation}
  \widehat{\xi}
  = \operatorname{hat}_{SE(3)}(\xi)
  =
  \begin{bmatrix}
    [\omega]_\times & v \\
    0_{1\times 3} & 0
  \end{bmatrix},
\end{equation}
and, for planar twists $\xi_{\mathrm{2D}} \in \mathbb{R}^3$ lifted into the spatial (strain) case via a fixed linear map $L:\mathbb{R}^3\to\mathbb{R}^6$, that the induced $4\times 4$ matrices reduce to the $\mathfrak{se}(2)$ representation $\Pi_{\mathrm{2D}}\!\bigl(\operatorname{hat}_{SE(3)}(L\xi_{\mathrm{2D}})\bigr) = \operatorname{hat}_{SE(2)}(\xi_{\mathrm{2D}})$,
where $\Pi_{\mathrm{2D}}$ projects the homogeneous $4\times 4$ matrix to its planar $3\times 3$ block. The exponential maps are tested by checking consistency with the matrix exponential and with the planar embedding: for random $\xi \in \mathbb{R}^6$ we assert $\operatorname{exp}_{SE(3)}(\xi) \approx \exp\bigl(\widehat{\xi}\bigr)$ component-wise up to a numerical tolerance, and for planar twists $\xi_{\mathrm{2D}}$ we require
\begin{equation}
  \Pi_{\mathrm{2D}}\!\bigl(\operatorname{exp}_{SE(3)}(L\xi_{\mathrm{2D}})\bigr)
  =
  \operatorname{exp}_{SE(2)}(\xi_{\mathrm{2D}}).
\end{equation}
The logarithm $\log_{SE(3)}$ is checked via round-trip and special-case tests: for pure rotations
\begin{equation*}
  g =
  \begin{bmatrix}
    R & 0\\
    0 & 1
  \end{bmatrix}
  \quad\Rightarrow\quad
  \log_{SE(3)}(g)
  =
  \begin{bmatrix}
    \omega\\
    0
  \end{bmatrix},
\end{equation*}
analogue for pure translations,
and for random twists $\xi$ we enforce the round-trip property $  \log_{SE(3)}\!\bigl(\operatorname{exp}_{SE(3)}(\xi)\bigr) \approx \xi$ while also checking numerical well-posedness at the identity $g = \mathbb{I}_4$ (no NaNs and $\log_{SE(3)}(\mathbb{I}_4) = 0$). The adjoint and coadjoint maps are validated against their known block structure: Namely, we assert
\begin{equation*}
\begin{split}
	\operatorname{Ad}^{SE(3)}_g(\xi)
  &=
  \begin{bmatrix}
    R & 0\\
    [p]_\times R & R
  \end{bmatrix}
  \begin{bmatrix}
    \omega\\
    v
  \end{bmatrix},\\
  \operatorname{Ad}^{*,SE(3)}_g(\xi) &=
  \begin{bmatrix}
    R & [p]_\times R\\
    0 & R
  \end{bmatrix},
\end{split}
\end{equation*}
that the inverse of the Adjoint is given by applying the Adjoint to the inverse homogenous transformation, i.e., $\operatorname{Ad}^{SE(3)}_{g^{-1}}\,
  \operatorname{Ad}^{SE(3)}_g
  = \mathbb{I}_6$,
and reduces to the planar formulas when restricted via the same $L$ and $\Pi_{\mathrm{2D}}$ (consistency with the $\operatorname{Ad}_{SE(2)}$ and its coadjoint $\operatorname{Ad}^*_{SE(2)}$). Finally, the left-trivialized tangent map and its derivative, $\operatorname{T}^{SE(3)}_{g_i}(\xi)$ and $\dot{\operatorname{T}}^{SE(3)}_{gi}(\xi, \dot{\xi)}$, are tested by enforcing the first-order consistency condition for small $\lVert\delta\xi\rVert$,
\begin{equation*}
  \operatorname{exp}_{SE(3)}(\xi + \delta\xi)
  \approx
  \operatorname{exp}_{SE(3)}(\xi)\,
  \operatorname{exp}_{SE(3)}\!\bigl(\operatorname{T}^{SE(3)}_{g_i}(\xi)\,\delta\xi\bigr),
\end{equation*}
checking that $\operatorname{T}^{SE(3)}_{g_i}(0) = \mathbb{I}_6$ and $\dot{\operatorname{T}}^{SE(3)}_{g_i}(\xi, 0) = 0$, and comparing the explicitly implemented $\dot{\operatorname{T}}(\xi, \dot{\xi})$ against $\operatorname{AD}_{\xi} (\operatorname{T}(\xi)) \, \dot{\xi} = \frac{\partial T}{\partial \xi} \, \dot{\xi}$, which is obtained via JAX \ac{AD}, while asserting that all entries remain finite at $\xi = 0$.

\subsubsection{Evaluation of Special Cases of the Kinematics and the Kinematic Loop}
First, consider a single-segment \ac{PCS} soft robot with reference strain $\xi_\mathrm{ref}=[0,0,0,1,0,0]^\top$, corresponding to a straight, unelongated shape. Then $\operatorname{FK}(0_n,s)$, where $q=0_n$ denotes zero strain displacement rather than zero full strain, gives the translation $p=[s,0,0]^\top$ and identity rotation $R=\mathbb{I}_3$. A pure axial-strain displacement $\Delta\sigma_\mathrm{ax}$ gives $p=[s(1+\Delta\sigma_\mathrm{ax}),0,0]^\top$, again with identity rotation; equivalently, in terms of the full axial strain $\sigma_\mathrm{ax}=1+\Delta\sigma_\mathrm{ax}$, this is $p=[s\sigma_\mathrm{ax},0,0]^\top$. Similarly, we can consider a single-segment \ac{PCC} robot and compare the forward kinematics and the Jacobian against the expressions given in \cite{webster2010design, della2020improved}.

Next, consider a \ac{PCS} soft robot, where the functions $g_\mathrm{tips} = \operatorname{FK}_\mathrm{tips}(q)$ and $q = \operatorname{IK}_\mathrm{tips}(g_\mathrm{tips})$ define the closed-form forward and inverse kinematics for the segment tips, respectively. Here, $g_\mathrm{tips} = (g_{\mathrm{tip},1}, \dots, g_{\mathrm{tip},N})$ is a tuple that contains the homogeneous transformation $g_{\mathrm{tip},i} = \operatorname{FK}_\mathrm{tip}(q,s_\mathrm{tip}) \in SE(3)$ to the tip of the $i$-th segment. Then, we need to naturally assert that $q = \operatorname{IK}_\mathrm{tips}(\operatorname{FK}_\mathrm{tips}(q))$.

\subsubsection{Test of Consistency of Kinematics and Dynamics via Auto-Differentiation}


Many expressions in (soft) robot models (e.g., Jacobians, Jacobian derivatives, Coriolis matrix, potential forces) can be written as derivatives of other quantities (e.g., forward kinematics, Jacobians, inertia matrix, potential energy). Consequently, they could in principle be implemented using \ac{AD} techniques in \emph{AD}. However, this is typically not desirable in practice when an explicit closed-form expression is available and can be implemented directly, since the use of \ac{AD} usually incurs higher computational compilation-, run-time, and memory costs, in particular when the algorithm is cascaded with higher-order derivatives (e.g., differentiating the state trajectory with respect to the initial condition).

In the following, let $\operatorname{AD}_{q}(f(q)) = \frac{\partial f}{\partial q}(q)$ denote the gradient of the function $f(q)$ with respect to $q$ obtained via auto-differentiation (e.g., the JAX \texttt{jacfwd} and \texttt{jacrev} functions).

\paragraph{Forward Kinematics $\rightarrow$ Inertial Frame Jacobian}

Let $g(q,s)\in SE(3)$ denote the homogeneous transform with rotation $R(q,s)\in\mathrm{SO}(3)$ given in explicit form, and let ${}^\mathrm{s}J(q,s) \in \mathbb{R}^{6 \times n}$ denote the explicit expression for the inertial-frame (spatial) Jacobian.

For each column $j=1,\dots,n$, we compute the directional derivative $\partial_j g(q,s) := \operatorname{AD}_q(g(q,s))[e_j]\in\mathbb{R}^{4\times 4}$, where $e_j$ is the $j$-th unit vector in Euclidean space, and left-trivialize it to the body Lie algebra by ${}^\mathrm{b}\widehat{\eta}_j := g(q,s)^{-1} \partial_j g(q,s) \in \mathfrak{se}(3)$. Writing
\[
{}^\mathrm{b}\widehat{\eta}_j
=
\begin{bmatrix}
[{}^\mathrm{b}\omega_j]_{\times} & {}^\mathrm{b}v_j \\
0^\top & 0
\end{bmatrix},
\]
with $[{}^\mathrm{b}\omega_j]_{\times} \in \mathbb{R}^{3 \times 3}$ the skew-symmetric matrix of the angular velocity vector, we obtain the body twist vector
\[
{}^\mathrm{b}\eta_j
=
\begin{bmatrix}
({}^\mathrm{b}\omega_j)^\top & ({}^\mathrm{b}v_j)^\top
\end{bmatrix}^\top.
\]
We then express this twist in the inertial frame using the pure-rotation adjoint (i.e., without translational coupling), ${}^\mathrm{s}\omega_j := R(q,s)\,{}^\mathrm{b}\omega_j$ and ${}^\mathrm{s}v_j := R(q,s)\,{}^\mathrm{b}v_j$, and set the $j$-th column of ${}^\mathrm{s}J_{\mathrm{AD}}(q,s)$ to
\[
\begin{bmatrix}
({}^\mathrm{s}\omega_j)^\top & ({}^\mathrm{s}v_j)^\top
\end{bmatrix}^\top.
\]
Finally, we assert element-wise agreement of
\[
{}^\mathrm{s}J_\mathrm{AD}(q,s) = {}^\mathrm{s}J(q,s).
\]

\paragraph{Jacobian $\rightarrow$ Jacobian Derivative}
Let $J(q,s)$ denote the inertial- or body-frame Jacobian and $\dot{J}(q,\dot{q},s)$ the corresponding time derivative, which can be written in explicit form for each system~\cite{renda2018discrete, mathew2025reduced}. Then we assert that
\[
\dot{J}(q,\dot{q},s) = \operatorname{AD}_{q}(J(q,s)) \, \dot{q}.
\]

\paragraph{Inertia Matrix $\rightarrow$ Coriolis Force}
Let $\tau_\mathrm{C} = C(q,\dot{q}) \, \dot{q}$ denote the Coriolis force computed using a Coriolis matrix in an arbitrary but valid factorization given in explicit form. We can verify this expression by comparing it to a Coriolis matrix derived via Christoffel symbols~\cite{siciliano2025foundations}:
\[
\begin{split}
    \tau_{\mathrm{C},i}
    &= \sum_{j=1}^n C_{ij}(q,\dot{q}) \, \dot{q}_j\\
    &=  \sum_{j = 1}^{n} \sum_{k = 1}^n c_{ijk} \, \dot{q}_j \, \dot{q}_k - \frac{1}{2} c_{jki} \, \dot{q}_j \, \dot{q}_k,
    \: \forall \, i \in \mathbb{N}_n,
\end{split}
\]
with $c_{ijk} = \operatorname{AD}_{q_k}(M_{ij}(q))$.

\paragraph{Kinetic Energy $\rightarrow$ Inertia Matrix, Coriolis Force}

If $\mathcal{T}(q,\dot{q})$ and $M(q)$ denote explicit expressions for the kinetic energy and inertia matrix, respectively, we need to assert that
\[
M(q) = \operatorname{AD}_{\dot{q}^2}(\mathcal{T}(q,\dot{q})),
\]
i.e., that $M(q)$ coincides with the second derivative of $\mathcal{T}$ with respect to $\dot{q}$~\cite{della2023model, liu2024physics, valadas2025learning}.

Similarly, the Coriolis force, independently of the chosen factorization of the Coriolis matrix, must satisfy
\[
\tau_\mathrm{C} = C(q,\dot{q}) \, \dot{q}
= \operatorname{AD}_{q}\bigl(\operatorname{AD}_{\dot{q}}(\mathcal{T}(q,\dot{q}))\bigr).
\]

\paragraph{Potential Energy $\rightarrow$ Potential Forces}
Given explicit expressions $\mathcal{U}(q)$ and $\tau_\mathcal{U}(q)$ for the potential energy and potential generalized force, respectively, the following relation must hold $\tau_\mathcal{U}(q) = \operatorname{AD}_{q}(\mathcal{U}(q))$ (see, e.g.,~\cite{siciliano2025foundations, della2023model, stolzle2025phdthesis}).

\paragraph{Actuator Coordinates $\rightarrow$ Actuation Matrix}
Let $\varphi_\mathrm{a}(q) \in \mathbb{R}^m$ denote the actuator coordinates, with corresponding actuation matrix $A(q) \in \mathbb{R}^{n \times m}$. Work conjugacy requires~\cite{murray2017mathematical, pustina2024input}
\[
    A^\top(q) = \operatorname{AD}_q\!\left(\varphi_\mathrm{a}(q)\right).
\]
For the tendon preset, $\varphi_\mathrm{a}(q)=-\ell(q)$, and therefore
\[
    A^\top(q) = -\operatorname{AD}_q\!\left(\ell(q)\right).
\]

\subsubsection{Verification of Inter-System Coherence}
For rod-based models, specialized formulations—such as \ac{PCC}~\cite{webster2010design}, polynomial curvature~\cite{della2019control}, and (Planar) \ac{PCS}~\cite{renda2016discrete, renda2018discrete}, which neglect certain strains and/or use specific (functional) strain bases—can typically be embedded in more general models like \ac{VS}~\cite{renda2020geometric, boyer2020dynamics, mathew2025reduced}, enabling systematic coherence tests between different systems.

We next outline a step-by-step procedure to verify coherence between the \emph{Planar PCS} and \emph{GVS} implementations.
First, we randomly sample system parameters (e.g., number of segments, active strains, segment lengths, mass density, and reference strain) for the \emph{Planar PCS} system and use them to construct the corresponding spatial \ac{GVS} model by selecting constant-strain bases for the circular segments and connecting them via fixed joints. The two models are assigned the same physical reference strain under the planar-to-spatial embedding, including the same axial reference component; for a straight, unelongated backbone, this component equals one.
Second, we sample \emph{Planar PCS} configurations and time derivatives $q, \dot{q}$, as well as backbone abscissae $s$, and lift each planar configuration to a spatial one by setting the out-of-plane generalized-coordinate components to zero (i.e., one bending component, the twist component, and one shear component). Because the reference strains are matched separately, equal lifted configurations then produce equal full strain fields.
Third, we evaluate the relevant class functions of both systems at $(q, \dot{q}, s)$, including forward kinematics, the Jacobian–velocity product $J(q)\dot{q}$, inertia and Coriolis matrices, potential forces, and kinetic and potential energies.
If necessary, we lift poses and velocities from $SE(2)$ to $SE(3)$.
Finally, we check that the lifted outputs from the \emph{Planar PCS} model agree with the \ac{GVS} predictions within a prescribed tolerance that accounts for floating-point errors and small discrepancies introduced by Gaussian quadrature.

\end{document}